\documentclass[runningheads]{llncs}

\usepackage{eccv}

\usepackage{wrapfig}
\usepackage{eccvabbrv}

\usepackage{graphicx}
\usepackage{booktabs}
\usepackage{longtable}
\usepackage{algorithm}
\usepackage{algorithmic}
\usepackage[table]{xcolor}
\usepackage{multirow}

\usepackage[accsupp]{axessibility}  

\usepackage{hyperref}

\usepackage{orcidlink}

\begin{document}

\title{EMAG: Self-Rectifying Diffusion Sampling with Exponential Moving Average Guidance} 


\author{Ankit Yadav\inst{1}\orcidlink{0009-0006-8323-3919} \and
Ta Duc Huy\inst{1}\orcidlink{0000-0001-6181-0478} \and
Lingqiao Liu\inst{1}\orcidlink{0000-0003-3584-795X}}

\authorrunning{A. Yadav et al.}
\titlerunning{EMAG: Self-Rectifying Diffusion Sampling with EMA Guidance}


\institute{Australian Institute for Machine Learning, Adelaide University, Australia \\
\email{\{ankit.yadav, huy.ta, lingqiao.liu\}@adelaide.edu.au}}

\maketitle

\begin{abstract}
  In diffusion and flow-matching generative models, guidance techniques are widely used to improve sample quality and consistency. Classifier-free guidance (CFG) is the de facto choice in modern systems and achieves this by contrasting conditional and unconditional samples. Recent work explores contrasting negative samples at inference using a weaker model, via strong/weak model pairs, attention-based masking, stochastic block dropping, or perturbations to the self-attention energy landscape. While these strategies refine the generation quality, they still lack a reliable control over the granularity or difficulty of the negative samples, and target-layer selection is often fixed. We propose \emph{Exponential Moving Average Guidance (EMAG)}, a training-free mechanism that modifies attention at inference time in diffusion transformers, with a statistics-based, adaptive layer-selection rule. Unlike prior methods, EMAG produces harder, semantically faithful negatives (fine-grained degradations), surfacing difficult failure modes, enabling the denoiser to refine subtle artifacts, boosting the quality and human preference score (HPS) by \textbf{+0.54} over CFG. We further demonstrate that EMAG naturally composes with advanced orthogonal guidance techniques, such as APG and CADS, further improving HPS. Code is available at \url{https://github.com/drkkgy/EMAG}.
\end{abstract}

\section{Introduction}
\label{sec:intro}

In recent times, Denoising Diffusion Models (DDMs) \cite{sohl2015deep,ho2020denoising,dhariwal2021diffusion,song2020score} and flow\mbox{-}matching models \cite{lipman2022flow,esser2024scaling} have surpassed GANs \cite{goodfellow2020generative} at modeling data across modalities (images \cite{rombach2022high}, audio \cite{wang2023audit}, video \cite{blattmann2023stable}). Their success stems from an iterative denoising process that starts from Gaussian noise and progressively removes it to sample the target distribution \cite{kingma2023understanding}. The advent of conditional diffusion models, where a condition such as a class \cite{dhariwal2021diffusion} or text prompt \cite{nichol2021glide} can be used to guide the generation, further propelled their popularity. However, naive guidance signals often result in insufficient quality and weaker prompt adherence \cite{mukhopadhyay2023diffusion}, leading to the introduction of strategies like classifier guidance \cite{dhariwal2021diffusion} and classifier-free guidance \cite{ho2022classifier}, where the prior method requires an explicit guidance model while the latter relies on the implicit Bayesian classifier of the same diffusion model by contrasting the conditional with the unconditional signal. This substantially improved conditional generation quality and prompt adherence, but can often reduce diversity and induce over\mbox{-}saturation \cite{sadat2024eliminating}. Recent work addresses these effects by providing more semantically aware guidance (S-CFG) \cite{shen2024rethinking}, decomposing the CFG update into parallel and orthogonal components and down-weights the parallel term (with rescaling/momentum) to mitigate oversaturation (APG) \cite{sadat2024eliminating}, and by perturbing the conditioning signal over an annealed schedule improving diversity (CADS) \cite{sadat2023cads}.

\begin{wrapfigure}{r}{0.5\linewidth}
\vspace{-24pt}

  \centering

  \includegraphics[width=1\linewidth]{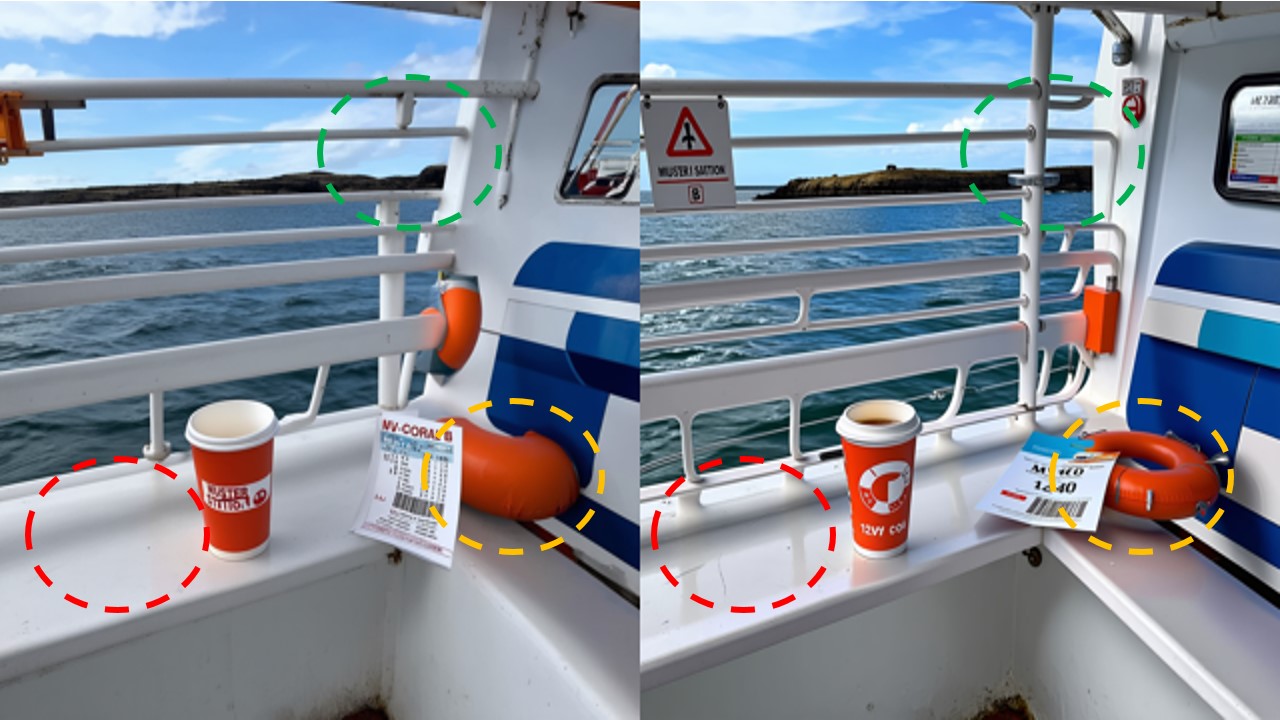}
   
    \caption{\textbf{CFG (left) vs. EMAG (right)} Compared to CFG \cite{ho2022classifier}, EMAG produces more \textit{semantically plausible} images, preserving global structure while sharpening fine details and suppressing minor artifacts. When applied to SD3-Medium, EMAG’s outputs align more closely with human-preference proxies (HPS \cite{wu2023human}).
    }
  \label{fig:front_image}
\vspace{-24pt}
\end{wrapfigure}

Another line of work exposes failure modes by contrasting strong and weak predictions. Such negative (weak) signals are obtained by perturbing attention maps \cite{ahn2024self,hong2023improving} or weights \cite{hong2024smoothed}, by guiding with deliberately weaker models (reduced capacity or degraded variants) \cite{karras2024guiding}, by sampling sub-networks \cite{chen2025s}, or via energy-based formulations such as ERG \cite{ifriqi2025entropy} and SEG \cite{hong2024smoothed}. These strategies are attractive because, unlike CFG, they apply to both conditional and unconditional settings and often yield \emph{Pareto-favorable} trade-offs. However, most approaches provide a mechanism to obtain an inferior model but do not explicitly construct \emph{hard negatives}. We hypothesize that if degradations are too strong (e.g., heavily distorting an image), they represent failure modes that modern diffusion models are implicitly powerful enough to avoid \cite{peebles2023scalable,esser2024scaling}, yielding \emph{easy} negatives. 

Motivated by this, we propose Exponential Moving Average Guidance (EMAG), which, unlike other approaches, by design can regulate how much \emph{high\mbox{-}frequency} information (fine-grained details) the weaker model is allowed to destroy by controlling the EMA updates, deliberately injecting subtle, semantically faithful artifacts. These yield \emph{hard negatives} that reveal failure modes that base models are more likely to miss, enabling more effective rectification during sampling, see Fig.~\ref{fig:front_image}. 
Empirically, EMAG improves HPS by +0.54 (Table~\ref{tab:main_results_SD3_cond}) over CFG and yields additional gains when combined with stronger orthogonal guidance baselines such as APG and CADS across most experimental settings, underscoring the value of controllable, fine-grained negatives in high-quality regimes. We also evaluate EMAG as a standalone guidance method against other baselines (see Supplementary Table~S17\footnote{Figures and tables prefixed with ``S'' refer to the supplementary material.}).

\paragraph{Contributions}
\begin{itemize}
    \item We propose \textbf{EMAG}, a training\mbox{-}free attention-map EMA guidance that produces \emph{controllable} weak signals without extra training. We also study the impact of negative sample hardness on guidance effectiveness, use cosine-similarity to quantify it, and show that EMAG's EMA $\beta$ controls it.
    \item We propose a statistics\mbox{-}based \emph{adaptive layer selection} that chooses target layers per timestep to create controllable degradations in negative samples.
    \item We conduct extensive experiments on DiT~\cite{peebles2023scalable} and MMDiT (SD3~\cite{esser2024scaling}) backbones, covering both class\mbox{-}conditional and text\mbox{-}to\mbox{-}image generation, and report competitive performance against strong guidance baselines, including notable improvement in Human Preference Score (HPS)~\cite{wu2023human}.
    \item We demonstrate the complementary nature of our approach with advanced orthogonal guidance methods such as APG\cite{sadat2024eliminating} and CADS\cite{sadat2023cads}, further improving HPS.
\end{itemize}


\section{Related work}
\label{sec:Related work}

\begin{figure*}[t]
  \centering
  \begin{subfigure}[t]{0.49\linewidth}
    \centering
    \includegraphics[width=\linewidth]%
      {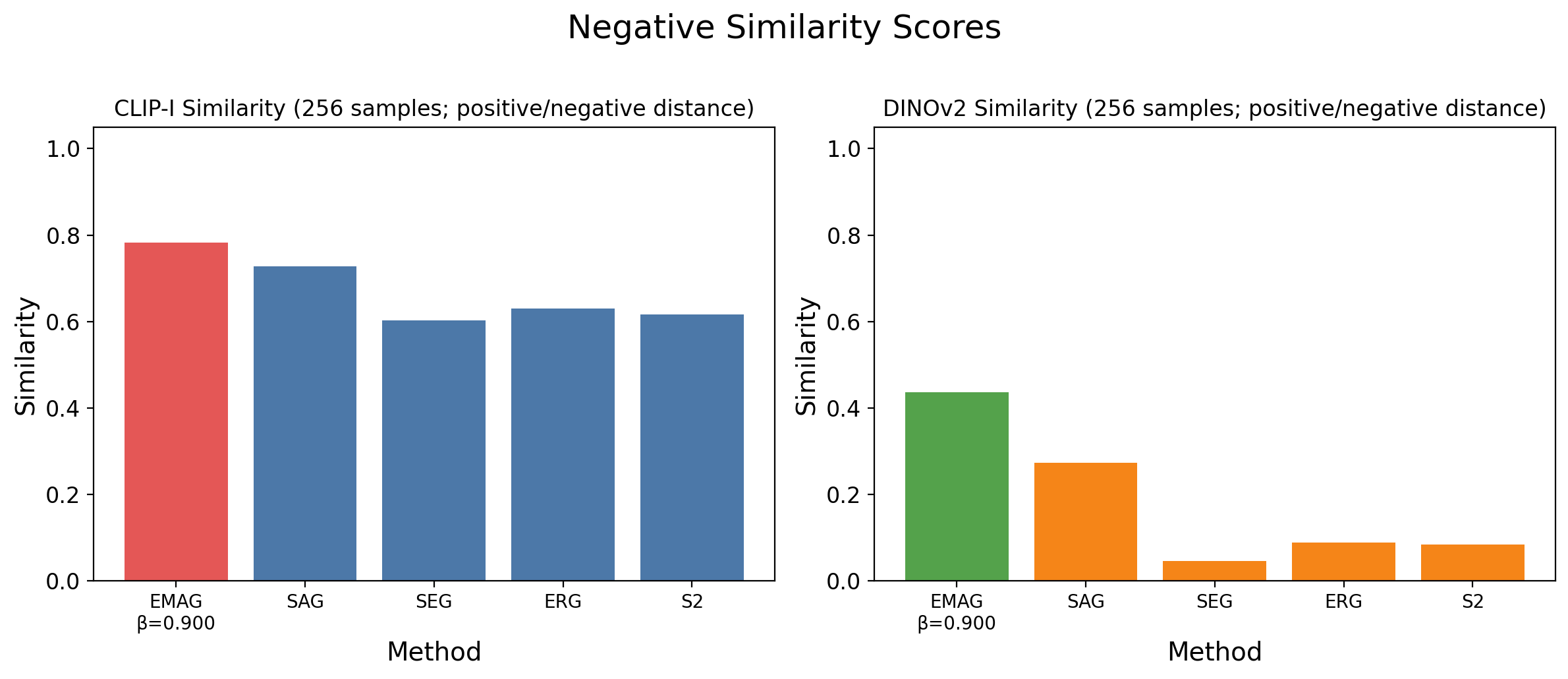}
    \caption{Mean similarity scores between positive and negative 
    samples across different baselines and EMAG. Higher similarity 
    indicates harder negative.}
    \label{fig: section_F_hard_negative_similarity_scores}
  \end{subfigure}
  \hfill
  \begin{subfigure}[t]{0.49\linewidth}
    \centering
    \includegraphics[width=\linewidth]%
      {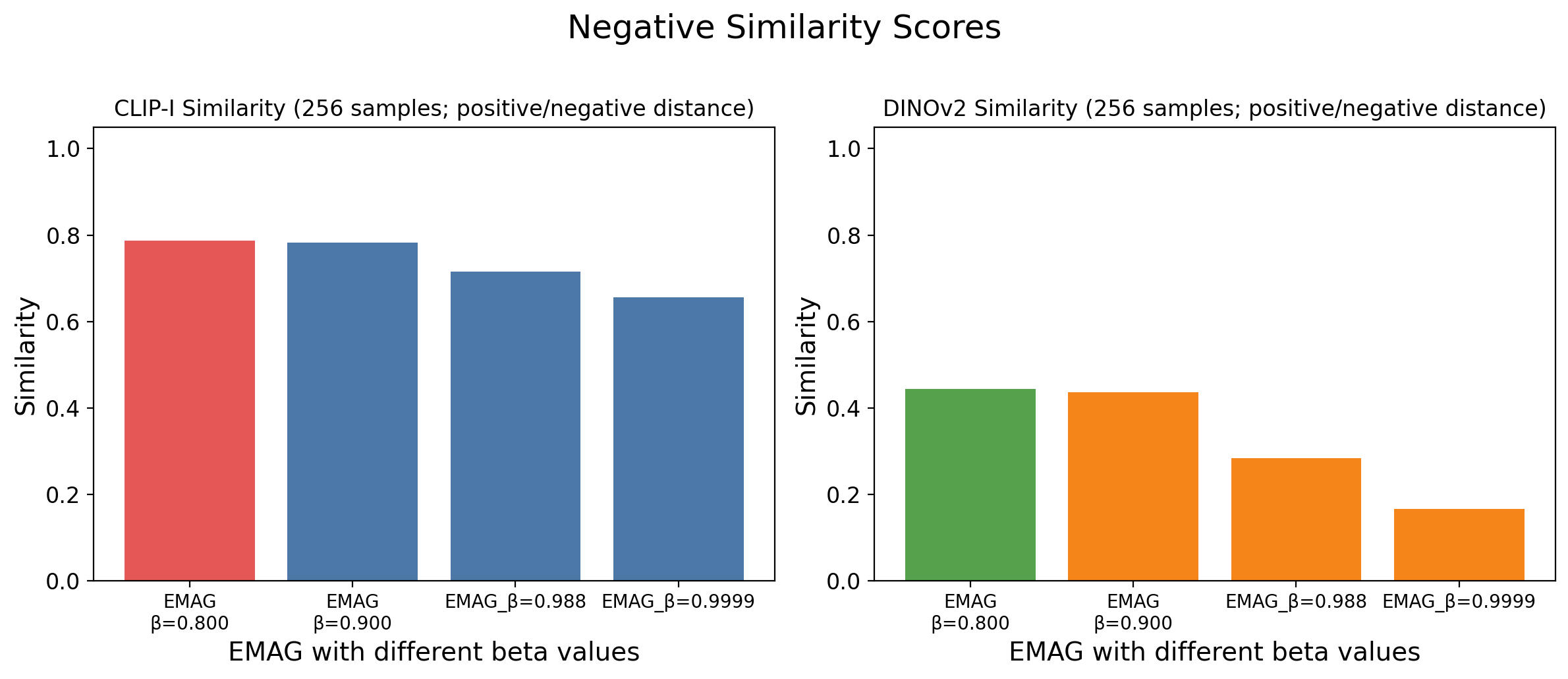}
    \caption{Mean similarity scores across different $\beta$ values 
    for EMAG. Higher similarity indicates harder negative.}
    \label{fig: section_F_hard_negative_similarity_scores_different_beta}
  \end{subfigure}
  \caption{\textbf{Negative sample hardness quantified via cosine 
  similarity (CLIP-I and DINOv2).} 
  \textbf{(a)}~EMAG produces harder negatives than prior guidance 
  methods across both similarity metrics. 
  \textbf{(b)}~Within EMAG, the decay factor $\beta$ directly 
  controls negative hardness.}
  \label{fig:hard_negative_similarity_combined}
\end{figure*}

Classifier-free guidance (CFG) \cite{ho2022classifier} combines conditional and unconditional predictions to improve prompt alignment and perceptual quality, but suffers from reduced generation diversity \cite{astolfi2024consistency, ifriqi2025entropy, sadat2023cads} and visual over-saturation \cite{ho2022classifier}.
To alleviate these issues, recent studies have revisited CFG and developed strategies that better balance semantic alignment and visual fidelity. Adaptive Projected Guidance (APG) \cite{sadat2024eliminating} decomposes the CFG update into orthogonal/parallel components and down-weights saturation-prone components, enabling higher guidance scales with fewer artifacts. Condition-Annealed Diffusion Sampler (CADS) \cite{sadat2023cads} boosts diversity via annealing the conditioning strength during inference, especially at high guidance scales.

A useful unifying view is \emph{auto-guidance} \cite{karras2024guiding}, where a strong model is contrasted with a deliberately weakened (reduced parameters or training steps) variant to steer sampling. This separation helps stabilize updates and expose failure modes, highlighting the artifacts typical of high guidance scales. \emph{Karras et al.}\ report that auto-guidance disentangles control over image quality while preserving variation, yielding more favorable quality–diversity trade-offs than CFG \cite{karras2024guiding}.

\begin{wrapfigure}{r}{0.50\linewidth}
\vspace{-32pt}
  \centering
  \includegraphics[width=\linewidth]{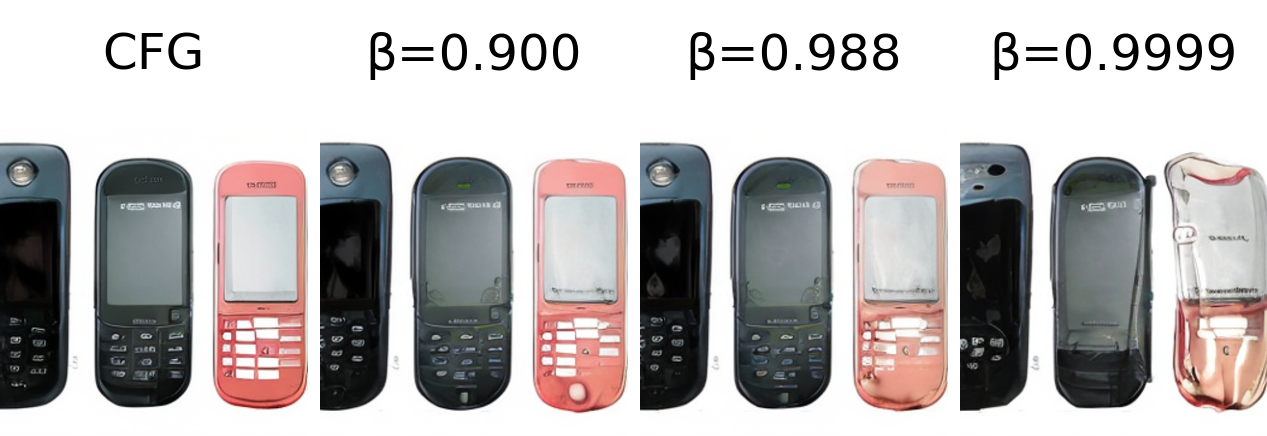}
    \caption{\textbf{CFG(left) vs\ EMAG hard negatives(right)} EMAG samples at varying $\beta$; larger $\beta$ produces stronger degradations.}
  \label{fig:Negative_sample_controlled}
\vspace{-16pt}
\end{wrapfigure}

However, explicitly maintaining strong/weak model pairs is not always practical; several strategies have been introduced to achieve a similar effect with a single model. Self-Attention Guidance (SAG) degrades the \emph{input} by blurring regions indicated by self-attention, yielding a weaker score that guides the original model \cite{hong2023improving}. Smoothed Energy Guidance (SEG) weakens the prediction by smoothing the attention energy (Gaussian blurring of attention weights), with an efficient \emph{query blurring} equivalent; in practice, one can approximate this by \emph{replacing image queries with their channel-wise means} \cite{hong2024smoothed}. Entropy Rectifying Guidance (ERG) perturbs the attention mechanism (including conditioning pathways) at inference to improve quality, diversity, and prompt consistency \cite{ifriqi2025entropy}. Perturbed-Attention Guidance (PAG) forms weak intermediates by replacing selected self-attention maps with the identity and then guiding away from them \cite{ahn2024self}. S$^2$-Guidance (Stochastic Self-Guidance) constructs weak sub-networks by stochastically dropping layers \cite{chen2025s}.
While guiding with weaker or negative guidance signals can outperform CFG, most methods offer limited control over the \emph{quality} (hardness and granularity) of these negatives. We introduce \emph{Exponential Moving Average Guidance (EMAG)}, a training-free approach that generates \emph{controllable, semantically faithful hard negatives} by regulating fine-grained (high-frequency) degradations through EMA-based attention smoothing, where the decay factor $\beta$ directly controls the degradation strength, together with adaptive layer selection. This design exposes hard failure modes that coarse degradations often miss and complements advanced guidance methods such as APG and CADS.

\section{Method}
\label{sec:Method}

\subsection{Diffusion preliminary}

\paragraph{Diffusion preliminaries.}

In diffusion-based generation~\cite{dhariwal2021diffusion,lipman2022flow,esser2024scaling}, a clean image \(x_0\) is progressively corrupted with Gaussian noise \(\epsilon\!\sim\!\mathcal{N}(0,1)\) over timesteps \(t\in[0,T]\), yielding noisy samples \(x_t\) and this is known as the forward process. This can be represented as a stochastic differential equation (SDE) 
\begin{equation}
\label{eq:preelim-1}
    dx \;=\; f(x,t)\,dt \;+\; g(t)\,dw,
\end{equation}
where \(f(x,t)\) and \(g(t)\) control the corruption process and \(w\) is a standard wiener process. To recover \(x_0\) from \(x_t\), one integrates the reverse-time SDE~\cite{song2020denoising}
\begin{equation}
\label{eq:preelim-2}
    dx \;=\; \big[f(x,t) \;-\; g(t)^2 \nabla_x \log p_t(x)\big]\,dt \;+\; g(t)\,d\bar{w},
\end{equation}
where \(\nabla_x \log p_t(x)\) is the score of the perturbed distribution and \(\bar w\) is a standard Wiener process in reverse time. A neural network (e.g., a Transformer) is trained to approximate the score, \(e_\theta(x,t)\!\approx\!\nabla_x\log p_t(x)\), enabling generation from random Gaussian noise. With external conditioning \(c\) (e.g., class labels or text), the conditional score is learned, \(e_\theta(x,t,c)\!\approx\!\nabla_x\log p_t(x\mid c)\)~\cite{dhariwal2021diffusion}. 

\paragraph{Flow matching}
Flow matching trains a \emph{deterministic} neural ODE to transport a simple prior to the data by regressing a time-dependent velocity field:
\begin{equation}
\label{eq:fm-ode}
\dot{x} \;=\; v_\theta(x,t),
\end{equation}
with the standard regression objective
\begin{equation}
\label{eq:fm-loss}
\min_{\theta}\;
\mathbb{E}_{t\sim\mathcal{U}(0,1),\,(x_0,x_1)}\;
\big\|\,v_\theta\!\big((1{-}t)x_0{+}t x_1,\,t\big) \;-\; (x_1{-}x_0)\big\|_2^{2},
\end{equation}

where using the linear path yields the popular \emph{rectified flow} instance.
In contrast, diffusion employs a \emph{stochastic} reverse SDE or its deterministic probability-flow ODE; the latter evolves the same marginals as the reverse SDE:

\begin{equation}
\label{eq:pf-ode}
\frac{dx}{dt} \;=\; f(x,t)\;-\;\tfrac{1}{2}\,g(t)^{2}\,\nabla_x \log p_t(x).
\end{equation}

\begin{wraptable}{r}{0.48\linewidth }
\vspace{-36pt}
\centering
\small
\setlength{\tabcolsep}{6pt}
\caption{Comparison of SOTA CFG-complementary guidance methods with and without EMAG on SD3 (COCO 2014, identical setup). \emph{*CFG+EMAG $\equiv$ EMAG by design}}
\label{tab:combination_experiment}
\begin{tabular}{lcc}
\toprule
\textbf{Guidance} & \textbf{HPS} $\uparrow$ & \textbf{HPS} $\uparrow$ \\
\midrule
 &  & \textbf{+EMAG} \\
\midrule

\textbf{CFG}     & 29.22 & \textbf{29.76}* \\
\textbf{ERG}     & 29.35 & \textbf{29.56} \\ 
\textbf{S2}      & 29.15 & \textbf{29.51} \\
\textbf{SAG + CFG} & 29.28 & - \\ 
\textbf{SEG + CFG} & 29.25 & \textbf{29.29} \\ 
\midrule
\textbf{EMAG}   & \textbf{29.76} & - \\  
\bottomrule
\end{tabular}
\vspace{-20pt}
\end{wraptable}
\subsection{Existing guidance approaches}

Conditioned Diffusion models have recently achieved the capability to generate high-quality multimedia. Classifier Free Guidance (CFG) \cite{ho2022classifier} is one of the major breakthroughs behind this dramatic improvement in the generation quality, and prompt alignment, where the unconditional and conditional score estimates are combined to create the final estimate with the help of guidance scale as depicted in Eq. \ref{eq:cfg_eq} where $\epsilon_\theta$ is the denoiser neural network, $x_t$ is the input latent, $c$ is the conditioning input and $\varnothing$ is null input. Hence $\epsilon_\theta(x_t, \varnothing)$ and $\epsilon_\theta(x_t, c)$ are the denoiser network with null-conditioning (unconditional) and conditioning and guidance scale $w > 1$.  
\begin{equation}
    \label{eq:cfg_eq}
    \tilde{\epsilon}_\theta(x_t, c) = \epsilon_\theta(x_t, \varnothing) + w \cdot (\epsilon_\theta(x_t, c) - \epsilon_\theta(x_t, \varnothing))
\end{equation}

However, CFG suffers from inherent drawbacks such as reduced generation diversity and visual over-saturation . To alleviate these issues, recent studies have revisited CFG and developed strategies that better balance semantic alignment and visual fidelity. One such direction is Auto-guidance , which employs two versions of the denoiser that share the same network structure but differ in capacity. Specifically as shown in Eq. \ref{eq:autoguidance}, 
\begin{equation}
    \label{eq:autoguidance}
    \tilde{\epsilon}(x_t, c) = \epsilon_0(x_t, c) + w \cdot (\epsilon_1(x_t, c) - \epsilon_0(x_t, c))
\end{equation}where the weaker model is denoted as $\epsilon_0$ and the stronger model as $\epsilon_1$. Here the weaker model’s score prediction guides the stronger one in Eq. \ref{eq:autoguidance}. 

This Auto-guidance framework in Eq. \ref{eq:autoguidance} can be achieved by using different ways to 
construct the weaker model through controlled degradation. For example, SEG forms the weaker model by directly perturbing the denoiser’s attention weights, such as by adding Gaussian noise or replacing image queries with their channel-wise means. Let the perturbed denoiser be $\epsilon_1'$. With this weaker version defined, replacing $\epsilon_0(x_t,c)$ with $\epsilon_1'(x_t,\varnothing)$ in Eq. \ref{eq:autoguidance} produces the SEG guidance formulation (only for unconditional).
More recent approaches, such as ERG, extend this perturbation to both the network and the conditioning encoder, resulting in a perturbed denoiser $\epsilon_1'$ and perturbed conditioning $c'$. Substituting these into Eq. \ref{eq:autoguidance} leads to the ERG objective, i.e., replacing $\epsilon_0(x_t,c)$ with $\epsilon_1'(x_t,c')$.
\begin{equation}
    \label{eq:PAG}
    \tilde{\epsilon}(x_t) = \epsilon_1(x_t) + w \cdot (\epsilon_1(x_t) - \epsilon_1'(x_t))
\end{equation}

\begin{wrapfigure}{l}{0.50\linewidth}
\vspace{-26pt}
  \centering
  \includegraphics[width=\linewidth]{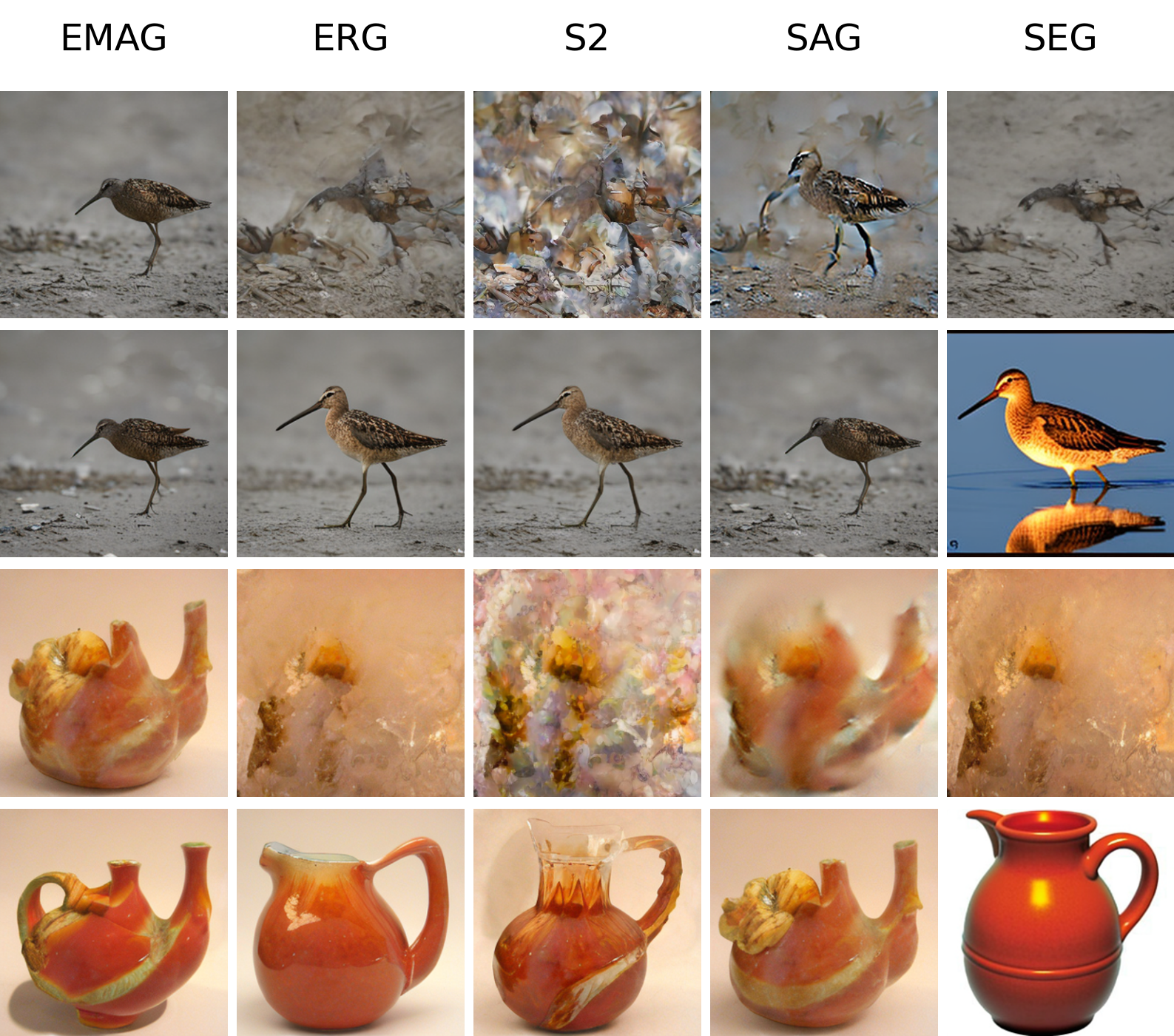}
      \caption{\textbf{Negative-sample comparison (top: negative samples; bottom: positive samples (clean image).} Prior guidance method (SAG \cite{hong2023improving}, auto guidance \cite{karras2024guiding}, ERG \cite{ifriqi2025entropy}, S$^2$-Guidance \cite{chen2025s}) often yield obvious “easy” degradations. In contrast, \textbf{EMAG} produces subtle, semantically near-miss negatives that reveal difficult failure modes yet retain global structure, enabling finer refinement. (Samples from DiT)}
    \label{fig: hard_negatives}
\vspace{-20pt}
\end{wrapfigure}

PAG perturbs self-attention by replacing selected attention maps with the identity, and then guides sampling away from the resulting degraded intermediates; its guidance rule departs from Eq.~\ref{eq:autoguidance} and is given in Eq.~\ref{eq:PAG}, where $\epsilon_1'$ denotes the network with perturbed attention. Similarly, SAG uses self-attention maps from the denoiser $\epsilon_\theta$ (the score prediction network) to detect high-frequency regions in the latent input. These regions are then selectively blurred to obtain a degraded latent $x_t'$, effectively producing a weaker model $\epsilon_1(x_t')$. Substituting this degraded score into Eq. \ref{eq:PAG} in place of perturbed noise $\epsilon_1'(x_t)$ yields the SAG objective.
Alternatively,
\emph{S$^2$-Guidance} constructs the weaker surrogate via stochastic layer dropping and, instead of Eq.~\ref{eq:autoguidance}, improves the CFG update by subtracting the perturbed model’s noise prediction scaled by $s$; equivalently, Eq.~\ref{eq:cfg_eq} is augmented with the term $-\,s \cdot \epsilon_1'(x_t, c, m)$, where $m$ controls which layers are dropped at each timestep, $\epsilon_1'$ is the perturbed sub-network, and $s$ is the S$^2$ scale.

In the auto-guidance framework, a weaker model helps steer sampling toward higher-quality regions\cite{karras2024guiding}. When derived from the same backbone, the weaker model tends to replicate the strong model’s errors but with greater magnitude. This discrepancy highlights failure modes that the strong model can correct. \cite{karras2024guiding}. Following this intuition, we target subtle, fine-grained errors (\emph{hard negatives}) by constructing lightly degraded surrogates that act as guidance, improving fidelity and HPS \cite{wu2023human}. Our approach follows the attention-perturbation family, e.g., ERG, PAG, and SEG, which instantiate such surrogates by modifying self-attention during sampling.

\subsection{Hard Negatives}
\label{sec:hard_negatives_supp}

\paragraph{Why quality of negative samples matters?}

In prior approaches, including auto-guidance \cite{karras2024guiding}, Entropy Rectifying Guidance (ERG) \cite{ifriqi2025entropy}, and S$^2$-Guidance \cite{chen2025s} instantiate a weaker surrogate to surface failure modes complementary to a strong model, thereby guiding sampling toward higher quality. However, these methods primarily specify how to obtain a weaker surrogate (e.g., capacity reduction, attention perturbation, or stochastic layer dropping) rather than offering direct, fine-grained control over which failure modes to elicit and to what extent. In this work, we introduce a strategy that explicitly controls the degradation level while preserving core semantics, yielding “hard negatives”, subtle, fine-grained degradations that expose nuanced errors without altering global structure and enable the model to rectify them see Fig.~\ref{fig: hard_negatives}. In Fig.~ \ref{fig:Negative_sample_controlled}, we observe that while the global structure of the mobile phone remains intact, subtle local deformations (e.g., distorted buttons) emerge; such minimally degraded yet semantically consistent samples constitute the hard negatives considered in this work. As evidenced in Fig.~\ref{fig:front_image} and Table~\ref{tab:combination_experiment}, this controllable hard-negative guidance improves the Human Preference Score (HPS) \cite{wu2023human} over both the baseline and several strong methods. We further quantify negative hardness in Fig.~\ref{fig: section_F_hard_negative_similarity_scores}, which reports the similarity between negative samples and their corresponding generated outputs; higher similarity indicates harder negatives. 

\paragraph{Measuring negative ``hardness'' via cosine similarity.}
We quantify how ``hard'' a negative is by measuring how \emph{close} it is to the final guided output under a semantic image representation.
Let $x^{+}$ denote the final guided image (or decoded final sample), and let $x^{-}$ denote the corresponding negative image produced by the EMA-perturbed branch (using the same condition/seed). 
We compute the hardness score as cosine similarity in a frozen embedding space:
\begin{equation}
\label{eq:hardness_cos}
\mathcal{H}(x^{-}, x^{+}) \;=\;
\cos\big(\phi(x^{-}), \phi(x^{+})\big)
\;=\;
\frac{\langle \phi(x^{-}), \phi(x^{+}) \rangle}{\|\phi(x^{-})\|_2 \, \|\phi(x^{+})\|_2},
\end{equation}
where $\phi(\cdot)$ denotes a pretrained image encoder. We compute cosine similarity using both CLIP image embeddings, which capture semantic similarity, and DINOv2 embeddings, which better reflect structural and perceptual similarity.

Intuitively, larger values of $\mathcal{H}$ indicate \emph{harder} negatives: the negative sample remains semantically close to the final output while differing primarily in fine-grained details. We compare mean similarity scores across baselines in Fig.~\ref{fig: section_F_hard_negative_similarity_scores}, where EMAG consistently produces the hardest negatives under both metrics, with SAG being the closest competitor. Within EMAG, smaller values of $\beta$ yield harder negatives (Fig.~\ref{fig: section_F_hard_negative_similarity_scores_different_beta}), as evidenced by higher similarity scores, particularly under DINOv2~\cite{oquab2023dinov2}, which is sensitive to structural and perceptual similarity. Qualitative examples are provided in
Fig.~S10

\subsection{Exponential Moving Average Guidance}

We adopt a strategy akin to prior works \cite{karras2024guiding, hong2023improving, hong2024smoothed} where we contrast denoising outputs from a strong model versus a deliberately degraded weaker prediction via Eq.~\ref{eq:autoguidance}. Following \cite{hong2024smoothed, ifriqi2025entropy, ahn2024self}, we perturb self‐attention by replacing the attention map at a selected layer and timestep \(t\), \(A_t\), with its exponential moving average \(\mathrm{EMA}(A_{<t})\), using our adaptive layer‐selection strategy (Sec~\ref{sec:layer selection}). The detailed steps are provided in Alg.~\ref{alg:EMAG-cond} and Alg.~2 (supplementary).
Intuitively, since self‐attention refines semantics from coarse to fine, swapping in the EMA suppresses high‐frequency refinements while preserving global structure, yielding a controllable, semantically faithful “hard negative.” We then contrast this negative with the original prediction during sampling. Fig.~\ref{fig: hard_negatives} shows that our EMA‐guided approach degrades selectively without losing core semantic details, and Fig.~\ref{fig:Negative_sample_controlled} and \ref{fig: section_F_hard_negative_similarity_scores_different_beta} demonstrate how adjusting the decay factor \(\beta\) controls the degradation strength. Now we will discuss our approach and its components in detail.

\begin{algorithm}[t]
\caption{EMAG (Conditional)}
\label{alg:EMAG-cond}
\begin{algorithmic}[1]
\REQUIRE $\epsilon_\theta$: denoiser, $\epsilon'_\theta$: perturbed denoiser,
         $\tau_s, \tau_e$: start/end timesteps, $w_e$: EMAG scale,
         $c$: condition, $w_{cfg}$: CFG scale
\STATE $x_T \sim \mathcal{N}(0,I)$
\FOR{$t = T, T-1, \ldots, 1$}
    \STATE $z_t^{c} \leftarrow \epsilon_\theta(x_t,c)$
    \STATE $z_t \leftarrow \epsilon_\theta(x_t)$
    \IF{$\tau_e < t < \tau_s$}
        \STATE $\hat{z}_t^{c} \leftarrow \epsilon'_\theta(x_t,c)$ \COMMENT{Eq.~\eqref{eq: layer_selection}, \eqref{eq: L1_norm_calcuation}, \eqref{eq: EMA replacement}}
        \STATE $\bar{z}_t^{c} \leftarrow \hat{z}_t^{c} + w_{e}\cdot(z_t^{c} - \hat{z}_t^{c})$ \COMMENT{Eq.~\eqref{eq:EMAG_gudiance_update}}
        \STATE $\bar{z}_t \leftarrow z_t + w_{cfg}\cdot(\bar{z}_t^{c} - z_t)$ \COMMENT{Eq.~\eqref{eq:EMAG_gudiance_update_2}}
    \ELSE
        \STATE $\bar{z}_t \leftarrow z_t + w_{cfg}\cdot(z_t^{c} - z_t)$ \COMMENT{Eq.~\eqref{eq:cfg_eq}}
    \ENDIF
\ENDFOR
\RETURN $\bar{z}_t$
\end{algorithmic}
\end{algorithm}

\subsubsection{Attention EMA calculation}
Let $A_t \in [0,1]^{B\times H\times Q\times K}$ denote the softmax-normalized attention at timestep $t$ (rows sum to $1$ over $K$).
We maintain an exponential moving average (EMA) $E_t$ of the same shape as $A_t$, initialized on the first call as $E_1 \!:=\! A_1$, and updated as mentioned in Eq.\ref{eq:EMA update}. 
\begin{equation}
    \label{eq:EMA update}
    E_t \;=\; \beta\, E_{t-1} \;+\; (1-\beta)\, A_t, \qquad \beta=0.988.
\end{equation}

\begin{figure*}[t]
  \centering
  \begin{subfigure}[t]{0.48\linewidth}
    \centering
    \includegraphics[width=\linewidth]{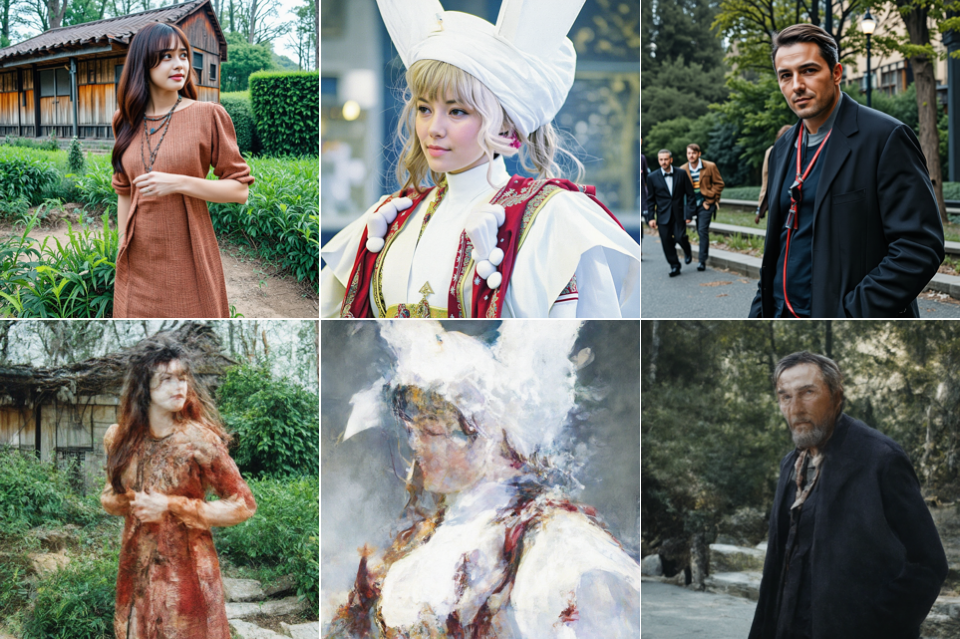}
    \caption{Unconditional generation. \textbf{Top:} EMAG. \textbf{Bottom:} unguided (baseline).}
    \label{fig:uncond-samples}
  \end{subfigure}\hfill
  \begin{subfigure}[t]{0.48\linewidth}
    \centering
    \includegraphics[width=\linewidth]{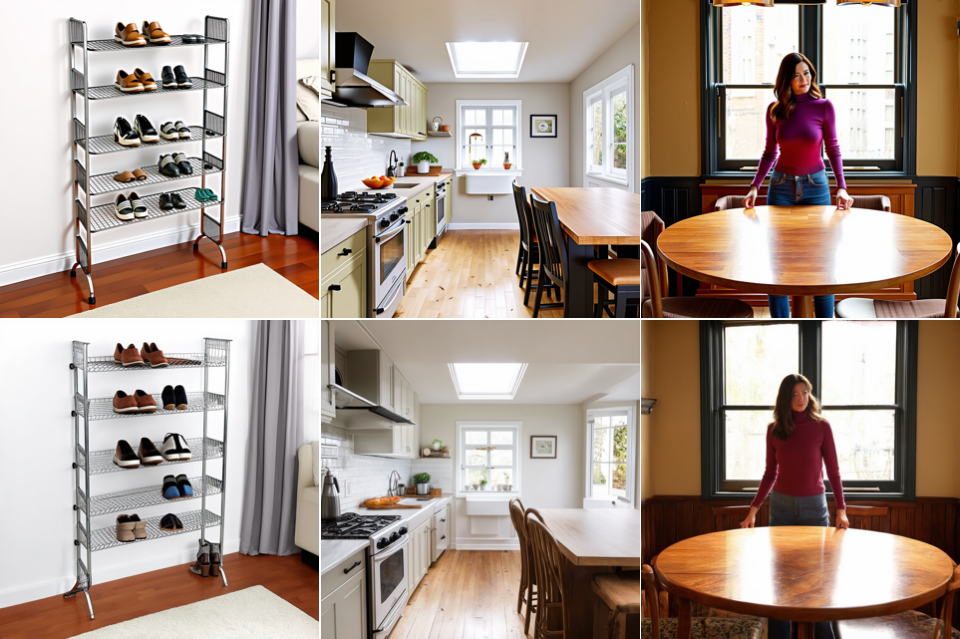}
    \caption{Class-conditional generation. \textbf{Top:} EMAG. \textbf{Bottom:} CFG (baseline).}
    \label{fig:cond-samples}
  \end{subfigure}
  \caption{Qualitative comparison for (a) Unconditional and (b) Text-conditional settings.}
  \label{fig:generation-samples}
\end{figure*}

We set $\beta$ by its EMA half-life $H$ via $\beta = e^{-\ln(2)/H}$: guidance becomes net-positive once the half-life exceeds $\approx\!0.7T$ of the sampling horizon, and our default $\beta{=}0.988$ (half-life $\approx\!2T$ for 28 steps) lies in this saturated regime. The full $\beta$ ablation is provided in Supp.~Tab.~S21. We also introduce parameters  $\delta_t, \tau_s, \tau_e$ where $\delta_t$ indicates the warmup period for the EMA to stabilize before we can start swapping the attention maps with the EMA version, $\tau_s$ indicates the timesteps when we start accumulating EMA and apply the perturbation, and $\tau_e$ indicates the timestep when we don't perturb the attention. Since the denoising happens in reverse, we have $t_0 < \tau_e < \tau_s<t_{max}$.

Replacing the attention map with its EMA effectively suppresses the incremental (high-frequency) refinements the model applies over timesteps while preserving the global structure. This induces inferior but semantically faithful predictions that guide sampling toward higher quality by introducing precise failure modes. In practice, the exponential smoothing prevents drift: even with $\delta_t = 1$, the guidance remains faithful to the original structure, unlike strong perturbation approaches \cite{chen2025s,ifriqi2025entropy} (see Fig.~\ref{fig: hard_negatives}).

\subsubsection{Layer Selection}
\label{sec:layer selection}

As prior work shows \cite{ifriqi2025entropy, hong2023improving, ahn2024self}, middle layers are especially effective for attention perturbations. For instance, \cite{ifriqi2025entropy} demonstrates that middle layers exhibit lower maximum attention probabilities than first or last layers. Methods like \cite{hong2023improving, ahn2024self} also extract masks or apply perturbations in these middle layers of the U-Net. Inspired by this, we apply EMAG specifically to middle layers: for DiT-XL/2 \cite{peebles2023scalable} we set $(l_{\min}, l_{\max}) = (12,15)$, and for SD3-Medium \cite{esser2024scaling} we set $(l_{\min}, l_{\max}) = (6,8)$.

At each timestep $t$ and layers $(l_{min},l_{max})$ we select a layer $l_n \text{ where } n \in (l_{min},l_{max})$ using the Eq.\ref{eq: layer_selection} and Eq. \ref{eq: L1_norm_calcuation} where $l^t_n$ is the selected layer at time step $t$ and $|A_t|$ is the total number of elements in the attention map. For per-layer selection, we compute $\Delta_{t}^{(\ell)}$ on the attention tensor of layer $\ell$ and choose layers with the largest $\Delta_{t}^{(\ell)}$.
\begin{align}
    \label{eq: layer_selection}
    l^t_n = \arg\max_{n \in \{l_{\min}, ..., l_{\max}\}} \Delta_{t}^{(n)}
\\
    \label{eq: L1_norm_calcuation}
    \Delta_{t} \;=\; \operatorname{MAE}(E_t, A_t) = \frac{1}{|A_t|}\sum_{i} \bigl| E_t^{(i)} - A_t^{(i)} \bigr|
\end{align}
Finally, we replace the EMA for the attention layer using a convex blend, see Eq.\ref{eq: EMA replacement} with $\lambda = 1$ (hard replace) as our default setting.  $\lambda \in [0, 1]$ allows for a soft update, and $\widetilde{A}_t$ is the perturbed attention map.
\begin{equation}
    \label{eq: EMA replacement}
    \widetilde{A}_t \;=\; (1-\lambda)\,A_t \;+\; \lambda\,E_t,
\end{equation}
This enables us to select the layer with the largest ~$\Delta_t$ at each timestep~$t$, ensuring maximal perturbation. We maintain EMA buffers for all layers in $(l_{\min},l_{\max})$ and at every timestep replace the attention map of the selected layer $l_{n}^{t}$. Since different layers contribute unequally across timesteps~\cite{ma2024learning}, our statistic-guided selection picks the layer actively inserting high-frequency content, yielding stronger negative samples. This approach also opens the possibility for providing more controlled negative samples by carefully designing optimization for the $\Delta_{t}^{(\ell)}$ as against the greedy approach Eq. \ref{eq: layer_selection}. We defer this to future work.

\subsubsection{EMAG-I}
In MMDiT's joint attention over text and image tokens, our default EMAG perturbs both Image$\rightarrow$Image and Image$\rightarrow$Text sub-blocks. While EMAG-I perturbs only the Image$\rightarrow$Image block. See appendix 
Section E

\subsubsection{EMAG-Q}
Instead of maintaining EMA over post-softmax attention maps, EMAG-Q tracks EMA over query embeddings and perturbs only queries, enabling fused SDPA and substantially reducing overhead. We defer the formulation and approximation analysis to the appendix 
section~G

\subsection{Guidance update}

Once we obtain the perturbed noise estimates, we perform the guidance update in two steps first, we leverage Eq. \ref {eq:autoguidance} for the guidance update and rewrite it as Eq.\ref{eq:EMAG_gudiance_update} presenting the first step of Guidance update for EMAG, where $\epsilon'_\theta(x_t, c)$ is the degraded score estimate and $w_{e}$ is the EMAG scale. Now using this perturbed signal, we perform a CFG-style update but with our perturbed conditional noise, see eq.\ref{eq:EMAG_gudiance_update_2} where $w$ is CFG scale.
\begin{align}
    \label{eq:EMAG_gudiance_update}
    &\tilde{\epsilon}(x_t, c)' = \epsilon'_\theta(x_t, c) + w_{e} \cdot (\epsilon_\theta(x_t, c) - \epsilon'_\theta(x_t, c))
 \\
    \label{eq:EMAG_gudiance_update_2}
    &\tilde{\epsilon}(x_t, c) = \epsilon_\theta(x_t, \varnothing) + w \cdot (\tilde{\epsilon}(x_t, c)' - \epsilon_\theta(x_t, \varnothing))
\end{align}
We can also apply EMAG for unconditional generation as presented in Eq. \ref{eq:EMAG_uncond_update}.
\begin{equation}
    \label{eq:EMAG_uncond_update}
    \tilde{\epsilon}(x_t) = \epsilon'_\theta(x_t , \varnothing) + w_{e} \cdot (\epsilon_\theta(x_t) - \epsilon'_\theta(x_t , \varnothing))
\end{equation}
\subsubsection{Combining EMAG with APG and CADS}
EMAG exhibits complementary behavior when combined with more advanced guidance techniques like APG \cite{sadat2024eliminating}, and CADS \cite{sadat2023cads}, where for the APG we follow the algorithm provided by the author, specifically the eq:\ref{eq:EMAG_gudiance_update} remains the same while APG is applied to the Eq: \ref{eq:EMAG_gudiance_update_2}. In essence, it's similar to applying APG to CFG, but in this case, we use our perturbed conditioning.

For CADS, we simply apply the interpolation between the positive tokens and the Gaussian noise to the encoder hidden states of the network. We apply this interpolation to the conditioning of the perturbed network as well.

\section{Experimentation and results}
\label{sec:Experiment}

\begin{wraptable}{r}{0.50\linewidth}
\vspace{-36pt}
\centering
\small
\setlength{\tabcolsep}{4pt}
\caption{Quantitative SD3 text-to-image results on the COCO 2014 validation set (identical data, sampling, and evaluation). Settings are selected from Pareto optimal scales (Supp Sec.~D). 
Bottom block highlights EMAG combos that surpass the corresponding baseline. For $\pm$\,std over 3 seeds; see Table~S22. $\dagger$ Incompatible with SD3, see supplementary section (E.2)}
\label{tab:main_results_SD3_cond}
\begin{tabular}{lcc}
\toprule
\textbf{Guidance} & \textbf{FID} $\downarrow$ & \textbf{HPS} $\uparrow$ \\
\midrule
(No CFG)        & 23.859 & 21.61 {\scriptsize $\pm$ 0.11} \\
CFG             & 22.877 & 29.22 {\scriptsize $\pm$ 0.05} \\
SAG\,+\,CFG     & 23.039 & 29.28 {\scriptsize $\pm$ 0.04} \\
PAG\,+\,CFG $\dagger$     & 24.066 & 28.73 {\scriptsize $\pm$ 0.04} \\
SEG\,+\,CFG     & 23.453 & 29.25 {\scriptsize $\pm$ 0.04} \\
APG             & 21.816 & 29.45 {\scriptsize $\pm$ 0.02} \\
CADS            & \textbf{18.320} & 28.36 \\
S$^{2}$         & 20.821 & 29.15 {\scriptsize $\pm$ 0.00} \\
ERG             & 23.989 & 29.35 {\scriptsize $\pm$ 0.03} \\
\midrule
\textbf{EMAG-I (ours)}  & 22.154 & 29.56 \\
\textbf{EMAG (ours)}    & 22.890 & \textbf{29.76} {\scriptsize $\pm$ 0.03} \\
\textbf{EMAG-Q (ours)}  & 23.530 & 29.64 \\
\midrule
\textbf{EMAG\,+\,APG}   & 21.819 & \textbf{29.79} {\scriptsize $\pm$ 0.04} \\
\textbf{EMAG\,+\,CADS}  & 19.950 & \textbf{28.86} {\scriptsize $\pm$ 0.02} \\
\textbf{EMAG-Q\,+\,APG}   & 22.480 & 29.77 \\
\textbf{EMAG-Q\,+\,CADS}  & 20.030 & 28.96 \\
\bottomrule
\end{tabular}
\vspace{-36pt}
\end{wraptable}

\subsection{Experimental setup}
\label{sec:experimental_setup}

\paragraph{Models}~
We study Transformer-based diffusion backbones: (i) \textbf{DiT-XL/2} in the \emph{class-conditional} setting at \textbf{256$\times$256} and \textbf{512$\times$512} resolutions \cite{peebles2023scalable}, and (ii) \textbf{MMDiT} via the \textbf{Stable Diffusion~3 Medium} checkpoint for \textbf{1024$\times$1024} \emph{text-to-image} generation \cite{esser2024scaling}. See additional details in Supplementary 
Sec.~A.

\paragraph{Datasets \& tasks}~
We evaluate both \emph{conditional} and \emph{unconditional} generation.
For \emph{class-conditional} experiments with DiT-XL/2 we use ImageNet-1K labels \cite{deng2009imagenet} and evaluate against the \emph{official ADM reference batches/statistics} from the Guided Diffusion repository \cite{dhariwal2021diffusion,deng2009imagenet}.
For \emph{text-conditioned} experiments with SD3 we use captions from the \textbf{COCO-2014} \textbf{40k} validation split \cite{lin2014microsoft}.
Unless stated otherwise, we generate \textbf{50K} samples for class-conditional ImageNet experiments and \textbf{40K} samples for text-to-image experiments (one per caption from the COCO-2014 validation split) with the seed set to 8.

\paragraph{Baselines and compared guidance methods}~
We compare against \emph{standard} conditional generation (no guidance), \emph{classifier-free guidance} (CFG), and recent \emph{attention-based} guidance methods:
\textbf{PAG}~\cite{ahn2024self},
\textbf{SAG}~\cite{hong2023improving},
\textbf{SEG}~\cite{hong2024smoothed},
\textbf{ERG}~\cite{ifriqi2025entropy},
and stochastic block dropping method \textbf{S$^2$} (Stochastic Self-Guidance)~\cite{chen2025s}.
We also evaluate two orthogonal guidance strategies,
\textbf{APG} (Adaptive Projected Guidance)~\cite{sadat2024eliminating} and
\textbf{CADS} (Condition-Annealed Diffusion Sampling)~\cite{sadat2023cads},
and further demonstrate that \emph{combining} APG/CADS with our approach yields additional gains (see Table~\ref{tab:main_results_SD3_cond}).
For methods originally proposed for U-Net backbones, we adapt them to Transformer backbones (DiT/MMDiT) by operating within the self-attention stack (e.g., perturbing queries or attention maps, or inserting attention-space degradations) while preserving the residual/MLP pathways; implementations are based on the strategies adopted in \cite{ifriqi2025entropy}. Specifics are provided in the Supplementary Sec.~E. 

\begin{table}[t]
\small
\setlength{\tabcolsep}{4pt}
\caption{Quantitative results for unconditional generation on the COCO,2014 validation set (SD3) and ImageNet-256 (DiT-XL/2). All methods share identical data, sampling steps, and evaluation code; settings follow author recommendations or are selected via hyperparameter sweep (Supp Sec.~D). \textbf{Baseline methods are reported where compatible with the corresponding backbone and sampling formulation.}$\dagger$ Incompatible with SD3, see supplementary section (E.2)}
\label{tab:merged_uncond_results}
\resizebox{\columnwidth}{!}{%
\begin{tabular}{lccccc}
\toprule
\textbf{Guidance} & \textbf{FID} $\downarrow$ & \textbf{Precision} $\uparrow$ & \textbf{Recall} $\uparrow$ & \textbf{Density} $\uparrow$ & \textbf{Coverage} $\uparrow$ \\
\midrule
\multicolumn{6}{l}{\textit{SD3 (MMDiT)\cite{esser2024scaling}}} \\
\midrule
(No-Guidance)                & 101.94 & 0.283 & \textbf{0.298} & 0.199 & 0.095 \\
SAG                     & 95.38   & 0.339 & 0.098 & 0.248 & 0.046 \\
PAG $\dagger$                    & 111.24   & 0.339 & 0.000 & 0.238 & 0.024 \\
SEG                       & 90.21   & \textbf{0.609} &  0.181 & \textbf{0.674} & 0.090 \\
I-ERG                    & \underline{82.11}   & 0.557 & 0.173 & 0.608 & \underline{0.090} \\
\midrule
\textbf{EMAG (ours)}    & \textbf{74.98} & \underline{0.562} & \underline{0.186} & \underline{0.642} & \textbf{0.112} \\

\midrule
\multicolumn{6}{l}{\textit{DiT-XL/2\cite{peebles2023scalable}} 256x256} \\
\midrule
(No-Guidance)           & 52.76  & 0.431 & \underline{0.664} & 0.392 & 0.554 \\
PAG                & 39.51  & \underline{0.626} & 0.357 & \underline{0.811} & \underline{0.703} \\
SAG                & 40.87  & 0.513 &\textbf{ 0.655} & 0.521 & 0.672 \\
SEG                & \underline{39.17}  & \textbf{0.638 }& 0.340 & \textbf{0.844} & 0.701 \\
\midrule
\textbf{EMAG (ours)}    & \textbf{29.85} & 0.621 & 0.561 & 0.805 & \textbf{0.768} \\
\bottomrule
\end{tabular}
}
\end{table}


\paragraph{Metrics}
We assess \emph{quality} with \textbf{HPS\,v2}~\cite{wu2023human}, \textbf{FID}~\cite{heusel2017gans} and \textbf{Precision}/\textbf{Density}~\cite{naeem2020reliable}; and \emph{diversity} with \textbf{Recall}/\textbf{Coverage}~\cite{naeem2020reliable}. We scale HPS by 100 when reporting. For text–image alignment, we report \textbf{CLIPScore}~\cite{hessel2021clipscore}. In text-to-image (conditional) results, we primarily report \textbf{HPS} and \textbf{FID}. For brevity, we refer to HPS\,v2 as HPS throughout the paper.

\paragraph{Hyperparameter sweeps}~
For each baseline (CFG, SAG, PAG, SEG, ERG, $S^2$, APG, CADS), we perform a grid search over combinations of the CFG scale and the method-specific control scale(s). For SD3, we construct the FID--HPS Pareto frontier for each method. We report configurations chosen from the Pareto-optimal scales of each method’s FID–HPS frontier. The full sweep results are reported in Table~S11, and the corresponding Pareto curves in Figs.~\ref{fig:pareto_frontier_curve} and S8. We apply the same selection protocol to DiT, with details provided in Supplementary Sec.~D.


\begin{table}[t]
\centering
\caption{Wall-clock inference timing on a single NVIDIA A100-SXM4-40GB. All methods use SD3 Medium at $1024{\times}1024$ with 28 steps, batch size 1. Timings are averaged over 45 images after 5 warmup iterations. Overhead is relative to vanilla CFG. FID\,/\,HPS are on the COCO 2014 validation set (same setting as Table~\ref{tab:main_results_SD3_cond}).}
\label{tab:timing_benchmark}
\small
\setlength{\tabcolsep}{4pt}
\renewcommand{\arraystretch}{1.05}
\begin{tabular}{lccccc}
\toprule
\textbf{Method}
& \textbf{s/image}
& \textbf{$\pm$ std}
& \textbf{Overhead}
& \textbf{Peak Mem.}
& \textbf{FID/HPS} \\
\midrule
CFG (baseline)              & 4.04  & 0.001 & --       & 16.89\,GB & 22.88 / 29.22 \\
SAG~\cite{hong2023improving} + CFG  & 7.22  & 0.005 & $+$78.5\%  & 24.85\,GB & 23.04 / 29.28 \\
SEG~\cite{hong2024smoothed} + CFG   & 6.05  & 0.001 & $+$49.6\%  & 18.76\,GB & 23.45 / 29.25 \\
PAG~\cite{ahn2024self} + CFG        & 9.41  & 0.002 & $+$132.7\% & 18.33\,GB & 24.07 / 28.73 \\
\textbf{EMAG(Ours)}  & 12.66 & 0.002 & $+$213.2\% & 29.58\,GB & 22.89 / \textbf{29.76} \\
\textbf{EMAG-Q(Ours)} & 7.86 & 0.005 & $+$94.5\%  & 16.96\,GB & 23.53 / 29.64 \\
\bottomrule
\end{tabular}
\end{table}
\subsection{Results}

\begin{wraptable}{r}{0.5\linewidth}
\vspace{-34pt}
\centering
\small
\setlength{\tabcolsep}{6pt}
\caption{Ablation for adaptive layer selection (5K samples, SD3, COCO 2014 val). All runs share identical sampling steps and evaluation code.}
\label{tab:Adaptive_layer_selection_ablation}
\begin{tabular}{lcc}
\toprule
\textbf{Guidance} & \textbf{FID} $\downarrow$ & \textbf{HPS} $\uparrow$ \\
\midrule
\textbf{EMAG (L 6)}      & \textbf{27.47}  & 29.54  \\
\textbf{EMAG (L 7)}      & 28.94            & 29.51  \\
\textbf{EMAG (L 8)}      & 28.23           & \textbf{29.66} \\
\textbf{EMAG (L all)}    & 35.60          & 28.80  \\
\midrule
\textbf{EMAG (Adaptive)} & 28.52          & 29.60  \\
\bottomrule
\end{tabular}
\vspace{-10pt}
\end{wraptable}

\label{Results}

\paragraph{Conditional generation}
We compare EMAG against conditional baselines, including both unguided and CFG-based sampling, as well as recent training-free guidance methods for text-to-image generation, including $S^{2}$-Guidance and ERG. Quantitative results in Table~\ref{tab:main_results_SD3_cond} show that EMAG achieves the highest HPS while remaining competitive on FID. 
We further evaluate compatibility with advanced guidance strategies such as APG and CADS, and find that it consistently yields additional gains in HPS without sacrificing their benefits on FID. Additional PRDC results in Supplementary Table~S1.

We also compare EMAG with guidance methods that are complementary to CFG and designed to boost its performance (Table~\ref{tab:combination_experiment}). EMAG delivers the largest HPS gain over CFG alone, attributable to EMA\mbox{-}induced hard negatives that encourage subtle refinements. It also serves as a complementary plug\mbox{-}in to strong guidance methods such as $S^{2}$ and ERG, further improving HPS. Qualitative examples are presented in Fig.~\ref{fig:generation-samples} and Supplementary Sec.~F. Tables~S8 and~S9 present the results for class-conditioned DiT experiments.

We additionally compare standalone EMAG against other baselines in Table~S17 and evaluate its behaviour under reduced inference budgets in Table~S20.

\begin{figure*}[t]
    \includegraphics[width=\textwidth]{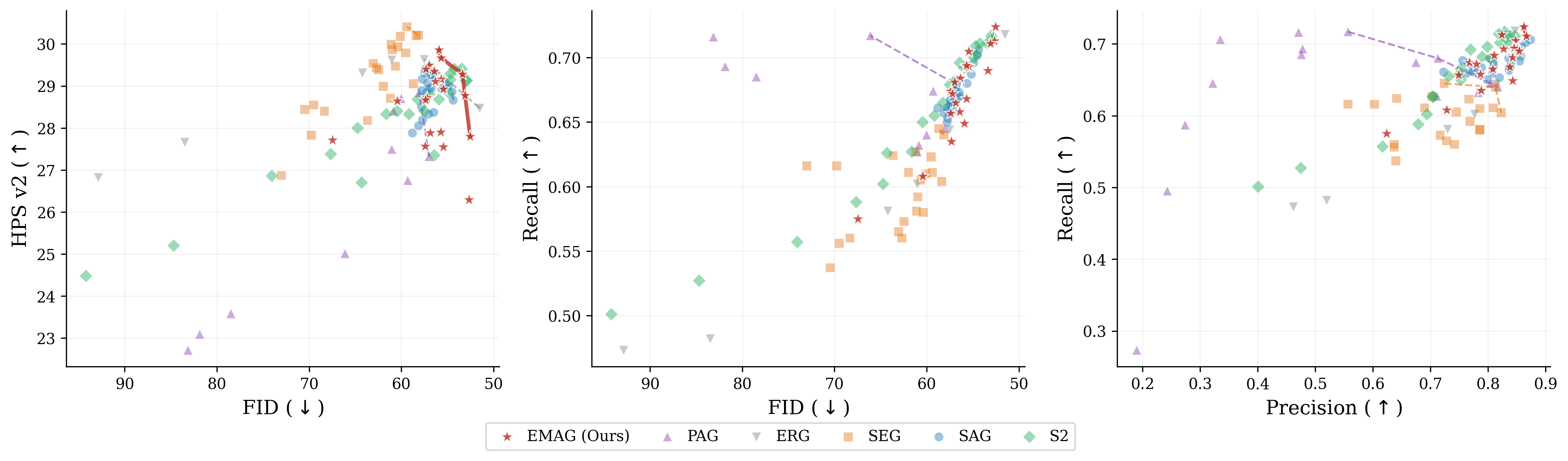}
    \caption{ \textbf{Pairwise Pareto frontiers across guidance methods on MS-COCO 2014 (1K samples, SD3).} Each point represents a specific guidance configuration; axes are oriented so that the \emph{top-right corner is optimal} in all three panels. \textbf{(Left)} FID vs.\ HPS: the primary quality--preference trade-off. EMAG (\textcolor{red}{$\bigstar$}) consistently occupies the top-right region, achieving a high HPS -FID tradeoff. \textbf{(Middle)} FID vs.\ Recall: fidelity--diversity trade-off. EMAG maintains competitive recall across the full FID range. \textbf{(Right)} Precision vs.\ Recall: quality--diversity trade-off. EMAG achieves a balanced operating point, avoiding the precision--recall collapse. Both method-specific and CFG scales are swept; full numerical results in Tab.~S11.} 
    \label{fig:pareto_frontier_curve}
\end{figure*}

We further note that, across methods, negative hardness alone does not determine final quality, and the perturbation mechanism also matters We further note that, across methods, negative hardness alone does not determine final quality---the perturbation mechanism also matters (e.g., $S^2$ and ERG show comparable hardness in Fig.~\ref{fig: section_F_hard_negative_similarity_scores} yet differ on the Pareto frontier in Fig.~\ref{fig:pareto_frontier_curve}).

\paragraph{Unconditional generation}~
We evaluate EMAG against a no\mbox{-}guidance baseline and representative guidance methods applicable in this setting, such as SAG, SEG (adapted to SD3), and I\mbox{-}ERG, and additionally report results for DiT\mbox{-}XL/2. Quantitative results in Table~\ref{tab:merged_uncond_results} show that EMAG attains the lowest FID and the highest Coverage across both settings, while remaining competitive on the remaining metrics. Qualitative comparisons in Fig.~\ref{fig:generation-samples}(a) show that, relative to the no-guidance baseline, EMAG preserves the core structure while refining fine details through EMA-induced hard negatives.

A practical limitation of standard EMAG is its computational overhead: maintaining EMA over post-softmax attention maps prevents the use of fused SDPA kernels and increases both runtime and memory (Table~\ref{tab:timing_benchmark}). To address this, we introduce \textbf{EMAG-Q}. It reduces inference overhead from $+213\%$ to $+95\%$ and peak memory from 29.58\,GB to 16.96\,GB, while preserving generation quality with only a $+0.63$ increase in FID and a $-0.12$ drop in HPS relative to EMAG (Table~S16). Further discussion is provided in Appendix Sec.~G.

\subsection{Ablation results}
\label{sec:Result_ablation}
\paragraph{Adaptive layer selection}

Table~\ref{tab:Adaptive_layer_selection_ablation} ablates the adaptive layer-selection mechanism against fixed single-layer and all-layers alternatives. Fixed single-layer variants may improve one metric but not both, while the all-layers strategy increases overhead and degrades performance. Our adaptive selection achieves the best FID–HPS balance.

\textbf{Guidance scale selection}~
Fig.~S3 shows how FID and HPS vary with CFG and EMAG scales. Unless noted otherwise (e.g., in ablations), we use a CFG scale of \textbf{7} and an EMAG scale of \textbf{1.5} for conditional and \textbf{5.125} for unconditional. A detailed Pareto frontier analysis is provided in Supplementary Section~D.5.

\section{Conclusion}
\label{sec:Conclusion}

We introduce \emph{Exponential Moving Average Guidance (EMAG)}, a training\mbox{-}free method that uses temporal attention smoothing to construct semantically faithful hard negatives for fine-grained refinement during sampling. EMAG improves HPS across conditional settings while remaining competitive on fidelity and diversity. It composes effectively with orthogonal techniques such as APG and CADS, as well as negative-signal guidance methods such as $S^{2}$.

\bibliographystyle{splncs04}
\bibliography{main}

\setcounter{figure}{0}
\setcounter{table}{0}
\renewcommand{\thefigure}{S\arabic{figure}}
\renewcommand{\thetable}{S\arabic{table}}

\appendix    

\section{Additional Experimental Setup Details and Discussion}
\label{sec:experimentl_setup_supp}


\paragraph{Samplers and guidance windows}
For \textbf{DiT-XL/2}, we use \textbf{DDIM} with \textbf{250} steps (deterministic, $\eta{=}0$), indexing timesteps as $t\in\{0,\ldots,t_{\max}\}$ with $t_{\max}{=}250$ and $t_0{=}0$ \cite{song2020denoising}.
For \textbf{SD3 Medium (MMDiT)}, we use the \textbf{FlowMatchEulerDiscrete} sampler with \textbf{28} steps \cite{lipman2022flow}.
Unless noted otherwise, we use a late-step (``tail'') schedule where EMAG is applied while timesteps traverse $(\tau_{s} \rightarrow \tau_{e})$ (high-to-low noise), with $(\tau_{s} = t_{\max})$ and $(\tau_{e} = \lfloor 0.2\, t_{\max} \rfloor)$. Concretely, this gives $(\tau_{e} = 5)$ for SD3 with 28 steps and $(\tau_{e} = 50)$ for DiT with DDIM-250; EMAG is disabled for $(t < \tau_{e})$.

\paragraph{Evaluation tooling.}
All distributional metrics (FID, Precision, Recall) are computed using the Guided Diffusion evaluation toolkit~\cite{dhariwal2021diffusion} with ImageNet reference statistics, while Density and Coverage follow the official PRDC implementation~\cite{naeem2020reliable}. Human preference scores use HPSv2~\cite{wu2023human}. Where applicable, we unify these under the EvalGIM library~\cite{hall2024evalgimlibraryevaluatinggenerative} to ensure reproducible pipelines across all experiments.

\paragraph{Multi-Seed Evaluation Protocol}
To assess sensitivity to random initialisation, we repeat all SD3 text-conditional experiments across three seeds (8, 56, 64) with 1{,}000 samples per seed. Tab.~\ref{tab:sd3_cond_std} reports the between-seed standard deviation for each metric. The main results in Tab.\ref{tab:main_results_SD3_cond} use 40{,}000 samples with a single seed; the reduced 1{,}000-sample runs here serve solely to estimate cross-seed variance at a practical compute budget while preserving the same prompts, sampler, and guidance configurations.


\subsection{Scaling with Model Capacity}
\label{sec:scaling_model_capacity}

The motivation for using EMA rather than injecting Gaussian noise is precision: EMA produces temporally coherent perturbations that preserve global semantic structure while selectively attenuating fine-grained,
step-to-step refinements. 
This exposes \emph{hard} failure modes( subtle structural degradations) that remain challenging even for high-capacity models, as opposed to the \emph{easy} negatives produced by random or spatially coarse perturbations, which stronger models can trivially self-correct.

\begin{figure*}[t]
  \centering
  \includegraphics[width=\linewidth]{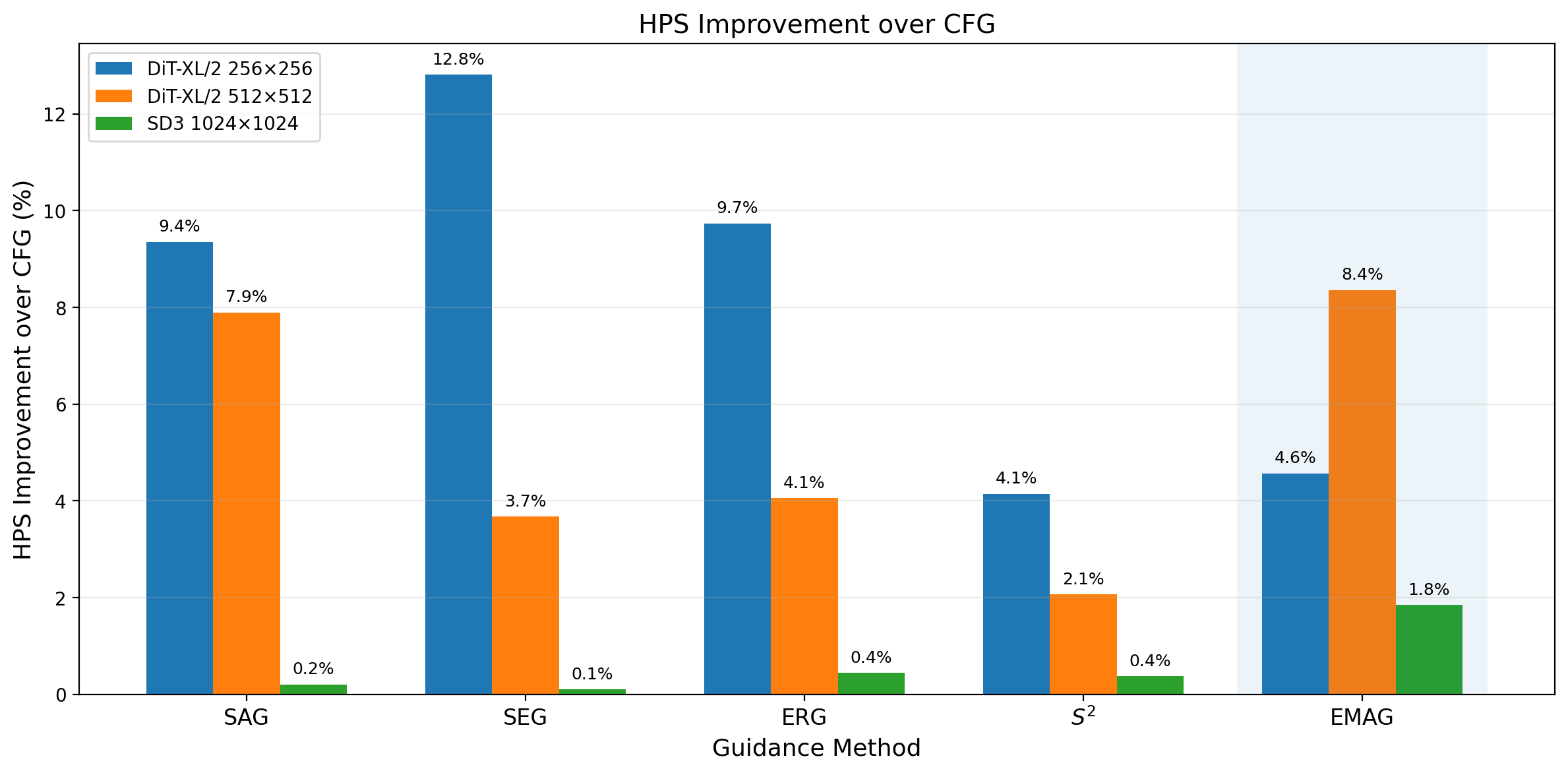}
    \caption{Relative HPS improvement (\%) over the CFG-only baseline across three model configurations: class-conditional DiT-XL/2 at 256$\times$256 and 512$\times$512 (ImageNet), and text-conditional SD3 at 1024$\times$1024 (COCO 2014). While existing guidance methods (SAG, SEG, ERG, S$^{2}$) achieve large gains on the lower-capacity DiT-256, their improvements diminishes with increasing model capacity and resolution. EMAG  maintains a consistent positive gain across all three settings (+4.6\%, +8.4\%, +1.8\%), supporting the hypothesis that high-capacity models require temporally coherent hard negatives rather than coarse spatial perturbations for effective guidance. Each method uses its best method-specific scales at fixed CFG; CFG scale is fixed at 1.5 for DiT and 7.0 for SD3.}
    \label{fig:hps_scaling_across_models}
\end{figure*}

We observe this empirically under a controlled fixed-CFG setting (CFG${=}1.5$ for DiT, CFG${=}7.0$ for SD3), where each method uses its best method-specific scale (Fig.~\ref{fig:hps_scaling_across_models}). Existing guidance methods achieve substantial HPS gains over CFG on the lower-capacity DiT-256 (SAG: $+9.4\%$, SEG: $+12.8\%$, ERG: $+9.7\%$), but their improvements diminish sharply as model capacity and resolution increase, with several methods providing near-zero benefit on SD3 (SAG: $+0.2\%$, SEG: $+0.1\%$, ERG: $+0.4\%$). In contrast, EMAG is the only method that maintains a consistent positive improvement across all three settings ($+4.6\%$, $+8.4\%$, $+1.8\%$), and notably \emph{increases} from DiT-256 to DiT-512. This supports the hypothesis that high-capacity models can trivially self-correct coarse spatial perturbations like as zeroed attention maps or smoothed eigenvectors---reducing their effectiveness, whereas EMAG's temporally coherent hard negatives expose subtler failure modes that persist even in stronger models.

\section{Extended Related Work}
\label{sec: Extended Related Work}

Another inference-time attention modification is PLADIS~\cite{kim2025pladis}, which improves conditional generation by leveraging \emph{sparse cross-attention}. 
PLADIS extrapolates query--key correlations by combining standard softmax attention with a sparse counterpart inside cross-attention layers, boosting text alignment and human preference without extra training ~\cite{kim2025pladis}. 

EMAG differs from PLADIS along two fundamental axes.
First, the two methods operate on \emph{opposite signal directions}: PLADIS enhances the \emph{positive} conditional signal by sharpening cross-attention to the text prompt, whereas EMAG constructs a deliberately weakened \emph{negative} prediction via temporal attention smoothing and guides \emph{away} from it.
Second, the perturbation mechanisms differ fundamentally: PLADIS modifies attention \emph{spatially} within a single forward pass by extrapolating between dense and sparse query--key correlations, while EMAG perturbs attention \emph{temporally} by replacing current attention maps (across both self-attention and cross-attention blocks) with EMA-smoothed maps accumulated from previous timesteps.

Importantly, PLADIS and EMAG are \emph{complementary}: one refines semantic alignment via cross-attention sharpening, while the other improves fine-grained quality by exposing hard negatives via self-attention smoothing.
As noted by~\cite{kim2025pladis}, PLADIS composes with guidance approaches such as CFG, PAG, and SEG; analogously, EMAG could be combined with PLADIS to simultaneously improve both text fidelity and structural detail. Exploring PLADIS combination with EMAG is a natural extension that we defer to future work.

\section{Background}
\label{sec:background}

\subsubsection{Theoretical Interpretation via Hopfield Networks}
\label{sec:hopfiled_view}
Hopfield Networks \cite{hopfield1982neural} are memory models to connect an input with similar patterns by building an energy landscape with the basins of attraction representing the desired patterns. Recently \cite{ramsauer2020hopfield} introduced Modern Hopfield networks with a new energy function extending Hopfield networks to continuous states Eq. \ref{eq:hopfield_1} \cite{ramsauer2020hopfield}, and for $ 0 \leq E \leq 2M^2 $ and using $p=softmax(\beta X^T \zeta)$ define a update rule Eq. \ref{eq:energy_update_rule} where $N$ is the count of stored patterns (keys) $x_i \in \mathcal{R^d}$ and $X = {x_1,...,x_N}$, $M=max_i||x_i||$ , state pattern (query) $\zeta \in \mathcal(R^d)$ and lse is \textit{log-sum-exp function} for $0 < \beta$ for more detailed view refer to \cite{ramsauer2020hopfield}.
\begin{equation}
\label{eq:hopfield_1}
    E = -lse(\beta,X^T\zeta) + \frac{1}{2}\zeta^T\zeta + \beta^{-1}logN + \frac{1}{2}M^{2} 
\end{equation}
\begin{equation}
    \label{eq:energy_update_rule}
    \zeta^{new} = X\,\operatorname{softmax}\,(\beta X^T \zeta) \quad \text{\cite{ramsauer2020hopfield}}
\end{equation}
From Eq.\ref{eq:energy_update_rule} we observe that the attention mechanism is equivalent to the update rule in modern hopfield network. 

Crucially, Ramsauer et\,al.\ prove that (i) the dynamics in~\eqref{eq:energy_update_rule} monotonically decrease the energy and converge to a stationary point (\textbf{Thm.\,1 and Thm.\,2: global convergence}); (ii) the model admits \emph{exponential} storage capacity in dimension $d$ (\textbf{Thm.\,3}); and (iii) retrieval can succeed after a \emph{single} update (\textbf{Thm.\,4 and Thm.\,5: one-step retrieval and small retrieval error })~\cite{ramsauer2020hopfield}. In the context of diffusion transformers, repeated attention applications across timesteps therefore act as successive energy-descent refinements of a state, empirically evolving attention from coarse to fine structure. 

Building on the intuitions from the Continuous Hopfield Networks \cite{ramsauer2020hopfield}  model of Self-Attention, see Eq.\ref{eq:energy_update_rule} obtained from \cite{ramsauer2020hopfield}., which suggests that for $t \rightarrow\infty$ the energy $E(\zeta^t) \rightarrow E(\zeta^*)$ and any sequence $\{\zeta^t\}^\infty_{t=0}$ generated from Eq.\ref{eq:energy_update_rule} will converge to a fixed point $\zeta^*$ where $\zeta \in \mathbb{R}^d $ is the state query in our case the query vector in self attention. Hence, from this observation, we can say that the self attention converges from coarse to finer detail with each time step or application of the update rule Eq.\ref{eq:energy_update_rule}. 
This Hopfield perspective offers an intuitive account of why EMA operates effectively in attention-space: as a temporal low-pass operator, EMA selectively attenuates the high-frequency refinements accumulated across recent denoising steps while preserving global semantic structure, yielding subtle, semantics-preserving degradations that constitute \emph{hard negatives} for guidance

\begin{figure}[t]
  \centering
  \includegraphics[width=\linewidth]{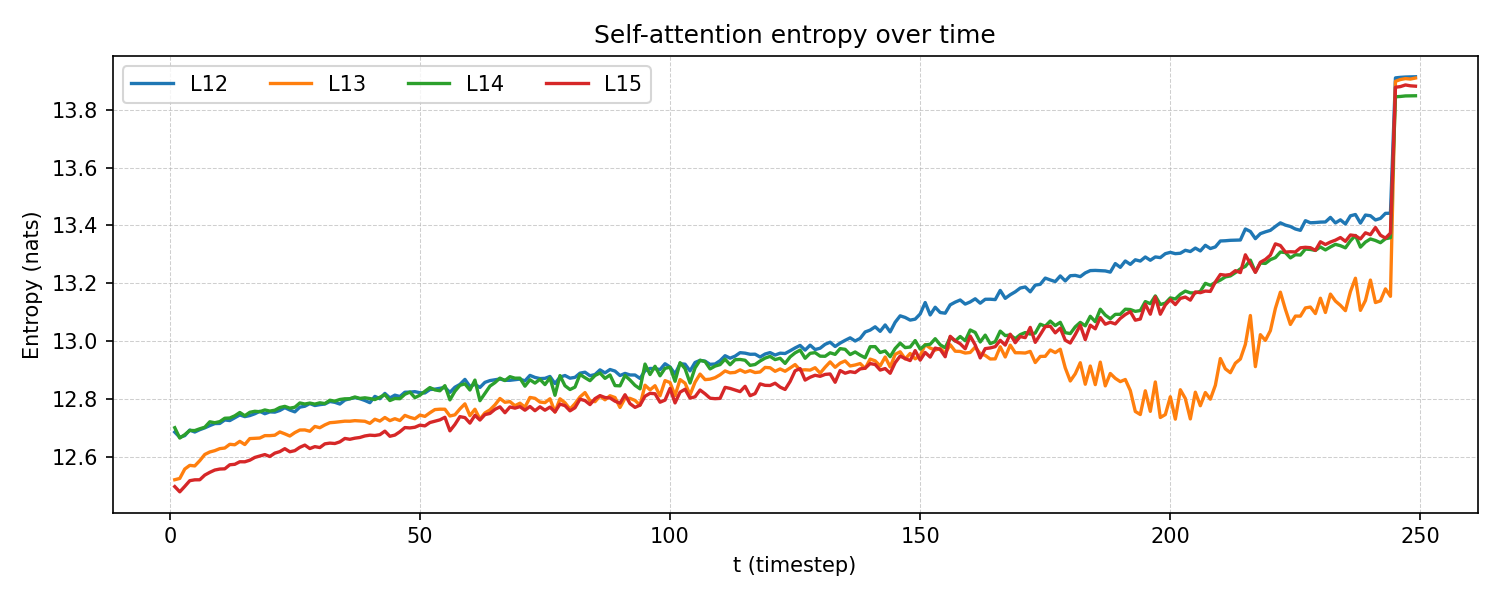}
     \caption{\textbf{Self-attention entropy over time.} Average token-wise entropy (in nats) of the self-attention maps in layers L12–L15 along the reverse diffusion trajectory of DiT-XL/2 with 250-step DDIM sampling.}
    \label{fig: section_a_entropy_delta_all_layers}
\end{figure}

\begin{figure}[t]
  \centering
  \includegraphics[width=\linewidth]{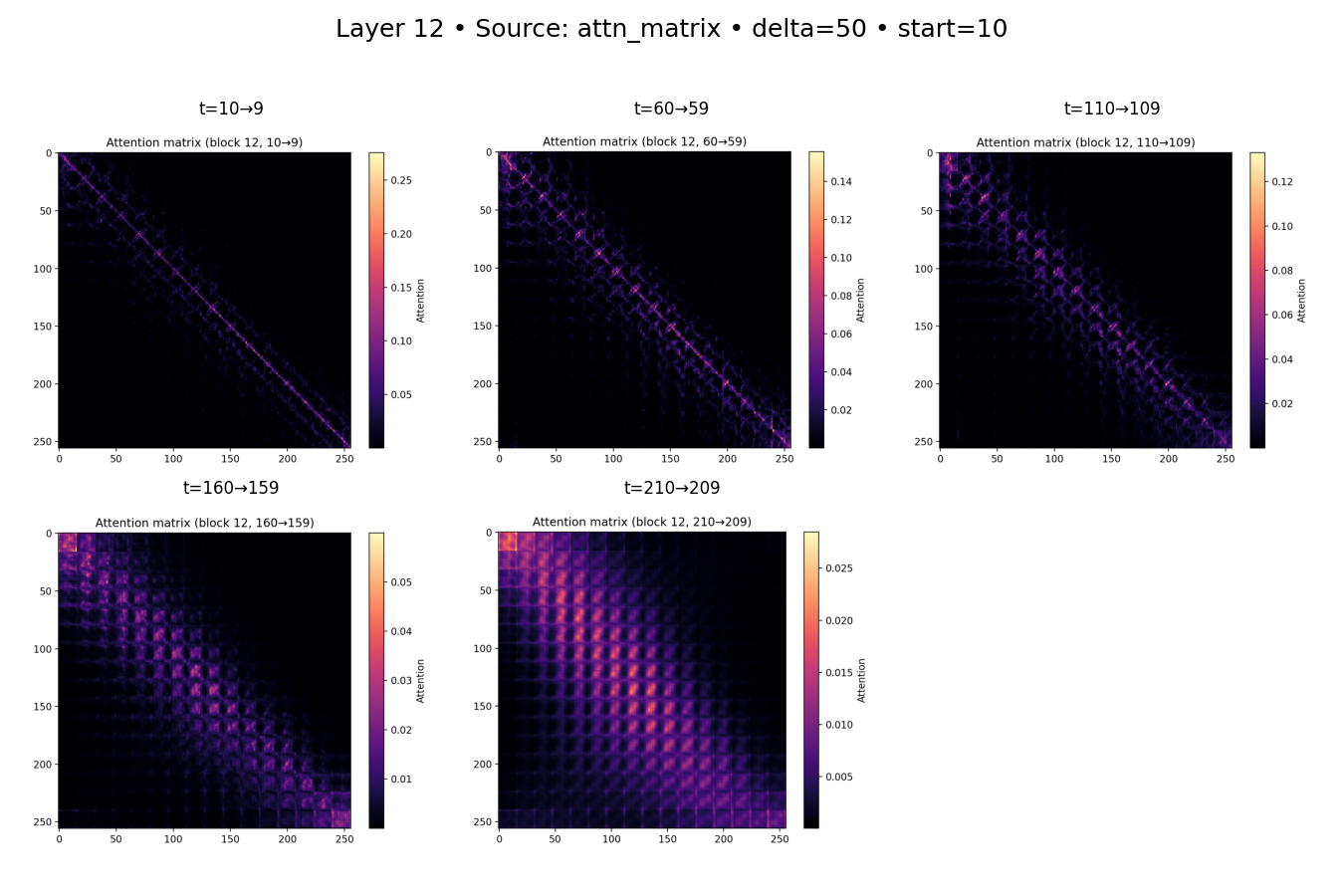}
     \caption{\textbf{Self-attention maps across timesteps.} Visualization of self-attention in layer \textbf{L12} along the reverse diffusion trajectory of DiT-XL/2 with 250-step DDIM sampling.}
    \label{fig: section_a_entropy_layer_wise_pattern}
\end{figure}

As discussed above, Eq.~\ref{eq:energy_update_rule} gives the update rule for modern Hopfield networks~\cite{ramsauer2020hopfield}. From Theorems~1 and~2 in~\cite{ramsauer2020hopfield}, for fixed patterns $X$, the iterations $\{\zeta^t\}_{t=0}^\infty$ generated by Eq.~\ref{eq:energy_update_rule} monotonically decrease the energy $E(\zeta^t)$ and converge to a fixed point $\zeta^*$, i.e., $E(\zeta^t) \rightarrow E(\zeta^*)$ as $t \rightarrow \infty$, where $\zeta \in \mathbb{R}^d$ plays the role of a query vector in self-attention. In our diffusion transformer, keys/values in a given self-attention layer $L_n$ at each reverse step $T_t \rightarrow T_0$ play the role of stored patterns, while token embeddings act as queries. Empirically, for a fixed layer we observe that attention patterns become more concentrated and temporally stable over the reverse process, and often settle into a small number of recurring association structures (see Fig.~\ref{fig: section_a_entropy_layer_wise_pattern}). We also observe that, as the timesteps progress from $T_t \rightarrow T_0$ in the reverse process, the average token-wise entropy of the attention maps tends to decrease (Fig.~\ref{fig: section_a_entropy_delta_all_layers}). This behaviour is qualitatively reminiscent of trajectories moving into attractor basins in a Hopfield energy landscape, although the effective “memory” and energy function here are time-dependent and influenced by stochastic noise, unlike the autonomous dynamics assumed in~\cite{ramsauer2020hopfield}. We therefore use the Hopfield perspective as an interpretive analogy for attention evolution across timesteps.

Theorem~4 of~\cite{ramsauer2020hopfield} further states that, under sufficient pattern separation $\Delta_i$, a single Hopfield update can reduce the distance $d_{\text{hopfield}}(f(\zeta), x^*)$ between the updated state $f(\zeta)$ and a stored pattern $x^*$ to an exponentially small value in the separation parameter. In our diffusion setting, however, the “patterns’’ (final images or latents) lie on a highly entangled, continuous data manifold rather than being well-separated discrete memories, so the effective separation $\Delta_i$ is small and one-step retrieval is neither realistic nor desirable. Instead, diffusion and flow-matching models are explicitly formulated as multi-step denoising / integration procedures that gradually transform noise into data by following a time-varying vector field or score function~\cite{song2020score,ho2020denoising,lipman2022flow}. This iterative refinement can be viewed as repeatedly applying Hopfield-like associative updates while slowly lowering an implicit energy, moving through intermediate latent states that become progressively closer to the target basin.

Within this viewpoint, EMAG can be interpreted as a controlled perturbation of the attention-induced energy landscape. Rather than perturbing the attention landscape at timestep $t$, by manipulating the temperature or replacing with channel-wise mean or an identity, as in prior approaches~\cite{ifriqi2025entropy,ahn2024self,hong2024smoothed}, EMAG replaces them with an exponential moving average over past steps, which we denote schematically as $\widetilde{A}_t \;=\; (1-\lambda)\,A_t \;+\; \lambda\,E_t$ (eq:\ref{eq: EMA replacement}) where we keep $\lambda=1$. Because $\widetilde{A}_t$ is a temporally smoothed version of $A_t$, it largely preserves the core semantic structure of the current attractor (which regions attend to which), but lags behind the most recent, highly sharpened configuration. Replacing $A_t$ by $\bar{A}_t$ therefore nudges the current state slightly away from the bottom of its local basin by a small energy increment $\Delta E$: the perturbation is strong enough to move the trajectory off its current path, yet too weak to collapse the global layout or object identities. In Hopfield terms, EMAG creates hard negatives that remain close to the attractor, subtly degraded versions of the same pattern rather than arbitrary corruptions. The reverse process is then forced to re-relax toward low-energy configurations from these slightly displaced starting points, encouraging the model to explore nearby trajectories in the energy landscape and correcting fine-grained artefacts (e.g., local texture and geometry) while keeping the overall structure intact.

\section{Additional experiments and ablation results}
\label{sec:additional_experimetns_and_ablation}

\subsection{Additional results for SD3 experiments}

\begin{figure}[t]
  \centering
  \begin{subfigure}[t]{0.5\linewidth}
    \centering
    \includegraphics[width=\linewidth]{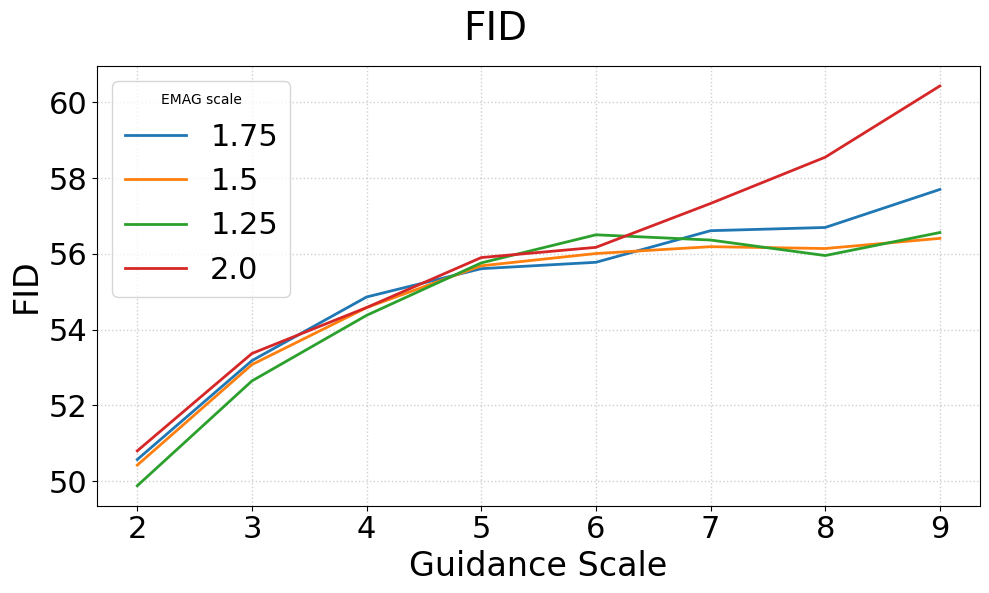}
    \caption{}
  \end{subfigure}\hfill
  \begin{subfigure}[t]{0.5\linewidth}
    \centering
    \includegraphics[width=\linewidth]{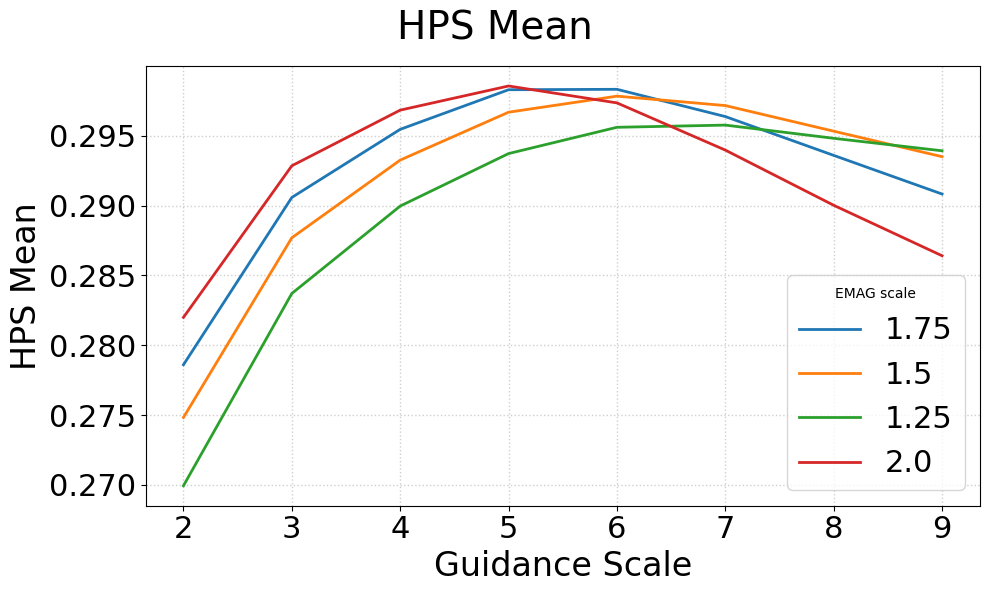}
    \caption{}
  \end{subfigure}
  \caption{HPS and FID vs guidance scale (1000 samples). Colors denote EMAG scales {1.25, 1.5, 1.75, 2.0}. (a) HPS ↑. (b) FID ↓. Same split, steps, and evaluation as main experiments.}
  \label{fig:guidance}
\end{figure}

In Table~\ref{tab:main_results_SD3_cond_supp}, we show that EMAG improves generation quality and human preference, as measured by HPS, while inducing only minimal changes in FID and other distributional metrics. When combined with CFG, EMAG consistently boosts HPS with minimal degradation in FID, indicating that its fine-grained perturbations largely preserve the behavior of the base guidance strategy. A similar pattern holds when EMAG is composed with stronger guidance baselines such as APG and CADS, as well as other SOTA guidance strategies like $S^2$.
Among these, CADS exhibits a slightly larger drop on the other metrics; this is expected, since its early-timestep weakening of the conditional signal delays the formation of a stable core structure in the self-attention landscape, which in turn reduces the relative benefit of EMAG’s fine-grained perturbations at early timesteps. For fairness, we keep EMAG’s hyperparameters fixed across methods; in practice, increasing the temporal-window parameters $\delta_t$ and $\tau_s$ can partially mitigate this effect, and we leave a more systematic study of this setting to future work.


Table~\ref{tab:sd3_baseline_tradeoff} compares EMAG to different attention-perturbation baselines for Pareto optimal scales. As the guidance scale is increased, other baselines shows the expected trade-off, with noticeable FID degradation in exchange for HPS gains. However, in our SD3 conditional setting, even at its optimal operating point, ERG and SAG still yield lower HPS than EMAG, suggesting that EMAG's noise-level perturbations produce a guidance signal more naturally aligned with human preference under comparable conditions. 

\begin{table*}[t]
\centering
\small
\setlength{\tabcolsep}{6pt}
\caption{Quantitative SD3 text-to-image results on COCO 2014 validation set. All methods share identical prompts, sampling steps, and evaluation code. Settings are selected from Pareto optimal scales (Sec.~\ref{sec:additional_experimetns_and_ablation}). The bottom block highlights cases where the combination surpasses the corresponding SOTA baseline. For $\pm$\,std over 3 seeds; see Table.~\ref{tab:sd3_cond_std}. $\dagger$ Incompatible with SD3, see supplementary section (\ref{sec:pag_implementation}).}
\label{tab:main_results_SD3_cond_supp}
\resizebox{\textwidth}{!}{%
\begin{tabular}{lccccccc}
\toprule
\textbf{Model / Guidance} & \textbf{FID} $\downarrow$ & \textbf{CLIP Score} $\uparrow$ & \textbf{Precision} $\uparrow$ & \textbf{Recall} $\uparrow$ & \textbf{Density} $\uparrow$ & \textbf{Coverage} $\uparrow$ & \textbf{HPS} $\uparrow$\\
\midrule
\multicolumn{6}{l}{\textit{SD3 / SD3.5 (MMDiT)}} \\
\midrule
(No CFG) & 23.86 & 23.79 & 0.494 & \textbf{0.371} & 0.591 & 0.505 & 21.61  \\
CFG      & 22.88 & 26.61 & 0.694 & 0.277 & 1.176 & 0.615 & 29.22  \\
SAG + CFG     & 23.04 & 26.71 & 0.679 & 0.265 & 1.121 & 0.597 & 29.28  \\
PAG + CFG $\dagger$ & 24.07  & 26.50 & 0.643 & 0.244 & 1.019 & 0.550 & 28.73  \\
SEG + CFG                    & 23.45  & 25.45 & 0.656 & 0.206 & 1.012 & 0.578 &  29.25  \\
APG   & 21.82  & 26.59 & \textbf{0.703} & 0.295 & \textbf{1.235} & 0.624 & 29.45  \\
CADS & \textbf{18.32} & 26.39 & 0.699 & 0.316 & 1.219 & \textbf{0.644} & 28.36  \\
S2 & 20.82  & \textbf{27.52} & 0.695 & 0.273 & 1.210 & 0.617 & 29.15 \\
ERG & 23.99  & 25.71 & 0.660 & 0.209 & 1.025 & 0.580 & 29.35 \\
\midrule
\textbf{EMAG-I (ours)}    & 22.15  & 26.63 & 0.663 & 0.263 & 1.082 & 0.595 & 29.56  \\  
\textbf{EMAG (ours)}  & 22.89  & 26.66 & 0.669 & 0.260 &  1.097 &  0.591 & \textbf{29.76} \\
\midrule
\textbf{EMAG + APG (ours)}    & 21.82  & 26.55 & 0.685 & 0.258 & 1.174 & 0.613 & \textbf{29.79}  \\  
\textbf{EMAG + CADS (ours)}    & 19.95  & 26.27 & 0.619 & 0.270 & 0.950 & 0.577 & \textbf{28.86}  \\  
\textbf{SD3 + S2 + EMAG}     & \textbf{20.12}  & 27.16  & 0.636 & 0.248 & 1.006   & 0.583  & \textbf{29.51}  \\  
\bottomrule
\end{tabular}
}
\end{table*}

\paragraph{EMAG performance under reduced inference steps}
Table~\ref{tab:emag_fewstep_ablation} shows that EMAG with default $\beta{=}0.988$ delivers consistent HPS gains across all step counts from 28 down to 4, even though the EMA half-life (${\approx}57$ steps) substantially exceeds the total sampling budget at low step counts. At $T{=}8$, EMAG improves both HPS (+1.15) and FID (58.49$\to$56.73) over CFG-only, indicating that even a partially warmed-up EMA provides a useful guidance signal under constrained budgets. Adapting $\beta$ to match the step count (Block~2) offers no clear advantage over the default, and in some cases degrades performance (e.g., $\beta{=}0.75$ at $T{=}14$ yields $\Delta$HPS$=-$0.20). This confirms that EMAG's default configuration generalizes across step regimes without requiring schedule-specific tuning.

\paragraph{Standalone alone gudiance comparason for baselines with EMAG}~

Several trends emerge from Table~\ref{tab:standalone_guidance_comparison}. It compares all methods operating as standalone guidance signals without CFG. For EMAG we just use Eq.~\ref{eq:EMAG_gudiance_update} 

\begin{itemize}
    \item EMAG standalone achieves the lowest, FID of any method, including CFG-only, demonstrating that temporal hard negatives provide a stronger distributional alignment signal than classifier-free guidance alone.
    \item EMAG exhibits the flattest trade-off curve as the FID remains stable across a wide range of guidance weights while HPS steadily improves, whereas SEG and SAG show rapid FID degradation at scales that improve perceptual quality
    \item PAG standalone performs poorly across all metrics, with FID nearly double that of other methods and HPS barely exceeding the no-guidance baseline, further supporting the architectural incompatibility discussed in Sec.~\ref{sec:pag_implementation}. 
\end{itemize}

Finally, combining standalone methods with CFG generally improves HPS at the cost of distributional fidelity, though at higher standalone scales, EMAG already captures much of the perceptual quality signal that CFG provides, reaching HPS competitive with CFG-only before any combination. Notably, EMAG also preserves the highest recall among all methods, indicating that its guidance does not sacrifice sample diversity to achieve quality gains.

\subsection{Additional experimental results on DiT-XL/2}
Since HPS requires a prompt–image pair, for class-conditioned experiments on ImageNet we follow the template prompt \textit{``A photo of \{class label\}.''}.

For DiT-XL/2 experiments we fix the CFG scale to \textbf{1.5}, as recommended by Peebles and Xie~\cite{peebles2023scalable}, and tune only the additional guidance scales introduced by each baseline method. For SAG, SEG, $S^2$, APG, EMAG, and related approaches, we perform a brief sweep of their additional scales 
and evaluate FID, PRDC on a 1k-sample subset (Table~\ref{tab:dit_cond_sag_seg_s2_erg_apg_cads}). For ERG, which does not define a separate additional scale, we instead sweep the CFG scale itself over a small range following the authors’ settings, to maintain fairness: other methods with additional scales can effectively strengthen or weaken the overall guidance signal, whereas ERG can only do so via CFG. This protocol allows us to assess the impact of each method under a fixed base CFG strength while fairly optimising its own additional scale for the DiT backbone. Methods such as SAG and SEG are combined with CFG in our setup to enable class-conditional sampling based on the settings provided by the authors.

We note that layering additional guidance on top of CFG often increases FID relative to pure CFG, but typically yields improvements in perceptual quality (HPS~v2) and recall, reflecting a shift along the fidelity--alignment trade-off rather than a degradation in generation quality.We report class-conditioned ImageNet results for DiT at $256^2$ and $512^2$ in Tables~\ref{tab:imagenet_classcond_256_sup} and~\ref{tab:imagenet_classcond_512_supp}, and the corresponding unconditional results in Tables~\ref{tab:merged_uncond_results} and~\ref{tab:imagenet_uncond_512_supp}. Qualitative samples and baseline comparisons are shown in Figures~\ref{fig: section_D_uncond_sd3_dit_512} and~\ref{fig: section_D_baseline_vs_emag_512_dit}.

On DiT-XL/2 at 256$\times$256 resolution, EMAG’s fine-grained refinements remain competitive but do not surpass the strongest baselines; its FID-HPS trade-off is generally tighter than most methods, with the notable exception of $S^{2}$. At 512$\times$512 resolution, where the base model already operates in a higher-quality regime, EMAG yields a more favorable balance between FID and HPS, and in our setting attains the best HPS among all variants while keeping FID comparable to other strong baselines. This trend is consistent with our SD3 results and further suggests that EMAG is particularly effective in high-resolution, high-quality regimes, where subtle attention-level refinements can translate into noticeable gains in human preference without substantially distorting the underlying sample distribution.

We also conduct a comprehensive Pareto comparison for both DiT-XL/2 at 256×256 and 512×512 resolutions. Full numerical results are provided in Tables~\ref{tab:dit_guidance_ablation_256} and~\ref{tab:dit_guidance_ablation_512}, and the corresponding Pareto frontiers are visualized in Figures~\ref{fig: Dit_256_pareto_frontier} and~\ref{fig: Dit_512_pareto_frontier}.

For unconditional DiT experiments, we ablate the guidance scales solely with respect to FID on a 1k-sample subset for each baseline (see Table~\ref{tab:dit_uncond_pag_sag_seg}) and then report final metrics on 50k generated samples using the best FID configuration, see Tables~\ref{tab:imagenet_uncond_512_supp} and \ref{tab:merged_uncond_results}.

\subsection{Further ablation on EMAG hyperparameters}

In Table~\ref{tab:Adaptive_layer_selection_ablation_supp}, the adaptive layer selection strategy yields more balanced performance across PRDC metrics, FID, and HPS. This suggests that adaptive selection navigates the trade-off more effectively than either using a single fixed layer or applying EMAG to all layers simultaneously. We additionally ablate all SD3 transformer layers for completeness; the results are reported in Table~\ref{tab:Adaptive_layer_selection_ablation_supp2}. We observe that the middle layers (L6–L8) provide a favorable trade-off between FID and HPS, reinforcing the design choice discussed in \cite{ifriqi2025entropy}. The adaptive layer selection further improves this balance by dynamically selecting a single layer from the L6–L8 range during guidance.

\begin{figure*}[t]
    \centering
    \includegraphics[width=\linewidth]{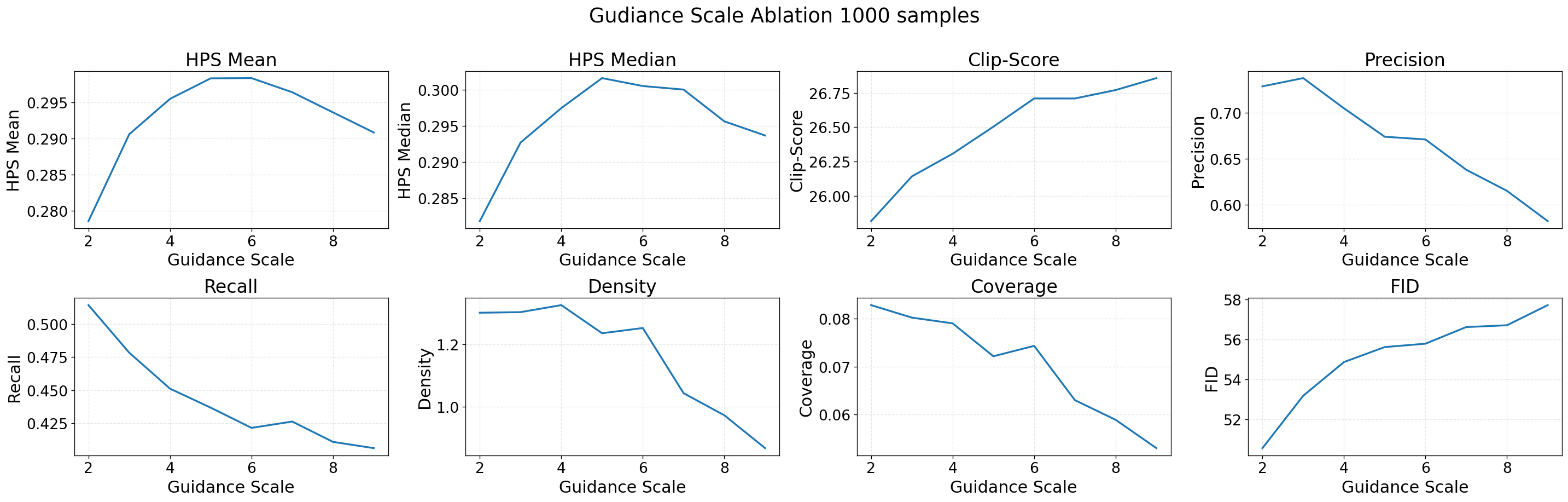}
    \caption{Variation of HPS, FID, Clip-Score, and PRDC metrics as a function of guidance scale for SD3 (1{,}000 samples).}
    \label{fig:emag_guidance_scale_ablation_supp}
\end{figure*}

\begin{figure*}[t]
    \centering
    \includegraphics[width=\linewidth]{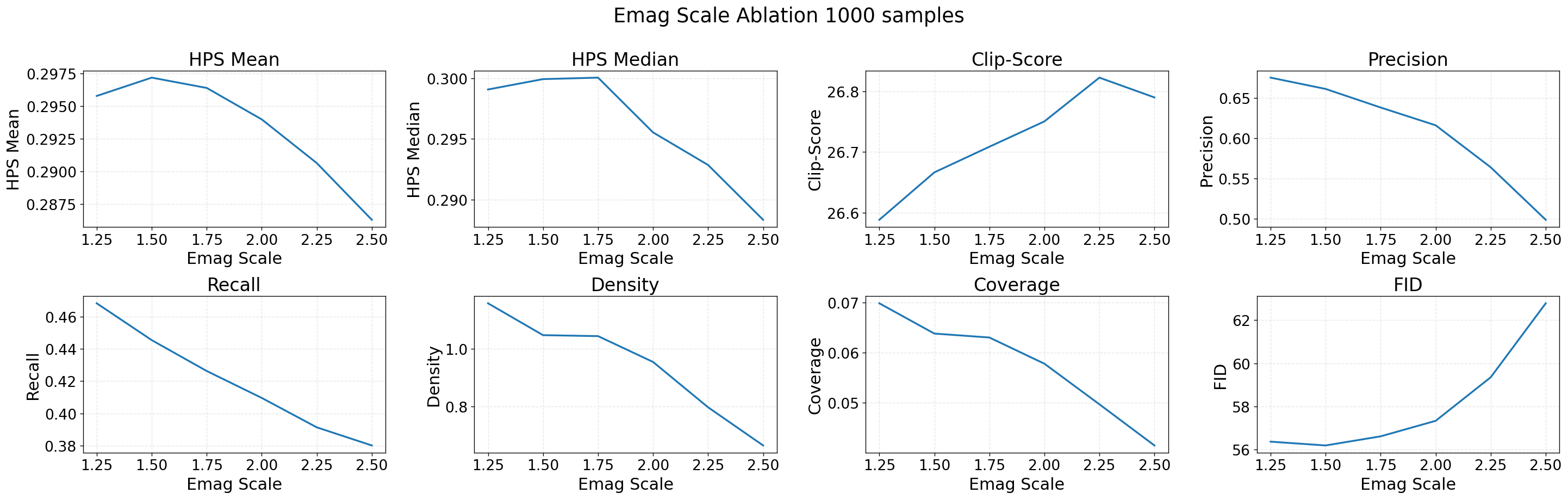}
    \caption{Variation of HPS, FID, Clip-Score, and PRDC metrics as a function of EMAG scale for SD3 (1{,}000 samples).}
    \label{fig:emag_scale_ablation_supp}
\end{figure*}

We further analyze the behaviour of EMAG with respect to PRDC metrics on SD3-Medium in Figures~\ref{fig:emag_guidance_scale_ablation_supp} and~\ref{fig:emag_scale_ablation_supp}, which we use to select a balanced combination of guidance and EMAG scales. While Fig.~\ref{fig:emag_guidance_scale_ablation_supp} examines the effect of varying the guidance scale, whereas Fig.~\ref{fig:emag_scale_ablation_supp} varies the EMAG scale; both studies are evaluated on 1{,}000 generated samples. 

\begin{figure*}[t]
  \centering
  \includegraphics[width=\linewidth]{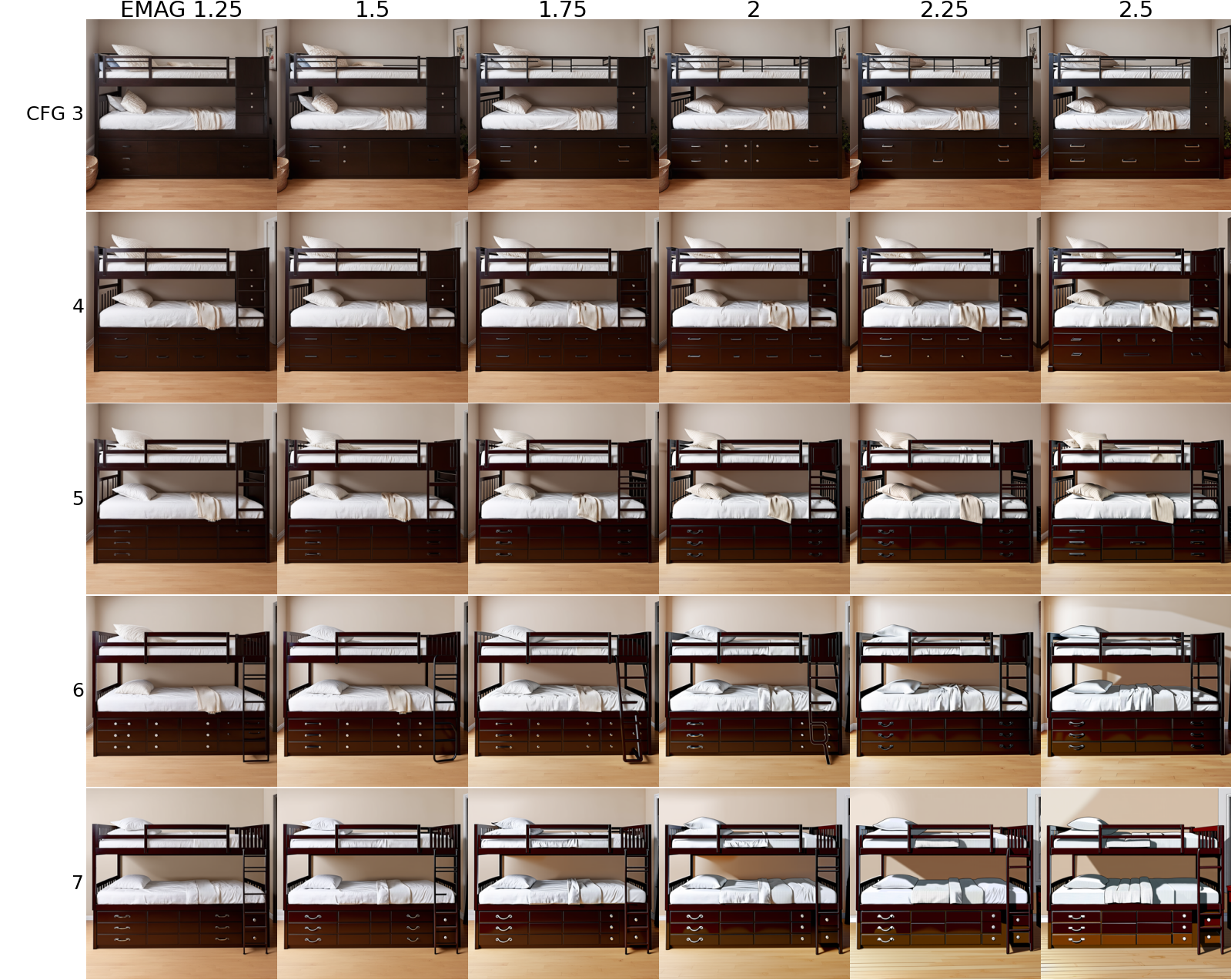}
     \caption{\textbf{Qualitative analysis of the effect of guidance scale and EMAG scale on text-to-image conditional generation with SD3. All samples are generated using the same prompt and the same pseudorandom seed; only the guidance scale (rows) and EMAG scale (columns) are varied. Prompt: \textit{A loft bed with a dresser underneath it.}}}

    \label{fig: section_D_ablation_scale_sd3}
\end{figure*}

We also provide qualitative example for SD3 in Fig.~\ref{fig: section_D_ablation_scale_sd3}, where we analyse the joint impact of guidance scale and EMAG scale on the generated samples. At very high EMAG scales, we observe overexposure and oversharpening artifacts, whereas smaller EMAG scales remain stable even under large guidance scales. In the medium EMAG regime, EMAG effectively refines background structure and fine details while preserving the overall composition.

\begin{figure*}[t]
  \centering
  \includegraphics[width=\linewidth]{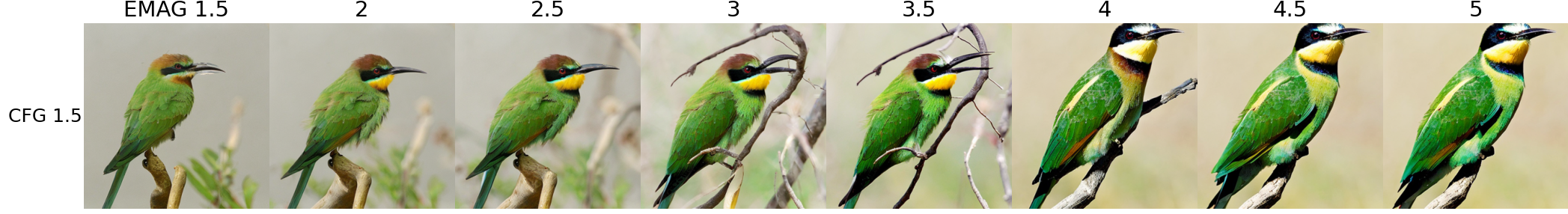}
     \caption{\textbf{Qualitative analysis of the effect of EMAG scale on class-conditional generation with DiT at $512^2$ resolution for fixed CFG scale. All samples are generated using the same prompt and the same pseudo-random seed; \textit{Class label: electric ray}}}
    \label{fig: section_D_ablation_scale_dit}
\end{figure*}

Similarly, Fig.~\ref{fig: section_D_ablation_scale_dit} illustrates the effect of varying the EMAG scale for a fixed guidance scale of 1.5 on DiT-XL/2 at 512$\times$512 resolution. We again find that excessively large EMAG scales lead to overexposed, overly sharpened generations, while medium EMAG scales provide the most visually appealing trade-off, yielding subtle refinements (e.g., in background texture and small object details) without destabilising the base sample.
For completeness, we summarise the EMAG hyperparameters and layer ranges used across different tasks and architectures in Table~\ref{tab:emag_hparams}.


In Table~\ref{tab:timestep_ablation_supp}, we ablate different choices of the start ($\tau_s$) and end ($\tau_e$) time steps for the SD3 experiment on the COCO 2014 validation set for 1000 samples. We observe that setting $\tau_s = T_{\max}$ and applying guidance over the final $20\%$ of denoising steps yields a balanced performance.

In Table~\ref{tab:emag_beta_ablation}, we ablate the impact of $\beta$ on the HPS scores and other PRDC metrics for EMAG. We can observe that at low values of $\beta$ $(\leq 0.6)$, the EMA buffer updates rapidly and closely tracks the current noise prediction, producing negative examples that are nearly identical to the generated output. Since the contrastive distance between the negative and the current prediction is negligible, the guidance signal effectively vanishes and all metrics remain within noise of the CFG-only baseline $(\Delta HPS \approx -0.4)$. As $\beta$ increases, the EMA integrates over a longer temporal window and the negative examples diverge progressively from the current prediction. This growing contrastive gap allows the guidance signal to carry meaningful directional information that steers the denoising trajectory toward higher perceptual quality. Notably, the transition from ineffective to effective guidance occurs when the EMA half-life exceeds the generation horizon: at $\beta = 0.95$ (half-life 13.5 steps), EMAG nearly matches CFG $(\Delta HPS = -0.07)$, while at $\beta = 0.988$ (half-life 57.4 steps, approximately twice the 28-step sampling budget), EMAG surpasses CFG with $\Delta HPS = +0.14$. 
Our finer-grained sweep pinpoints the zero-crossing at $\beta = 0.965$ (half-life $\approx 0.7T$), confirming that guidance becomes net-positive precisely when the EMA half-life exceeds roughly 70\% of the sampling horizon---the point at which the buffer begins retaining information across the majority of the denoising trajectory. Beyond this threshold, returns diminish gracefully: the jump from $\beta = 0.988$ to $\beta = 0.999$ yields only $+0.08$ additional HPS Meanwhile, FID remains stable across the entire $\beta$ range (55.5--56.3), \textit{indicating that the hard negative difficulty primarily modulates per-image perceptual quality rather than distributional fidelity}. The monotonic relationship between $\beta$ and $\Delta HPS$ confirms that $\beta$ acts as a direct control over the information gap between the hard negative and the final generation: lower $\beta$ preserves nearly all information in the negative (easy, uninformative), while higher $\beta$ progressively abstracts away fine-grained details (hard, informative), enabling the guidance signal to recover precisely those details during sampling.


\subsection{Hyperparameter selection for baselines}

In this section, we report the ablation experiments used to select guidance scales for all baseline guidance methods for our main experiments. For SD3, ablations on conditional text-to-image generation are shown in Table~\ref{tab:sd3_cond_sag_seg_s2_erg_apg_cads} and \ref{tab:guidance_methods_comparison_full}, while unconditional generation with null prompts is reported in Table~\ref{tab:sd3_uncond_sag_seg_ierg}. For DiT, the unconditional ablations are provided in Table~\ref{tab:dit_uncond_pag_sag_seg}, and the class conditional ablations are summarized in Tables~\ref{tab:dit_cond_sag_seg_s2_erg_apg_cads},\ref{tab:dit_guidance_ablation_256} and \ref{tab:dit_guidance_ablation_512}.

\begin{table*}[t]
\centering
\small
\setlength{\tabcolsep}{5pt}
\caption{Ablation of guidance scales for unconditional generation using SD3 (null prompts) across multiple baselines. Lower is better for FID; higher is better for all other metrics. All results are reported on 1{,}000 generated samples.}

\label{tab:sd3_uncond_sag_seg_ierg}
\resizebox{\textwidth}{!}{%
\begin{tabular}{lcccccc}
\toprule
\textbf{Method} & \textbf{Hyperparameters} & \textbf{FID} $\downarrow$ & \textbf{Precision} $\uparrow$ & \textbf{Recall} $\uparrow$ & \textbf{Density} $\uparrow$ & \textbf{Coverage} $\uparrow$ \\
\midrule
\multicolumn{7}{l}{\textit{SD3 (Unconditional)}} \\
\midrule
SAG    & SAG Scale $s=0.1$                     & 126.85 & 0.513 & 0.516 & 0.239 & 0.024 \\
SAG    & SAG Scale $s=0.2$                     & 151.01 & 0.389  & 0.408 & 0.147 & 0.015 \\
SAG    & SAG Scale $s=0.3$                     & 214.21 & 0.249 & 0.166 & 0.074 & 0.006 \\
\midrule
SEG    & SEG Scale $s=2$                       & 119.36 & 0.770  & 0.524 & 0.609 & 0.051 \\
SEG    & SEG Scale $s=3$                       & 138.06 & 0.781 & 0.423 & 0.600 & 0.047 \\
SEG    & SEG Scale $s=4$                       & 149.59 & 0.773  & 0.354 & 0.510 & 0.032 \\
\midrule
I-ERG  & ERG Scale $s=2$                       & 111.73 & 0.730 & 0.581 & 0.594 & 0.056 \\
I-ERG  & ERG Scale $s=3$                       & 123.80 & 0.743 & 0.466 & 0.545 & 0.053 \\
I-ERG  & ERG Scale $s=4$                       & 134.66 & 0.681  & 0.340 &  0.474 & 0.043 \\
\midrule
PAG    & PAG Scale $s=0.1$      & 206.54 & 0.336 & 0.000 & 0.251 & 0.003 \\
PAG    & PAG Scale $s=0.3$      & 225.62 & 0.348 & 0.000 & 0.228 & 0.003 \\
PAG    & PAG Scale $s=0.5$      & 261.76 & 0.348 & 0.000 & 0.257 & 0.003 \\
PAG    & PAG Scale $s=0.7$      & 293.69 & 0.308 & 0.000 & 0.241 & 0.002 \\
PAG    & PAG Scale $s=1$        & 308.16 & 0.204 & 0.000 & 0.147 & 0.002 \\
PAG    & PAG Scale $s=2$         & 333.52 & 0.116 & 0.000 & 0.112 & 0.001 \\
PAG    & PAG Scale $s=3$         & 327.39 & 0.072 & 0.000 & 0.069 & 0.001 \\
\bottomrule
\end{tabular}
}
\end{table*}
%

\begin{table*}[t]
\centering
\small
\setlength{\tabcolsep}{5pt}
\caption{Ablation of guidance scales for unconditional generation using DiT-XL/2 at 256$\times$256 and 512$\times$512 resolutions. Lower is better for FID; higher is better for all other metrics. All results are reported on 1{,}000 generated samples per setting.}
\label{tab:dit_uncond_pag_sag_seg}
\resizebox{\textwidth}{!}{%
\begin{tabular}{lcccccc}
\toprule
\textbf{Method} & \textbf{Hyperparameters} & \textbf{FID} $\downarrow$ & \textbf{Precision} $\uparrow$ & \textbf{Recall} $\uparrow$ & \textbf{Density} $\uparrow$ & \textbf{Coverage} $\uparrow$ \\
\midrule
\multicolumn{7}{l}{\textit{DiT-XL/2 (Unconditional), $256^2$}} \\
\midrule
PAG & PAG Scale $s=2$                 & 69.05 & 0.657 & 0.364 & 0.9052 & 0.222 \\
PAG & PAG Scale $s=3$                 & 75.71 & 0.670 & 0.314 & 0.869 & 0.203 \\
PAG & PAG Scale $s=4$                 & 80.12 & 0.635 & 0.287 & 0.848 & 0.196 \\
\midrule
SAG & SAG Scale $s=0.1$                 & 77.87 & 0.535 & 0.536 & 0.567 & 0.156 \\
SAG & SAG Scale $s=0.2$                 & 74.58 & 0.541 & 0.524 & 0.582 & 0.156 \\
SAG & SAG Scale $s=0.3$                 & 73.16 & 0.563 & 0.519 & 0.596 & 0.159 \\
\midrule
SEG & SEG Scale $s=2$                 & 66.36 & 0.682 & 0.388 & 0.961 & 0.238 \\
SEG & SEG Scale $s=3$                 & 75.58 & 0.658 & 0.297 & 0.903 & 0.200 \\
SEG & SEG Scale $s=4$                 & 81.39 & 0.579 & 0.271 & 0.762 & 0.174 \\
\midrule
\multicolumn{7}{l}{\textit{DiT-XL/2 (Unconditional), $512^2$}} \\
\midrule
PAG & PAG Scale $s=2$                 & 84.25 & 0.656 & 0.330 & 1.206 & 0.776 \\
PAG & PAG Scale $s=3$                 & 92.05 & 0.611 & 0.252 & 1.122 & 0.678 \\
PAG & PAG Scale $s=4$                 & 96.64 & 0.587 & 0.250 & 0.996 & 0.682 \\
\midrule
SAG & SAG Scale $s=0.1$                 & 81.55 & 0.601 & 0.451 & 0.997 & 0.781 \\
SAG & SAG Scale $s=0.2$                 & 79.32 & 0.633 & 0.449 & 1.052 & 0.777 \\
SAG & SAG Scale $s=0.3$                 & 79.20 & 0.653 & 0.450 & 1.114 & 0.748 \\
\midrule
SEG & SEG Scale $s=2$                 & 81.15 & 0.703 & 0.336 & 1.410 & 0.767 \\
SEG & SEG Scale $s=3$                 & 93.86 & 0.598 & 0.256 & 1.084 & 0.621 \\
SEG & SEG Scale $s=4$                 & 104.42 & 0.525 & 0.197 & 0.833 & 0.556 \\
\bottomrule
\end{tabular}
}
\end{table*}
\begin{table*}[t]
\centering
\small
\setlength{\tabcolsep}{5pt}
\caption{APG and CADS ablation for SD3 text-to-image generation on the COCO 2014 validation set. Lower is better for FID; higher is better for all other metrics. For APG, the \textit{Scale} column reports the swept guidance scale, and \textit{Hyperparameters} lists the momentum ($\beta$) and rescaling factor ($r$). CADS uses the authors' recommended configuration for Stable Diffusion. All results are reported on 1{,}000 generated images.}
\label{tab:sd3_cond_sag_seg_s2_erg_apg_cads}
\resizebox{\textwidth}{!}{%
\begin{tabular}{lccccccc}
\toprule
\textbf{Method} & \textbf{Scale} & \textbf{Hyperparameters} & \textbf{FID} $\downarrow$ & \textbf{Precision} $\uparrow$ & \textbf{Recall} $\uparrow$ & \textbf{Density} $\uparrow$ & \textbf{Coverage} $\uparrow$ \\
\midrule
\multicolumn{8}{l}{\textit{SD3 (Conditional)}} \\
\midrule
APG         & 10 & $\beta=-0.5,\;r=7.5$    & 55.62 & 0.850 & 0.694 & 1.329 & 0.184 \\
APG         & 12 & $\beta=-0.5,\;r=7.5$    & 55.50 & 0.839 & 0.685 & 1.271 & 0.178 \\
APG         & 15 & $\beta=-0.5,\;r=7.5$    & 55.20 & 0.835 & 0.690 & 1.174 & 0.167 \\
APG         & 10 & $\beta=-0.75,\;r=7.5$   & 55.33 & 0.860 & 0.692 & 1.358 & 0.189 \\
APG         & 12 & $\beta=-0.75,\;r=7.5$   & 55.29 & 0.861 & 0.706 & 1.262 & 0.179 \\
APG         & 15 & $\beta=-0.75,\;r=7.5$   & 54.84 & 0.845 & 0.693 & 1.276 & 0.180 \\
APG         & 10 & $\beta=-0.5,\;r=15.0$   & 55.23 & 0.833 & 0.675 & 1.327 & 0.186 \\
APG         & 12 & $\beta=-0.5,\;r=15.0$   & 55.34 & 0.847 & 0.676 & 1.259 & 0.175 \\
APG         & 15 & $\beta=-0.5,\;r=15.0$   & 54.95 & 0.830 & 0.679 & 1.086 & 0.154 \\
APG         & 10 & $\beta=-0.75,\;r=15.0$  & 55.63 & 0.844 & 0.695 & 1.311 & 0.184 \\
APG         & 12 & $\beta=-0.75,\;r=15.0$  & 55.09 & 0.847 & 0.683 & 1.306 & 0.184 \\
APG         & 15 & $\beta=-0.75,\;r=15.0$  & 55.42 & 0.815 & 0.704 & 1.192 & 0.169 \\
\midrule
\multicolumn{8}{l}{\textit{CADS (settings from the original paper for Stable Diffusion: $w_{\text{cfg}}{=}9.0,\;\tau_1{=}0.6,\;\tau_2{=}0.9,\;s{=}0.25,\;\psi{=}1$)}} \\
\bottomrule
\end{tabular}
}
\end{table*}

\begin{table*}[t]
\centering
\small
\setlength{\tabcolsep}{5pt}

\caption{Ablation of guidance configurations for class-conditional DiT-XL/2 on ImageNet at 256$\times$256 and 512$\times$512 resolutions. Lower is better for FID; higher is better for all other metrics. Each \textit{Resolution} block corresponds to a fixed sampling resolution. For SAG, SEG, and $S^{2}$, the \textit{Scale} column reports the fixed CFG scale, while \textit{Hyperparameters} lists the method-specific scale swept in this ablation (e.g., SAG/SEG scale $s$, $S^{2}$ scale $w$). For ERG, the \textit{Scale} column itself is the swept guidance scale. APG and CADS use the authors' recommended configurations for the corresponding resolution. All results are reported on 1{,}000 generated images. For detailed Pareto Analysis see Tables~\ref{tab:dit_guidance_ablation_256} and \ref{tab:dit_guidance_ablation_512} }

\label{tab:dit_cond_sag_seg_s2_erg_apg_cads}
\resizebox{\textwidth}{!}{%
\begin{tabular}{lccccccc}
\toprule
\textbf{Method} & \textbf{Scale} & \textbf{Hyperparameters} & \textbf{FID} $\downarrow$ & \textbf{Precision} $\uparrow$ & \textbf{Recall} $\uparrow$ & \textbf{Density} $\uparrow$ & \textbf{Coverage} $\uparrow$ \\
\midrule
\multicolumn{8}{l}{\textit{DiT-XL/2 (Conditional), $256^2$}} \\
\midrule
SAG + CFG   & 1.5 & SAG scale $s=0.1$               & 54.56 &  0.936 & 0.702 & 1.351 & 0.455 \\
SAG + CFG   & 1.5 & SAG scale $s=0.2$               & 54.86 & 0.927 & 0.697 & 1.353 & 0.451 \\
SAG + CFG   & 1.5 & SAG scale $s=0.3$               & 54.44 & 0.920 & 0.694 & 1.359 & 0.456 \\
\midrule
SEG + CFG   & 1.5 & SEG scale $s=2$                 & 52.25 & 0.859 & 0.644 & 1.198 & 0.412 \\
SEG + CFG   & 1.5 & SEG scale $s=3$                 & 54.53 & 0.760 & 0.610 & 1.039 & 0.359 \\
SEG + CFG   & 1.5 & SEG scale $s=4$                 & 59.51 & 0.656 & 0.573 & 0.873 &  0.301 \\
\midrule
$S^{2}$     & 1.5 & $w=0.01$                      & 44.39 & 0.808 & 0.735 & 1.154 & 0.420 \\
$S^{2}$     & 1.5 & $w=0.05$                      & 48.84 & 0.719 & 0.727 & 0.940 & 0.353 \\
$S^{2}$     & 1.5 & $w=0.1$                      & 83.72 & 0.403 & 0.625 & 0.379 & 0.147 \\
$S^{2}$     & 1.5 & $w=0.15$                      & 140.74 & 0.231 & 0.447 & 0.135 & 0.041 \\
$S^{2}$     & 1.5 & $w=0.20$                      & 187.93 & 0.209 & 0.241 & 0.100 & 0.010 \\
$S^{2}$     & 1.5 & $w=0.25$                      & 233.99 & 0.189 & 0.035 & 0.117 & 0.003 \\
$S^{2}$     & 1.5 & $w=0.30$                      & 270.78 & 0.206 & 0.003 & 0.193 & 0.002 \\
\midrule
ERG         & 1.5 & Setting from Original paper          & 44.27 & 0.857 & 0.702 & 1.278 & 0.458 \\
ERG         & 2.0 & Setting from Original paper          & 48.09 & 0.896 & 0.673 & 1.376 & 0.467 \\
ERG         & 3.0 & Setting from Original paper          & 51.90 & 0.857 & 0.650 & 1.255 & 0.436 \\
ERG         & 4.0 & Setting from Original paper          & 53.59 & 0.815 & 0.625 & 1.142 & 0.383 \\
\midrule
\multicolumn{8}{l}{\textit{APG (We use the settings from the original paper for DiT ImageNet-256)}} \\
\midrule
\multicolumn{8}{l}{\textit{CADS (We use the settings from the original paper for DiT ImageNet-256)}}\\

\midrule
\multicolumn{8}{l}{\textit{DiT-XL/2 (Conditional), $512^2$}} \\
\midrule
SAG + CFG   & 1.5 & SAG scale $s=0.1$                      & 53.87 & 0.845 & 0.713 & 1.024 & 0.956 \\
SAG + CFG   & 1.5 & SAG scale $s=0.2$                      & 53.39 & 0.844 & 0.709 & 1.068 & 0.966 \\
SAG + CFG   & 1.5 & SAG scale $s=0.3$                      & 53.17 & 0.850 & 0.694 & 1.117 & 0.967 \\
\midrule
SEG + CFG   & 1.5 & SEG scale $s=2$                      & 57.91 & 0.777 & 0.567 & 1.252 & 0.951 \\
SEG + CFG   & 1.5 & SEG scale $s=3$                      & 68.38 & 0.695 & 0.519 & 1.072 & 0.918 \\
SEG + CFG   & 1.5 & SEG scale $s=4$                      & 79.57 & 0.610 & 0.458 & 0.867 & 0.828 \\
\midrule
$S^{2}$     & 1.5 & $w=0.01$                      & 44.94 & 0.850 & 0.695 & 1.216 & 0.977 \\
$S^{2}$     & 1.5 & $w=0.05$                      & 58.00 & 0.701 & 0.662 & 0.883 & 0.949 \\
$S^{2}$     & 1.5 & $w=0.1$                      & 124.33 & 0.315 & 0.497 & 0.258 & 0.593 \\
$S^{2}$     & 1.5 & $w=0.15$                      & 200.08 & 0.131 & 0.263 & 0.065 & 0.135 \\
$S^{2}$     & 1.5 & $w=0.20$                      & 253.80 & 0.154 & 0.049 & 0.075 & 0.065 \\
$S^{2}$     & 1.5 & $w=0.25$                      & 305.95 &  0.170 & 0.000 & 0.047 & 0.024 \\
$S^{2}$     & 1.5 & $w=0.30$                      & 387.04 & 0.051 & 0.000 & 0.010 & 0.002 \\
\midrule
ERG         & 1.5 & Setting from Original paper                      & 45.26 & 0.864 & 0.669 & 1.508 & 0.984 \\
ERG         & 2.0 & Setting from Original paper                      & 48.67 & 0.860 & 0.635 & 1.365 & 0.978 \\
ERG         & 3.0 & Setting from Original paper                      & 54.51 & 0.831 & 0.573 & 1.311 & 0.960 \\
ERG         & 4.0 & Setting from Original paper                      & 60.76 & 0.767 & 0.533 & 1.171 & 0.943 \\
\midrule
\multicolumn{8}{l}{\textit{APG (We use the settings from the original paper for DiT ImageNet-512)}}\\
\midrule
\multicolumn{8}{l}{\textit{CADS (We use the settings from the original paper for DiT ImageNet-512)}}\\
\bottomrule
\end{tabular}
}
\end{table*}
\begin{table*}[t]
\centering
\small
\setlength{\tabcolsep}{5pt}
\caption{FID--HPS trade-off comparison between attention-perturbation baselines (SAG, SEG, ERG) and EMAG on SD3. Baselines exhibit a pronounced trade-off: configurations that improve FID degrade HPS, and vice versa. EMAG achieves competitive performance on both axes simultaneously without navigating this trade-off. All methods evaluated on 40K images conditioned on COCO 2014 validation captions. Best per group is \underline{underlined}; overall best is \textbf{bolded}.}
\label{tab:sd3_baseline_tradeoff}
\resizebox{\textwidth}{!}{%
\begin{tabular}{llccccccc}
\toprule
\textbf{Method} & \textbf{Config} & \textbf{FID} $\downarrow$ & \textbf{CLIP} $\uparrow$ & \textbf{Precision} $\uparrow$ & \textbf{Recall} $\uparrow$ & \textbf{Density} $\uparrow$ & \textbf{Coverage} $\uparrow$ & \textbf{HPS} $\uparrow$ \\
\midrule
\multicolumn{9}{l}{\textit{ERG — varying guidance scale (no CFG)}} \\
\midrule
ERG (FID-optimal)  & Scale$=$2  & \underline{\textbf{17.89}} & 25.34 & \underline{\textbf{0.709}} & \underline{\textbf{0.307}} & \underline{\textbf{1.199}} & \underline{\textbf{0.679}} & 28.36  \\
ERG (HPS-optimal)  & Scale$=$3  & 23.99 & \underline{25.71} & 0.660 & 0.209 & 1.025 & 0.580 & \underline{29.35} \\
\midrule
\multicolumn{9}{l}{\textit{SAG — varying SAG scale (CFG$=$7)}} \\
\midrule
SAG                & $s$$=$0.1, CFG$=$7  & \underline{22.90} & 26.66 & \underline{0.682} & \underline{0.276} & \underline{1.128} & \underline{0.597} & \underline{29.30} \\
SAG                & $s$$=$0.25, CFG$=$7 & 23.04 & \underline{\textbf{26.71}} & 0.679 & 0.265 & 1.121 & 0.597 & 29.28  \\
\midrule
\multicolumn{9}{l}{\textit{SEG — varying SEG scale (CFG$=$7)}} \\
\midrule
SEG (FID-optimal)  & $s$$=$3.0, CFG$=$7  & \underline{23.45} & 25.45 & \underline{0.656} & \underline{0.206} & \underline{1.012} & \underline{0.578} & 29.25 \\
SEG (HPS-optimal)  & $s$$=$2.0, CFG$=$7  & 26.18 & \underline{26.70} & 0.637 & 0.183 & 0.984 & 0.538 & \underline{\textbf{30.15}} \\
\midrule
\multicolumn{9}{l}{\textit{EMAG (ours)}} \\
\midrule
EMAG               & $w_e$$=$1.75, CFG$=$7  & \underline{22.15} & 26.63 & 0.663 & \underline{0.263} & 1.082 & \underline{0.595} & 29.56 \\
EMAG               & $w_e$$=$1.50, CFG$=$7  & 22.89 & \underline{26.66} & \underline{0.669} & 0.260 & \underline{1.097} & 0.591 & \underline{29.76} \\
\bottomrule
\end{tabular}
}
\end{table*}

\begin{table*}[t]
\centering
\small
\setlength{\tabcolsep}{6pt}
\caption{We compare the impact of the adaptive layer selection strategy. The experiments are conducted for 5000 samples. All runs use the identical COCO 2014 validation set, sampling steps, and evaluation code. }
\label{tab:Adaptive_layer_selection_ablation_supp}
\resizebox{\columnwidth}{!}{
\begin{tabular}{lccccccc}
\toprule
\textbf{Model / Guidance} & \textbf{FID} $\downarrow$ & \textbf{CLIP Score} $\uparrow$ & \textbf{Precision} $\uparrow$ & \textbf{Recall} $\uparrow$ & \textbf{Density} $\uparrow$ & \textbf{Coverage} $\uparrow$ & \textbf{HPS} $\uparrow$\\
\midrule
\multicolumn{8}{l}{\textit{SD3(MMDiT)}} \\
\midrule
\textbf{EMAG (L 6)}    & \textbf{27.47} & 26.604 & \textbf{0.824} & \textbf{0.548} & \textbf{1.102} & \textbf{0.452} & 29.54 \\ 
\textbf{EMAG (L 7)}    & 28.94  & 26.733 & 0.788 & 0.508 & 1.008 & 0.424 & 29.51  \\ 
\textbf{EMAG (L 8)}    & 28.23  & 26.768 & 0.810 & 0.543 & 1.040 & 0.431 & \textbf{29.66}  \\ 
\textbf{EMAG (L all)}    & 35.60  & \textbf{26.849} & 0.688 & 0.465 & 0.671 & 0.311 & 28.80 \\ 
\midrule
\textbf{EMAG (Adaptive)}    & 28.52  & 26.733 & 0.806 & 0.530 & 1.040 & 0.430 & 29.60 \\ 
\bottomrule
\end{tabular}
}
\end{table*}


\begin{table*}[t]
\centering
\small
\setlength{\tabcolsep}{6pt}
\caption{Quantitative class-conditional results on ImageNet at 256$\times$256 resolution. Lower is better for FID; all other metrics are higher-is-better. All methods share identical samples, sampling steps, and evaluation code. We fix the CFG scale to 1.5 for all methods; for additional  guidance scales are selected via a brief hyperparameter sweep (Sec.~\ref{sec:additional_experimetns_and_ablation}).}

\label{tab:imagenet_classcond_256_sup}
\resizebox{\textwidth}{!}{%
\begin{tabular}{lcccccc}
\toprule
\textbf{Model / Guidance} & \textbf{FID} $\downarrow$ & \textbf{Precision} $\uparrow$ & \textbf{Recall} $\uparrow$ & \textbf{Density} $\uparrow$ & \textbf{Coverage} $\uparrow$ & \textbf{HPS} $\uparrow$ \\
\midrule
\multicolumn{7}{l}{\textit{DiT-XL/2 \cite{peebles2023scalable}} 256x256} \\
\midrule
(No CFG)          &  9.53 & 0.668 & 0.678 & 0.855 & 0.997 & 21.64\\
CFG               &  \textbf{2.30} & 0.826 & 0.575 & 1.188 & \textbf{0.999} & 23.41 \\
SAG  & 13.53 & 0.932 & 0.214 & 1.335 & 0.990 & 25.60 \\
SEG & 13.14 & 0.838 & 0.167 & 1.208 & 0.984 & \textbf{26.41} \\
APG  &  11.53 & 0.930 & 0.353 & 1.338 & 0.997 & 25.33 \\
CADS       &  2.70 & 0.711 & 0.674 & 0.918 & 0.998 & 22.97 \\
$S^{2}$  &  2.40 & 0.891 & \textbf{0.689} & 1.161 & \textbf{0.999} & 24.38 \\
I-ERG &  8.37 & \textbf{0.945} & 0.355 & \textbf{1.350} & 0.997  & 25.69\\
\midrule
\textbf{EMAG (ours)}&  4.16 & 0.872 & 0.517 & 1.292 & \textbf{0.999} & 24.48\\
\bottomrule
\end{tabular}
}
\end{table*}

\begin{table*}[t]
\centering
\small
\setlength{\tabcolsep}{6pt}
\caption{Quantitative class-conditional results on ImageNet at 512$\times$512 resolution. Lower is better for FID; all other metrics are higher-is-better. All methods share identical samples, sampling steps, and evaluation code. We fix the CFG scale to 1.5 for all methods; for additional guidance scales are select via a brief hyperparameter sweep (Sec.~\ref{sec:additional_experimetns_and_ablation}).}
\label{tab:imagenet_classcond_512_supp}
\resizebox{\textwidth}{!}{%
\begin{tabular}{lcccccc}
\toprule
\textbf{Model / Guidance} & \textbf{FID} $\downarrow$ & \textbf{Precision} $\uparrow$ & \textbf{Recall} $\uparrow$ & \textbf{Density} $\uparrow$ & \textbf{Coverage} $\uparrow$ & \textbf{HPS} $\uparrow$ \\
\midrule
\multicolumn{7}{l}{\textit{DiT-XL/2 \cite{peebles2023scalable}} 512x512} \\
\midrule
(No CFG)           &  11.78 & 0.750 & 0.646 & 1.071 & 1.000 & 21.40 \\
CFG                 &  \textbf{3.08} & 0.838 & 0.532 & 1.219 & 1.000 & 23.68 \\
SAG                 &  18.58 & 0.843 & 0.134 & 1.075 & 1.000 & 25.55 \\
SEG                 &  22.51 & 0.789 & 0.069 & 1.298 & 1.000 & 24.55 \\
APG &  10.24 & 0.849 & 0.347 & 1.102 & 1.000 & 25.25 \\
CADS       &  11.19 & 0.688 & \textbf{0.719} & 0.785 & 1.000 & 21.54 \\
$S^{2}$ &  6.20 & 0.831 & 0.005 & 1.178 & 1.000 & 24.17 \\
I-ERG    &  5.00 & 0.854 & 0.435 & \textbf{1.477} & 1.000 &  24.64 \\
\midrule
\textbf{EMAG (ours)}&  7.59 & \textbf{0.864} & 0.094 & 1.343 & 1.000  & \textbf{25.66}\\
\bottomrule
\end{tabular}
}
\end{table*}

\begin{table*}[t]
\centering
\small
\setlength{\tabcolsep}{6pt}
\caption{Quantitative results for unconditional ImageNet at 512$\times$512 resolution. Lower is better for FID; all other metrics are higher-is-better. All methods share identical data, sampling steps, and evaluation code. Settings follow authors' recommendations or, if unspecified, a brief hyperparameter sweep (Sec.~\ref{sec:additional_experimetns_and_ablation}).}
\label{tab:imagenet_uncond_512_supp}
\resizebox{\textwidth}{!}{%
\begin{tabular}{lccccc}
\toprule
\textbf{Model / Guidance} & \textbf{FID} $\downarrow$ & \textbf{Precision} $\uparrow$ & \textbf{Recall} $\uparrow$ & \textbf{Density} $\uparrow$ & \textbf{Coverage} $\uparrow$ \\
\multicolumn{6}{l}{\textit{DiT-XL/2 \cite{peebles2023scalable}} 512x512} \\
\midrule
\midrule
(No Guidance)           &  55.58 & 0.519 & \textbf{0.652} & 0.579 & 0.995  \\
PAG              &  54.62 & 0.622 & 0.327 & 1.112 & 0.991  \\
SAG               &  46.55 & 0.586 & 0.633 & 0.764 & \textbf{0.998}  \\
SEG  &  45.55 & 0.711 &  0.001 & 1.422 & 0.987  \\
\midrule
\textbf{EMAG (ours)}     &  \textbf{45.01} & \textbf{0.762} & 0.001 & \textbf{1.664} & 0.985 \\
\bottomrule
\end{tabular}
}
\end{table*}

\subsection{Pareto Frontier Analysis}
\label{sec:pareto_frontier_analysis}

We perform an extensive Pareto frontier analysis across both the SD3-Medium (text-conditional) and DiT-XL/2 (class-conditional) architectures. For each method, we jointly sweep the classifier-free guidance (CFG) scale and the method-specific control parameter (where applicable), yielding a comprehensive mapping of the fidelity--alignment--diversity trade-off space.

\subsubsection{SD3-Medium}

\begin{figure*}[t]
    \centering
    \includegraphics[width=\textwidth]{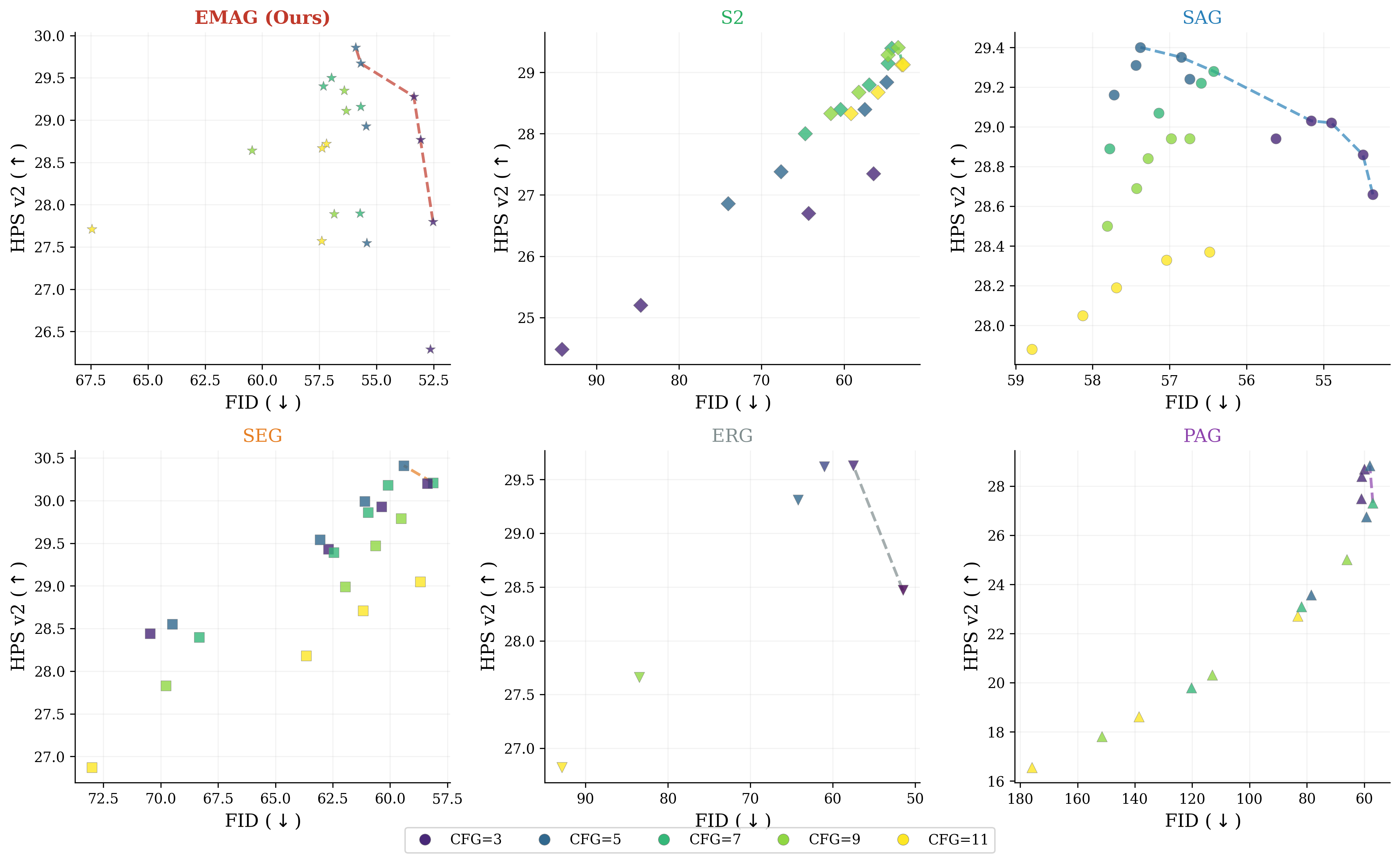}
    \caption{Pareto frontiers for SD3 guidance-scale and  method-specific scale sweeps for baseline guidance methods and EMAG.}
    \label{fig:pareto_frontier_curve_method_specific}
\end{figure*}

We evaluate approximately 108 unique configurations across all methods on MS-COCO 2014 (1K samples). Each configuration is assessed on FID ($\downarrow$), HPS~v2 ($\uparrow$), Precision ($\uparrow$), and Recall ($\uparrow$). The pairwise Pareto frontiers are shown in
Fig.~\ref{fig:pareto_frontier_curve}, and the per-method frontiers in Fig.~\ref{fig:pareto_frontier_curve_method_specific}; full numerical results appear in Table~\ref{tab:guidance_methods_comparison_full}.

\paragraph{Quality--Alignment Trade-off (FID vs.\ HPS).} As shown in the left panel of Fig.~\ref{fig:pareto_frontier_curve}, EMAG consistently occupies the top-right region of the FID--HPS space, achieving the most favorable trade-off among all methods. SAG provides a competitive frontier at lower FID ranges, while PAG collapses to low HPS at elevated FID values, consistent with its architectural incompatibility with MMDiT joint attention (see Sec.~\ref{sec:pag_implementation}). ERG exhibits a narrow operating range with few Pareto-optimal points.

\paragraph{Fidelity--Diversity Trade-off (FID vs.\ Recall).} The middle panel reveals that EMAG maintains competitive recall across the full FID range without the sharp recall degradation observed in methods such as SEG
and S$^{2}$ at low FID. This indicates that EMAG's quality gains do not come at the expense of sample diversity.

\paragraph{Quality--Diversity Trade-off (Precision vs.\ Recall).} The right panel shows that EMAG achieves a balanced operating point, avoiding the precision--recall collapse (high precision, low recall) that characterizes
overly aggressive guidance. In contrast, several baselines cluster in the high-precision, low-recall corner at their strongest guidance settings.

\paragraph{Per-Method Frontier Analysis.} Fig.~\ref{fig:pareto_frontier_curve_method_specific} decomposes the joint Pareto analysis into individual method frontiers, color-coded by CFG scale. EMAG traces a smooth, monotonic frontier across CFG values, indicating stable behavior under guidance scaling. PAG, by contrast, shows erratic FID--HPS trajectories with several dominated configurations, and ERG yields a near-degenerate frontier with only a small number of non-dominated points.

\subsubsection{DiT-XL/2}

\begin{figure*}[t]
  \centering
  \includegraphics[width=\linewidth]{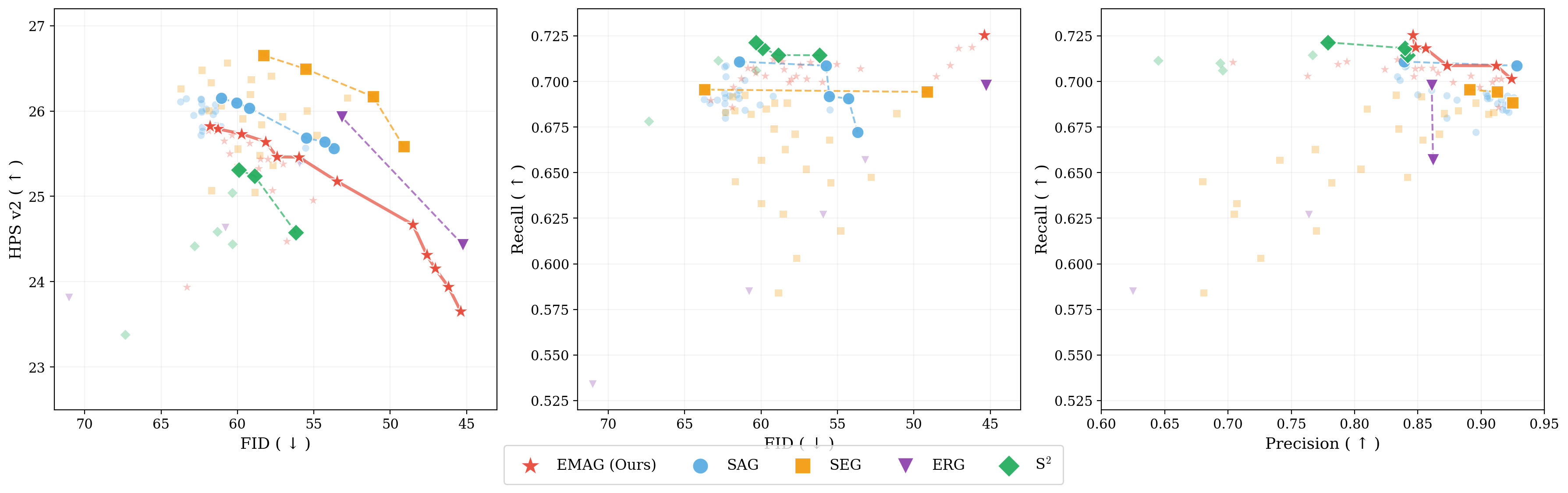}
      \caption{\textbf{Pairwise Pareto frontiers across guidance methods on ImageNet-1K dataset (1K samples, DIT-XL/2 256)}. Each point represents a specific guidance configuration; axes are oriented so that the \textit{top-right corner is optimal} in all the three panels. \textbf{(Left)} FID vs. HPS: the primary quality-preference trade-off. \textbf{(Middle)} FID vs. Recall: fidelity-diversity trade-off. \textbf{(Right)} Precision vs. Recall: quality-diversity trade-off.}

    \label{fig: Dit_256_pareto_frontier}
\end{figure*}

\begin{figure*}[t]
  \centering
  \includegraphics[width=\linewidth]{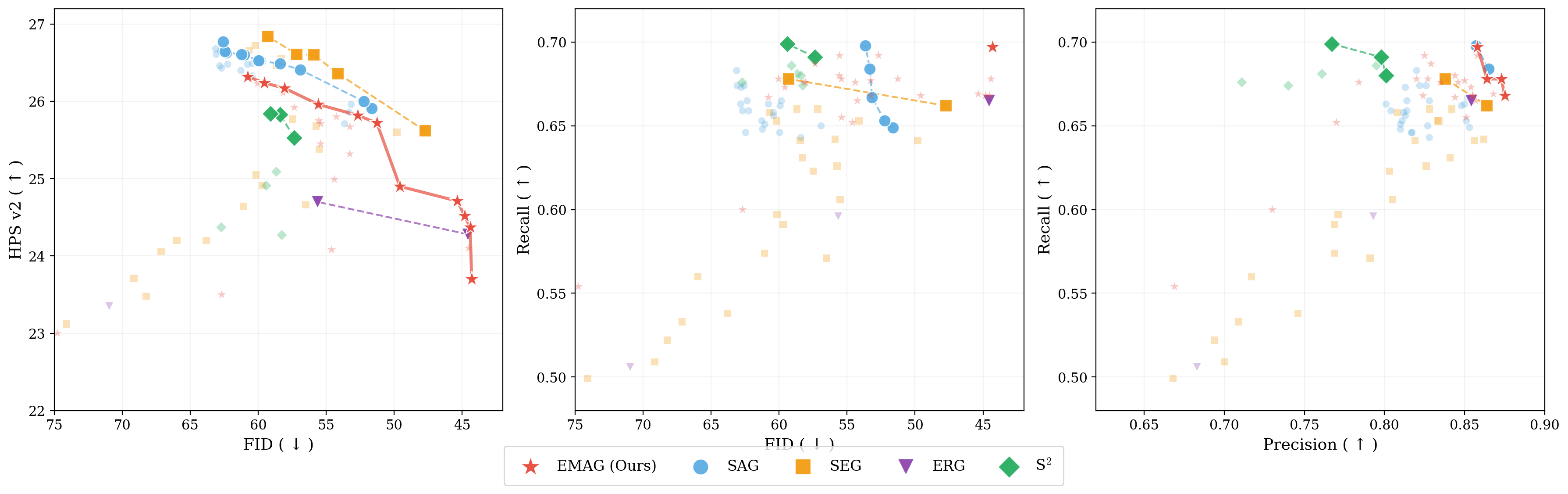}
      \caption{\textbf{Pairwise Pareto frontiers across guidance methods on ImageNet-1K dataset (1K samples, DIT-XL/2 512)}. Each point represents a specific guidance configuration; axes are oriented so that the \textit{top-right corner is optimal} in all the three panels. \textbf{(Left)} FID vs. HPS: the primary quality-preference trade-off. \textbf{(Middle)} FID vs. Recall: fidelity-diversity trade-off. \textbf{(Right)} Precision vs. Recall: quality-diversity trade-off.} 

    \label{fig: Dit_512_pareto_frontier}
\end{figure*}

We conduct analogous sweeps for DiT-XL/2 on ImageNet-1K at both $256\times256$ and $512\times512$ resolutions. Full numerical results are provided in Tables~\ref{tab:dit_guidance_ablation_256} and~\ref{tab:dit_guidance_ablation_512}; the corresponding Pareto frontiers are visualized in Figs.~\ref{fig: Dit_256_pareto_frontier} and~\ref{fig: Dit_512_pareto_frontier}.

\begin{figure*}[t]
  \centering
  \includegraphics[width=\linewidth]{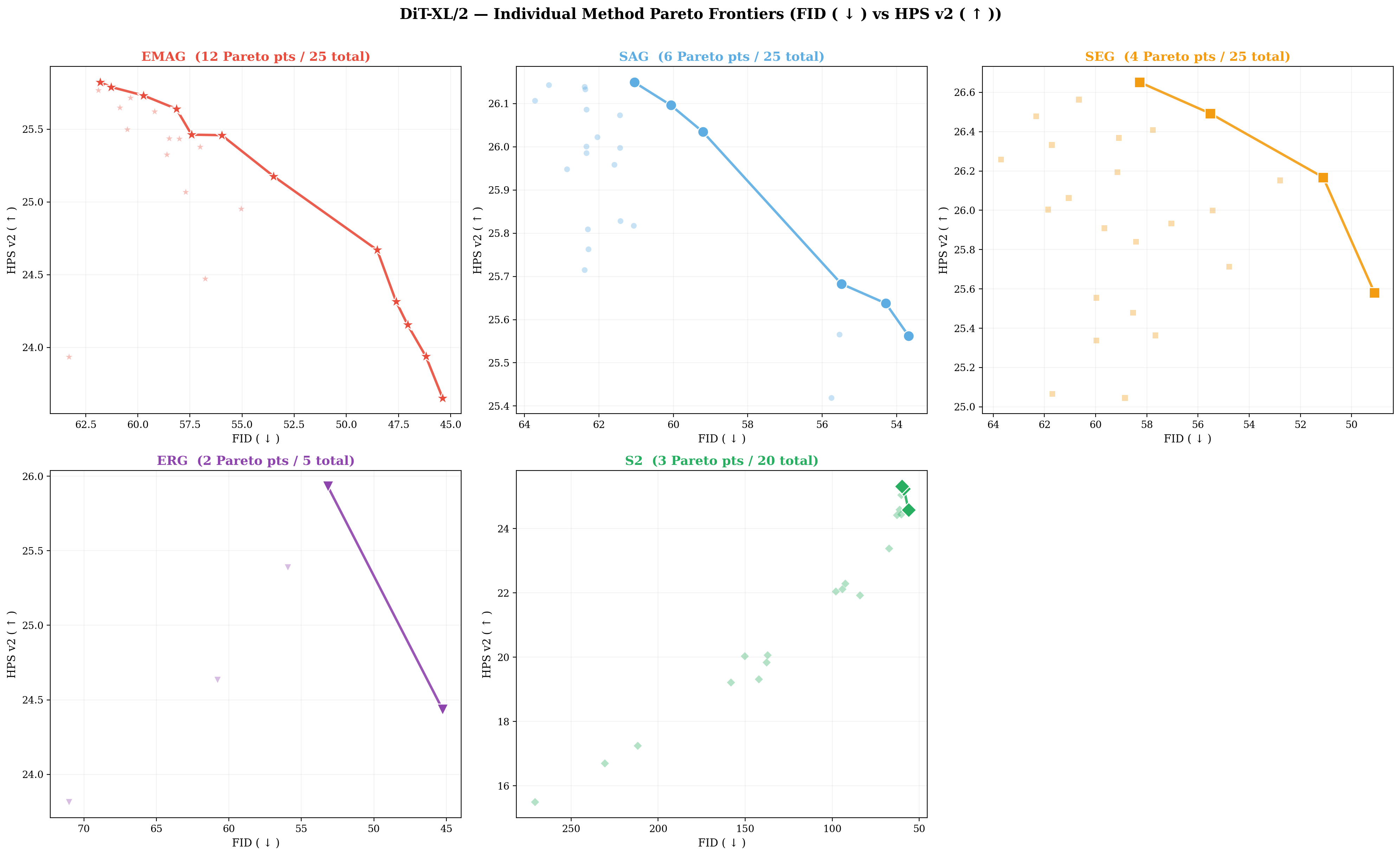}
      \caption{Pareto frontiers for guidance-scale and method-specific scale sweeps for baseline guidance methods and EMAG on DIT-XL/2 256x256 }

    \label{fig: Dit_256_pareto_frontier_by_method}
\end{figure*}

\paragraph{256$\times$256 Resolution.}
As shown in Fig.~\ref{fig: Dit_256_pareto_frontier} (left panel), the methods are more tightly clustered in FID--HPS space compared to SD3, reflecting the simpler class-conditional generation setting. Nevertheless, EMAG achieves the highest Pareto efficiency: 12 out of 25 configurations are Pareto-optimal (Fig.~\ref{fig: Dit_256_pareto_frontier_by_method}), compared to 6/25 for SAG, 4/25 for SEG, and only 2/5 for ERG. This indicates that EMAG produces fewer dominated (wasteful) configurations across its parameter space. In the FID--Recall and Precision--Recall panels, EMAG again maintains a balanced trade-off without collapsing toward either extreme.

\paragraph{512$\times$512 Resolution.}

\begin{figure*}[t]
  \centering
  \includegraphics[width=\linewidth]{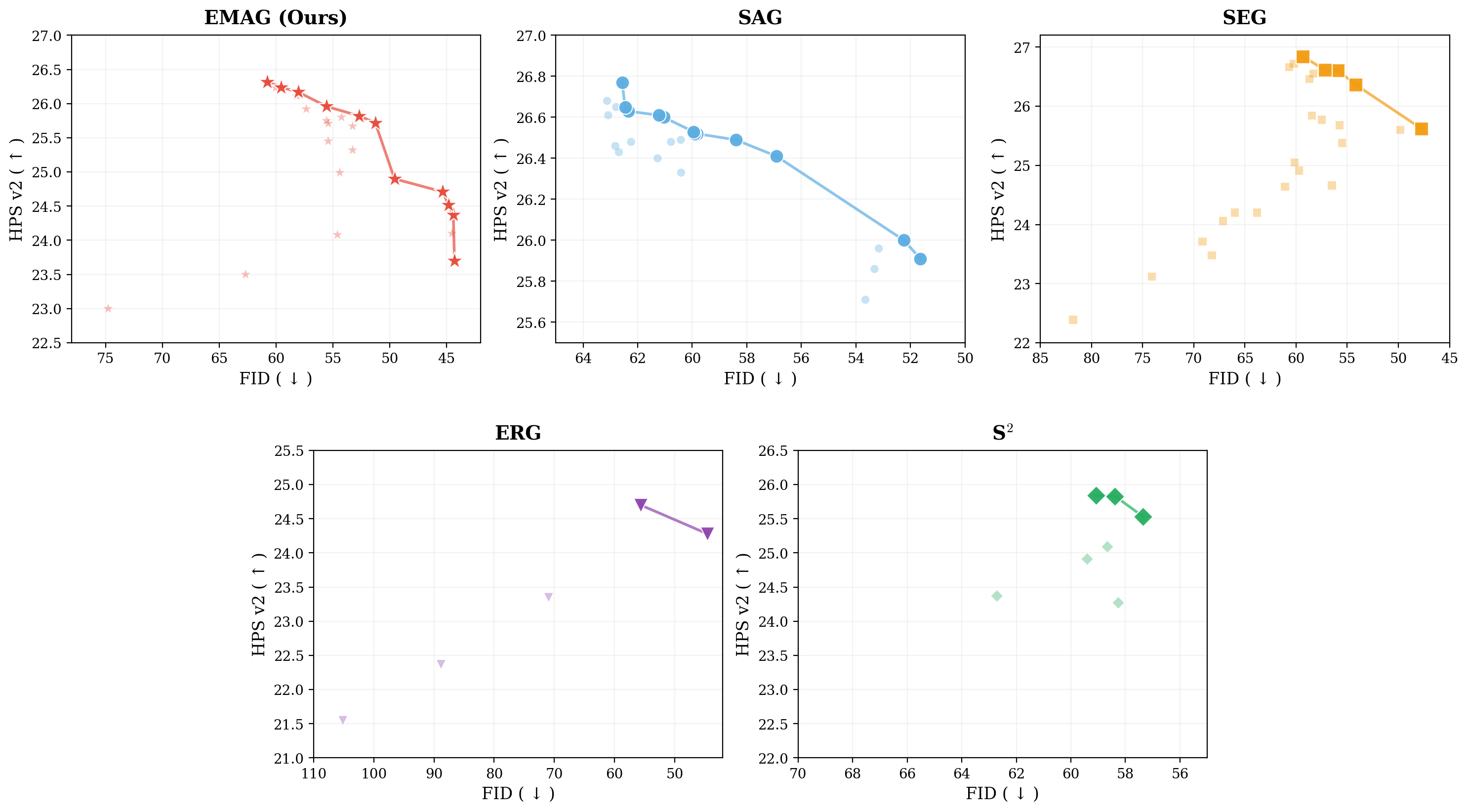}
      \caption{Pareto frontiers for guidance-scale and method-specific scale sweeps for baseline guidance methods and EMAG on DIT-XL/2 512x512 }

   \label{fig: Dit_512_pareto_frontier_by_method}
\end{figure*}

Scaling to $512{\times}512$ amplifies the separation between methods (Figs.~\ref{fig: Dit_512_pareto_frontier} and~\ref{fig: Dit_512_pareto_frontier_by_method}). EMAG achieves the lowest FID of any method (44.30) with the highest recall (0.697), and remains stable under increasing CFG scale as FID stays within 44--61 across all scales. ERG, competitive at $256{\times}256$, degrades catastrophically: FID rises from 44.58 to 105.14 across CFG 1.5--9, a $+$28-point degradation at CFG$=$7 versus only $-$1.0 for EMAG. SAG and SEG remain stable but never reach below FID 51 and 47 respectively. S$^2$ is the most fragile, with scales $w{\geq}0.25$ producing FID$>$100. EMAG yields 11 Pareto-optimal points spanning the widest operating range of any method, while ERG collapses to just 2. These results suggest that EMAG's prediction-level guidance scales more favourably with resolution, as compared to competing methods that perturb internal attention mechanisms.




\begin{figure*}[t]
    \centering
    \includegraphics[width=0.98\textwidth]{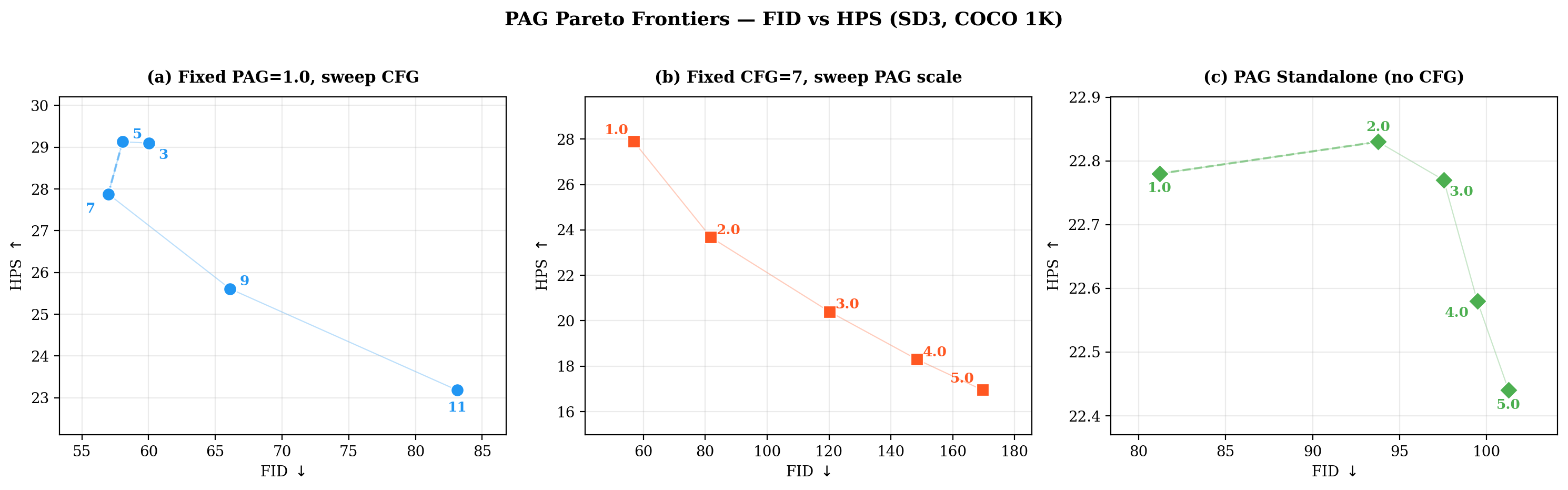}
    \caption{Pareto frontiers for PAG on SD3-Medium for guidance-scale, method-specific scale and standalone guidance sweep.}
    \label{fig:pag_pareto_frontier_curve}
\end{figure*}
\clearpage
{\scriptsize
\begin{longtable}{llccccccc}
\caption{Full comparison of guidance methods with varying hyperparameters. All experiments are conducted on 1000 samples from the COCO 2014 validation set using identical sampling steps, evaluation protocols, and a fixed seed of 8. Corresponding Pareto frontier curves are shown in Fig.~\ref{fig:pareto_frontier_curve}.}
\label{tab:guidance_methods_comparison_full} \\
\toprule
\textbf{Method} & \textbf{Scale} & \textbf{CFG Scale} & \textbf{FID} $\downarrow$ & \textbf{Precision} $\uparrow$ & \textbf{Recall} $\uparrow$ & \textbf{Density} $\uparrow$ & \textbf{Coverage} $\uparrow$ & \textbf{HPS} $\uparrow$\\
\midrule
\endfirsthead
\multicolumn{9}{l}{\small\textit{Table~\ref{tab:guidance_methods_comparison_full} continued from previous page}} \\
\toprule
\textbf{Method} & \textbf{Scale} & \textbf{CFG Scale} & \textbf{FID} $\downarrow$ & \textbf{Precision} $\uparrow$ & \textbf{Recall} $\uparrow$ & \textbf{Density} $\uparrow$ & \textbf{Coverage} $\uparrow$ & \textbf{HPS} $\uparrow$\\
\midrule
\endhead
\midrule
\multicolumn{9}{r}{\small\textit{Continued on next page}} \\
\endfoot
\bottomrule
\endlastfoot
\multicolumn{9}{l}{\textit{SEG}} \\
\midrule
SEG & $s=1.0$ & 7 & 58.13 & 0.813 & 0.640 & 1.069 & 0.150 & 30.21 \\
SEG & $s=2.0$ & 3 & 58.38 & 0.823 & 0.604 & 1.114 & 0.160 & 30.20 \\
SEG & $s=2.0$ & 5 & 59.40 & 0.809 & 0.611 & 1.055 & 0.148 & 30.41 \\
SEG & $s=2.0$ & 7 & 60.10 & 0.786 & 0.610 & 0.971 & 0.139 & 30.18 \\
SEG & $s=2.0$ & 9 & 59.52 & 0.767 & 0.623 & 0.961 & 0.138 & 29.79 \\
SEG & $s=2.0$ & 11 & 58.68 & 0.724 & 0.645 & 0.867 & 0.125 & 29.05 \\
SEG & $s=3.0$ & 3 & 60.37 & 0.786 & 0.580 & 0.919 & 0.137 & 29.93 \\
SEG & $s=3.0$ & 5 & 61.10 & 0.786 & 0.581 & 0.919 & 0.137 & 29.99 \\
SEG & $s=3.0$ & 7 & 60.96 & 0.769 & 0.592 & 0.866 & 0.127 & 29.86 \\
SEG & $s=3.0$ & 9 & 60.63 & 0.745 & 0.605 & 0.822 & 0.122 & 29.47 \\
SEG & $s=3.0$ & 11 & 61.18 & 0.703 & 0.627 & 0.695 & 0.107 & 28.71 \\
SEG & $s=4.0$ & 3 & 62.69 & 0.742 & 0.560 & 0.841 & 0.124 & 29.43 \\
SEG & $s=4.0$ & 5 & 63.05 & 0.728 & 0.565 & 0.834 & 0.122 & 29.54 \\
SEG & $s=4.0$ & 7 & 62.45 & 0.717 & 0.573 & 0.722 & 0.111 & 29.39 \\
SEG & $s=4.0$ & 9 & 61.96 & 0.690 & 0.611 & 0.715 & 0.104 & 28.99 \\
SEG & $s=4.0$ & 11 & 63.66 & 0.641 & 0.624 & 0.643 & 0.095 & 28.18 \\
SEG & $s=6.0$ & 3 & 70.47 & 0.640 & 0.537 & 0.565 & 0.088 & 28.44 \\
SEG & $s=6.0$ & 5 & 69.50 & 0.637 & 0.556 & 0.585 & 0.088 & 28.55 \\
SEG & $s=6.0$ & 7 & 68.33 & 0.637 & 0.560 & 0.498 & 0.077 & 28.40 \\
SEG & $s=6.0$ & 9 & 69.77 & 0.603 & 0.616 & 0.487 & 0.072 & 27.83 \\
SEG & $s=6.0$ & 11 & 73.00 & 0.557 & 0.616 & 0.485 & 0.074 & 26.87 \\
\midrule
\multicolumn{9}{l}{\textit{SAG}} \\
\midrule
SAG & $s=0.1$ & 3 & 54.36 & 0.874 & 0.706 & 1.331 & 0.187 & 28.66 \\
SAG & $s=0.1$ & 5 & 56.74 & 0.857 & 0.683 & 1.211 & 0.169 & 29.24 \\
SAG & $s=0.1$ & 7 & 56.28 & 0.841 & 0.673 & 1.143 & 0.162 & 29.23 \\
SAG & $s=0.1$ & 9 & 56.74 & 0.803 & 0.680 & 0.952 & 0.139 & 28.94 \\
SAG & $s=0.1$ & 11 & 56.48 & 0.765 & 0.673 & 0.861 & 0.129 & 28.37 \\
SAG & $s=0.25$ & 3 & 54.49 & 0.866 & 0.702 & 1.349 & 0.189 & 28.86 \\
SAG & $s=0.25$ & 5 & 56.85 & 0.855 & 0.676 & 1.236 & 0.172 & 29.35 \\
SAG & $s=0.25$ & 7 & 56.43 & 0.840 & 0.670 & 1.159 & 0.161 & 29.28 \\
SAG & $s=0.25$ & 9 & 56.98 & 0.810 & 0.676 & 0.949 & 0.137 & 28.94 \\
SAG & $s=0.25$ & 11 & 57.04 & 0.756 & 0.677 & 0.859 & 0.128 & 28.33 \\
SAG & $s=0.5$ & 3 & 54.90 & 0.858 & 0.696 & 1.322 & 0.184 & 29.02 \\
SAG & $s=0.5$ & 5 & 57.38 & 0.837 & 0.667 & 1.223 & 0.169 & 29.40 \\
SAG & $s=0.5$ & 7 & 56.59 & 0.810 & 0.669 & 1.122 & 0.157 & 29.22 \\
SAG & $s=0.5$ & 9 & 57.28 & 0.775 & 0.664 & 0.924 & 0.134 & 28.84 \\
SAG & $s=0.5$ & 11 & 57.69 & 0.763 & 0.659 & 0.845 & 0.123 & 28.19 \\
SAG & $s=0.75$ & 3 & 55.16 & 0.850 & 0.687 & 1.265 & 0.177 & 29.03 \\
SAG & $s=0.75$ & 5 & 57.44 & 0.837 & 0.663 & 1.177 & 0.163 & 29.31 \\
SAG & $s=0.75$ & 7 & 57.14 & 0.808 & 0.659 & 1.065 & 0.149 & 29.07 \\
SAG & $s=0.75$ & 9 & 57.43 & 0.789 & 0.667 & 0.901 & 0.129 & 28.69 \\
SAG & $s=0.75$ & 11 & 58.13 & 0.750 & 0.657 & 0.794 & 0.117 & 28.05 \\
SAG & $s=1.0$ & 3 & 55.62 & 0.836 & 0.680 & 1.226 & 0.171 & 28.94 \\
SAG & $s=1.0$ & 5 & 57.72 & 0.820 & 0.653 & 1.104 & 0.156 & 29.16 \\
SAG & $s=1.0$ & 7 & 57.78 & 0.799 & 0.650 & 0.987 & 0.138 & 28.89 \\
SAG & $s=1.0$ & 9 & 57.81 & 0.775 & 0.661 & 0.849 & 0.124 & 28.50 \\
SAG & $s=1.0$ & 11 & 58.79 & 0.723 & 0.661 & 0.731 & 0.111 & 27.88 \\
\midrule
\multicolumn{9}{l}{\textit{EMAG (Ours)}} \\
\midrule
EMAG & $w_e=0.5$ & 3 & 52.64 & 0.824 & 0.713 & 1.197 & 0.176 & 26.29 \\
EMAG & $w_e=0.5$ & 5 & 55.43 & 0.848 & 0.705 & 1.093 & 0.159 & 27.55 \\
EMAG & $w_e=0.5$ & 7 & 55.72 & 0.827 & 0.693 & 1.076 & 0.155 & 27.90 \\
EMAG & $w_e=0.5$ & 9 & 56.85 & 0.808 & 0.665 & 1.026 & 0.147 & 27.89 \\
EMAG & $w_e=0.5$ & 11 & 57.40 & 0.768 & 0.674 & 0.840 & 0.128 & 27.57 \\
EMAG & $w_e=1.0$ & 3 & 52.53 & 0.862 & 0.724 & 1.312 & 0.192 & 27.80 \\
EMAG & $w_e=1.0$ & 5 & 55.46 & 0.851 & 0.693 & 1.159 & 0.165 & 28.93 \\
EMAG & $w_e=1.0$ & 7 & 55.70 & 0.845 & 0.694 & 1.171 & 0.167 & 29.16 \\
EMAG & $w_e=1.0$ & 9 & 56.33 & 0.811 & 0.684 & 1.088 & 0.153 & 29.11 \\
EMAG & $w_e=1.0$ & 11 & 57.20 & 0.780 & 0.672 & 0.875 & 0.131 & 28.72 \\
EMAG & $w_e=1.5$ & 3 & 53.08 & 0.867 & 0.711 & 1.333 & 0.193 & 28.77 \\
EMAG & $w_e=1.5$ & 5 & 55.69 & 0.838 & 0.668 & 1.221 & 0.169 & 29.67 \\
EMAG & $w_e=1.5$ & 7 & 56.98 & 0.843 & 0.681 & 1.085 & 0.155 & 29.50 \\
EMAG & $w_e=1.5$ & 9 & 56.41 & 0.787 & 0.658 & 1.034 & 0.146 & 29.35 \\
EMAG & $w_e=1.5$ & 11 & 57.39 & 0.749 & 0.657 & 0.815 & 0.122 & 28.67 \\
EMAG & $w_e=2.0$ & 3 & 53.37 & 0.854 & 0.690 & 1.283 & 0.186 & 29.28 \\
EMAG & $w_e=2.0$ & 5 & 55.91 & 0.843 & 0.649 & 1.161 & 0.162 & 29.86 \\
EMAG & $w_e=2.0$ & 7 & 57.33 & 0.788 & 0.635 & 0.979 & 0.139 & 29.40 \\
EMAG & $w_e=2.0$ & 9 & 60.44 & 0.728 & 0.608 & 0.739 & 0.111 & 28.64 \\
EMAG & $w_e=2.0$ & 11 & 67.45 & 0.624 & 0.575 & 0.528 & 0.081 & 27.71 \\
\midrule
\multicolumn{9}{l}{\textit{ERG}} \\
\midrule
ERG & -- & 2 & 51.49 & 0.847 & 0.718 & 1.177 & 0.178 & 28.47 \\
ERG & -- & 3 & 57.51 & 0.814 & 0.644 & 0.999 & 0.148 & 29.63 \\
ERG & -- & 4 & 61.02 & 0.777 & 0.602 & 0.879 & 0.135 & 29.62 \\
ERG & -- & 5 & 64.21 & 0.730 & 0.581 & 0.779 & 0.117 & 29.31 \\
ERG & -- & 9 & 83.48 & 0.520 & 0.482 & 0.388 & 0.059 & 27.66 \\
ERG & -- & 11 & 92.87 & 0.462 & 0.473 & 0.295 & 0.045 & 26.82 \\
\midrule
\multicolumn{9}{l}{\textit{PAG}} \\
\midrule
PAG & $s=1.0$ & 3 & 60.01 & 0.817 & 0.640 & 1.130 & 0.158 & 28.70 \\
PAG & $s=1.0$ & 5 & 58.05 & 0.805 & 0.645 & 1.102 & 0.157 & 28.84 \\
PAG & $s=1.0$ & 7 & 56.98 & 0.713 & 0.680 & 0.806 & 0.122 & 27.32 \\
PAG & $s=1.0$ & 9 & 66.10 & 0.557 & 0.717 & 0.537 & 0.084 & 25.01 \\
PAG & $s=1.0$ & 11 & 83.14 & 0.471 & 0.716 & 0.321 & 0.051 & 22.71 \\
PAG & $s=2.0$ & 3 & 60.85 & 0.783 & 0.632 & 0.963 & 0.136 & 28.40 \\
PAG & $s=2.0$ & 5 & 59.29 & 0.675 & 0.674 & 0.736 & 0.108 & 26.75 \\
PAG & $s=2.0$ & 7 & 81.86 & 0.478 & 0.693 & 0.348 & 0.056 & 23.09 \\
PAG & $s=2.0$ & 9 & 113.00 & 0.335 & 0.706 & 0.135 & 0.019 & 20.31 \\
PAG & $s=2.0$ & 11 & 138.57 & 0.274 & 0.587 & 0.091 & 0.012 & 18.61 \\
PAG & $s=3.0$ & 3 & 61.03 & 0.711 & 0.627 & 0.753 & 0.112 & 27.49 \\
PAG & $s=3.0$ & 5 & 78.48 & 0.476 & 0.685 & 0.361 & 0.057 & 23.58 \\
PAG & $s=3.0$ & 7 & 120.27 & 0.322 & 0.645 & 0.137 & 0.019 & 19.80 \\
PAG & $s=3.0$ & 9 & 151.46 & 0.243 & 0.495 & 0.074 & 0.008 & 17.81 \\
PAG & $s=3.0$ & 11 & 175.97 & 0.190 & 0.273 & 0.042 & 0.003 & 16.54 \\
\midrule
\multicolumn{9}{l}{\textit{S2}} \\
\midrule
S2 & $w=0.10$ & 3 & 56.42 & 0.799 & 0.696 & 0.985 & 0.148 & 27.35 \\
S2 & $w=0.10$ & 5 & 54.85 & 0.837 & 0.700 & 1.245 & 0.177 & 28.84 \\
S2 & $w=0.10$ & 7 & 54.21 & 0.839 & 0.711 & 1.304 & 0.183 & 29.40 \\
S2 & $w=0.10$ & 9 & 53.44 & 0.847 & 0.711 & 1.232 & 0.172 & 29.41 \\
S2 & $w=0.10$ & 11 & 52.99 & 0.829 & 0.717 & 1.159 & 0.167 & 29.13 \\
S2 & $w=0.15$ & 3 & 64.30 & 0.705 & 0.626 & 0.671 & 0.104 & 26.70 \\
S2 & $w=0.15$ & 5 & 57.48 & 0.806 & 0.679 & 1.005 & 0.145 & 28.40 \\
S2 & $w=0.15$ & 7 & 54.66 & 0.821 & 0.702 & 1.209 & 0.173 & 29.15 \\
S2 & $w=0.15$ & 9 & 54.68 & 0.831 & 0.709 & 1.202 & 0.167 & 29.29 \\
S2 & $w=0.15$ & 11 & 52.86 & 0.819 & 0.714 & 1.093 & 0.160 & 29.13 \\
S2 & $w=0.20$ & 7 & 56.95 & 0.790 & 0.682 & 1.023 & 0.148 & 28.80 \\
S2 & $w=0.25$ & 3 & 84.68 & 0.475 & 0.527 & 0.300 & 0.048 & 25.20 \\
S2 & $w=0.25$ & 5 & 67.65 & 0.679 & 0.588 & 0.653 & 0.098 & 27.38 \\
S2 & $w=0.25$ & 7 & 60.44 & 0.753 & 0.650 & 0.856 & 0.128 & 28.40 \\
S2 & $w=0.25$ & 9 & 58.22 & 0.753 & 0.665 & 0.891 & 0.128 & 28.68 \\
S2 & $w=0.25$ & 11 & 55.90 & 0.770 & 0.692 & 0.902 & 0.135 & 28.68 \\
S2 & $w=0.30$ & 3 & 94.21 & 0.401 & 0.501 & 0.219 & 0.035 & 24.48 \\
S2 & $w=0.30$ & 5 & 74.04 & 0.617 & 0.557 & 0.487 & 0.076 & 26.86 \\
S2 & $w=0.30$ & 7 & 64.73 & 0.694 & 0.602 & 0.679 & 0.104 & 28.00 \\
S2 & $w=0.30$ & 9 & 61.62 & 0.704 & 0.627 & 0.730 & 0.109 & 28.33 \\
S2 & $w=0.30$ & 11 & 59.16 & 0.732 & 0.655 & 0.764 & 0.116 & 28.33 \\
\end{longtable}
}

\section{Additional implementation details}
\label{sec:additional_implementation_details}

\subsection{EMAG-I vs.\ EMAG}
For MMDiT-based diffusion transformers, at a given timestep $t$, we partition the per-head attention over image queries into
image-to-image and image-to-text blocks:
\begin{equation}
    A_t^{\mathrm{img}}
    \;=\;
    \big[\,A_t^{\mathrm{ii}} \;\big|\; A_t^{\mathrm{it}}\,\big],
\end{equation}
where $A_t^{\mathrm{ii}} \in \mathbb{R}^{Q_{\mathrm{img}} \times K_{\mathrm{img}}}$ denotes image$\to$image attention
and $A_t^{\mathrm{it}} \in \mathbb{R}^{Q_{\mathrm{img}} \times K_{\mathrm{txt}}}$ denotes image$\to$text attention.

\paragraph{EMAG-I (image-to-image only).}
EMAG-I maintains an EMA only over the image-to-image block and perturbs that block:
\begin{align}
    E_t^{\mathrm{ii}}
    &= \beta\,E_{t-1}^{\mathrm{ii}} + (1-\beta)\,A_t^{\mathrm{ii}}, \\
    \widetilde{A}_t^{\mathrm{ii}}
    &= (1-\lambda)\,A_t^{\mathrm{ii}} + \lambda\,E_t^{\mathrm{ii}}.
\end{align}

\paragraph{EMAG (image-to-image + image-to-text).}
For EMAG, we maintain EMA over the full image-query row (image-to-image + image-to-text):
\begin{align}
    E_t^{\mathrm{img}}
    &= \beta\,E_{t-1}^{\mathrm{img}} + (1-\beta)\,A_t^{\mathrm{img}} \\
    \widetilde{A}_t^{\mathrm{img}}
    &= (1-\lambda)\,A_t^{\mathrm{img}} + \lambda\,E_t^{\mathrm{img}}
\end{align}

In the case of DiT-based architectures, EMAG-I is our default mode; throughout this work, any reference to ``EMAG'' in the context of DiT-based diffusion transformers corresponds to EMAG-I.

\begin{algorithm}[t]
\caption{EMAG: (Unconditional)}
\label{alg:EMAG-uncond}
\begin{algorithmic}[1]
\REQUIRE $\tau_s$: starting timestep for EMAG
\REQUIRE $\tau_e$: end timestep for EMAG
\REQUIRE $\epsilon_\theta$: Denoiser network
\REQUIRE $\epsilon_\theta^{'}$: Perturbed Denoiser network
\REQUIRE $x_t$: input at timestep $t$
\REQUIRE $w_{e}$: EMAG Scale
\STATE $x_T \sim \mathcal{N}(0,I)$
\FOR{$t = T, T-1, \ldots, 1$}
    \STATE $z_t \leftarrow \epsilon_\theta(x_t)$
    \IF{$\tau_e < t < \tau_s$}
        \STATE $\hat{z}_t \leftarrow \epsilon'_\theta(x_t)$ \COMMENT{Eq.~\eqref{eq: layer_selection}, \eqref{eq: L1_norm_calcuation}, \eqref{eq: EMA replacement}}
        \STATE $\bar{z}_t \leftarrow \hat{z}_t + w_{e}\cdot(z_t - \hat{z}_t)$ \COMMENT{Eq.~\eqref{eq:EMAG_uncond_update}}
    \ELSE
        \STATE $\bar{z}_t \leftarrow z_t$
    \ENDIF
\ENDFOR
\RETURN $\bar{z}_t$
\end{algorithmic}
\end{algorithm}

\subsection{Implementation details for baselines}

For APG and CADS, we follow the authors' reference pseudocode and adapt it to both DiT and SD3 codebases. For ERG, we use the official reference implementation and plug it into DiT and SD3 in the same way. S2 is reimplemented directly from the algorithm provided in the original paper. SEG is adapted from the official Stable Diffusion XL (SDXL) implementation, replacing the UNet attention processor with the joint-attention processor used by SD3-medium. For UNet-based approaches such as SAG and PAG, we adapt them to DiT and SD3 following the recommendations in \cite{ifriqi2025entropy}. PAG is used only for the DiT unconditional setting.

We now summarize the main implementation choices for SAG, SEG, and PAG.

\paragraph{SAG}
For SAG, we follow the authors’ SDXL implementation and adapt it to both DiT and SD3. We compute a global average pooling over the selected attention maps to obtain a spatial mask, threshold this mask at $1$ as in the original paper, and then apply a Gaussian blur with kernel size $9$. The SAG guidance scales for conditional and unconditional generation are chosen according to the ablations reported in Tables~\ref{tab:guidance_methods_comparison_full},\ref{tab:sd3_uncond_sag_seg_ierg},
\ref{tab:dit_cond_sag_seg_s2_erg_apg_cads},and~\ref{tab:dit_uncond_pag_sag_seg}. For DiT, we apply SAG at block~\textbf{13}, mirroring the original setup. For SD3 (MMDiT), following the ERG paper’s recommendation that blocks 6--8 are comparable to the UNet middle block, we apply SAG at layer~\textbf{7} to construct the SAG mask.

\paragraph{SEG}
For SEG, we start from the official SDXL implementation and adapt it to SD3-medium and DiT. In SD3, we adapt the UNet \texttt{AttentionProcessor2\_0} to the joint-attention processor used by the MMDiT backbone. In the original formulation, SEG operates on the unconditional branch; we preserve this behavior in the CFG setting by manipulating only the unconditional path when combining SEG with CFG for conditional generation. Throughout the paper, SEG+CFG is treated as the default SEG configuration for conditional setups. Following the recommendations in \cite{ifriqi2025entropy}, where SEG is applied at the UNet middle block, we target the corresponding transformer layers: blocks 6--8 in SD3 and blocks 12--15 in DiT. In all cases, we perturb only the image-to-image portion of the attention map, leaving image-to-text entries unchanged when applicable.

\paragraph{PAG}
\label{sec:pag_implementation}

We adapt Perturbed-Attention Guidance (PAG) to SD3's MMDiT architecture following the joint-attention adaptation strategy described in the \cite{ifriqi2025entropy}, which provides the closest existing reference for applying self-attention perturbation methods to fused image-text attention blocks. Specifically, we replace only the image-to-image sub-block of the joint attention matrix with an identity matrix while preserving the image-to-text and text-to-text attention blocks, then use the difference between normal and degraded predictions as a guidance signal following the standard PAG formulation. We apply this perturbation to transformer blocks 6–7–8 (the middle layers), consistent with the authors' recommendation and also discussed in \cite{ifriqi2025entropy} for targeting mid-network layers where spatial structure is most concentrated. We run a sweep across CFG scales and report all configurations in Table~\ref{tab:guidance_methods_comparison_full} and \ref{tab:pag_pareto_sweep}. We note that adapting PAG to joint attention architectures is inherently more difficult than to UNet-based models, as the self-attention perturbation unavoidably interacts with text conditioning within the fused attention operation; this architectural mismatch may partially account for the reduced effectiveness observed relative to UNet-based results reported in the original work. For DiT, we follow the authors' unconditional configuration directly. 

\begin{figure*}[t]
  \centering
  \includegraphics[width=\linewidth]{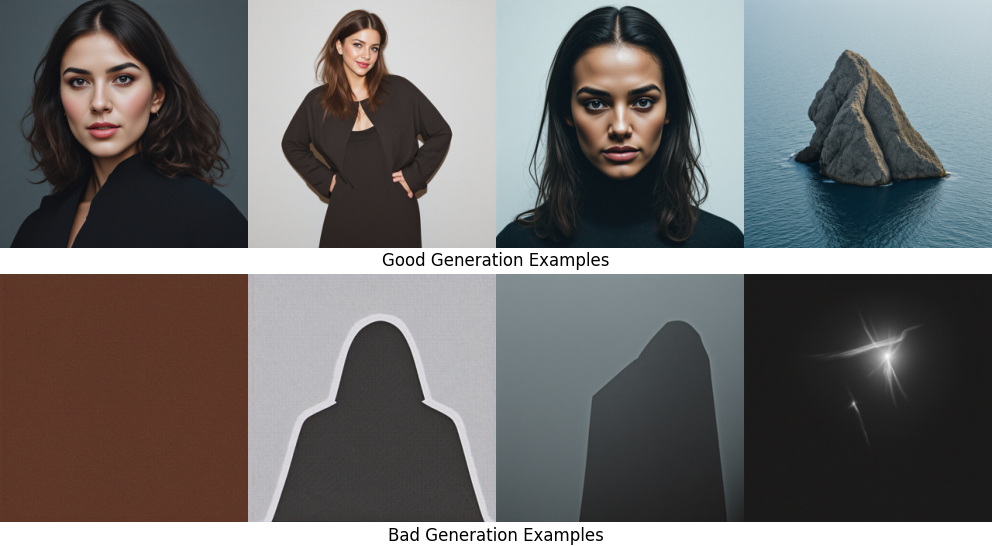}
      \caption{\textbf{Qualitative samples for SD3 PAG implementation for PAG Scale=1}}

    \label{fig:section_D_uncond_pag_analysis}
\end{figure*}

\paragraph{Architectural limitations of PAG on joint-attention models.}
On further analysis (Table~\ref{tab:sd3_uncond_sag_seg_ierg}), we find that PAG's identity perturbation on image tokens inadvertently disrupts text conditioning due to SD3's joint attention design. Since image and text tokens attend to each other within shared attention layers, perturbing the image-token attention inevitably corrupts the conditional signal as well — even though only image tokens are directly targeted. This leads to generation collapse even at lower scales. Figure~\ref{fig:section_D_uncond_pag_analysis} illustrates this: while some unconditional generations retain acceptable quality, the majority exhibit severe degradation, and the effect is equally pronounced in conditional generation. A similar architectural incompatibility for PAG on DiT-based models, including SD3, has been independently reported by Rajabi et al.~\cite{rajabi2025token} and Ahn et al.~\cite{ahn2025and}. A potential design choice is to perturb the entire joint attention or perform re-normalization; however, this is out of the scope for our discussion, as it builds on the original implementation with multiple fixes possible. 

\textbf{Note.} For simplicity, all DiT experiments use the DDIM sampler, while all SD3 experiments use the FlowMatchEulerDiscrete sampler. For HPS, we randomly subsample 1{,}000 images (without replacement, using a fixed random seed \textbf{8} ) from the pool of generated samples for each method and report the mean HPS over these images.

\begin{table*}[t]
\centering
\small
\setlength{\tabcolsep}{6pt}
\caption{EMAG hyperparameter configurations for each task and backbone considered in our experiments.}
\label{tab:emag_hparams}
\begin{tabular}{lccccccc}
\toprule
\textbf{Task} & $w_e$ & $\tau_{s}$ & $\tau_{e}$ & $\delta_{t}$ & $\beta$ & $l_{min}$ & $l_{max}$ \\
\midrule
T2I (SD3-Medium, MMDiT)               & 1.5  & 28  & 5 & 5 & 0.988 & 6 & 8  \\
\midrule
Unconditional T2I (SD3-Medium)        & 5.125 & 28 & 5 & 5 & 0.988 & 6  & 8 \\
\midrule
Class-conditional (DiT-XL/2)          & 1.5 & 250  & 50 & 50 & 0.988 & 12 & 15 \\
\midrule
Unconditional (DiT-XL/2)              & 7  & 250 & 50 & 50  & 0.988 & 12 & 15 \\
\bottomrule
\end{tabular}
\end{table*}


\section{Additional qualitative results}
\label{sec:additional_qualitative_results}

\begin{figure*}[t]
  \centering
  \includegraphics[width=\linewidth]{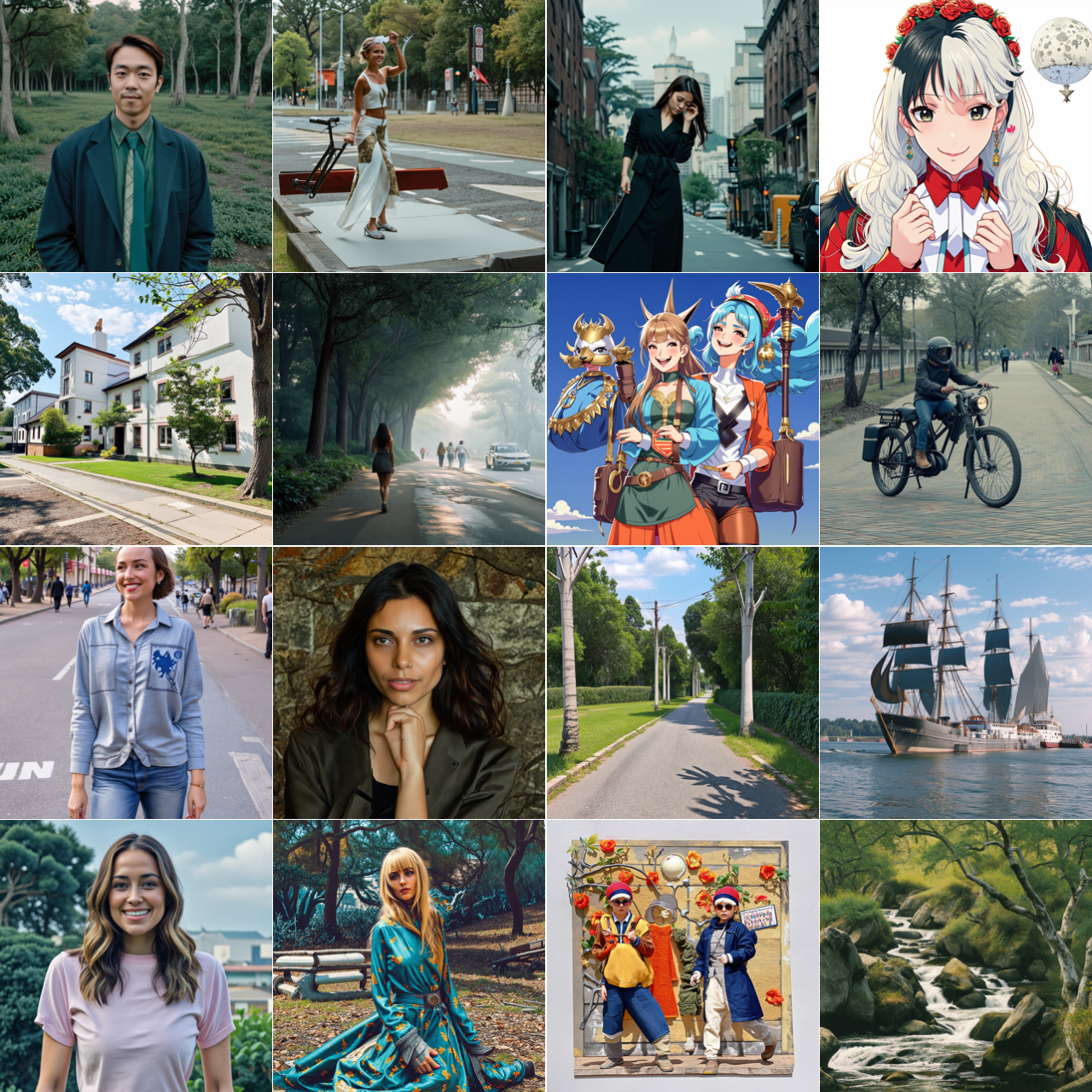}
      \caption{\textbf{Additional qualitative samples for unconditional EMAG with $ w_{emag} = 5.125 $} on SD3-Medium}
    \label{fig: section_D_uncond_sd3_EMAG}
\end{figure*}

\begin{figure*}[t]
  \centering
  \includegraphics[width=\linewidth]{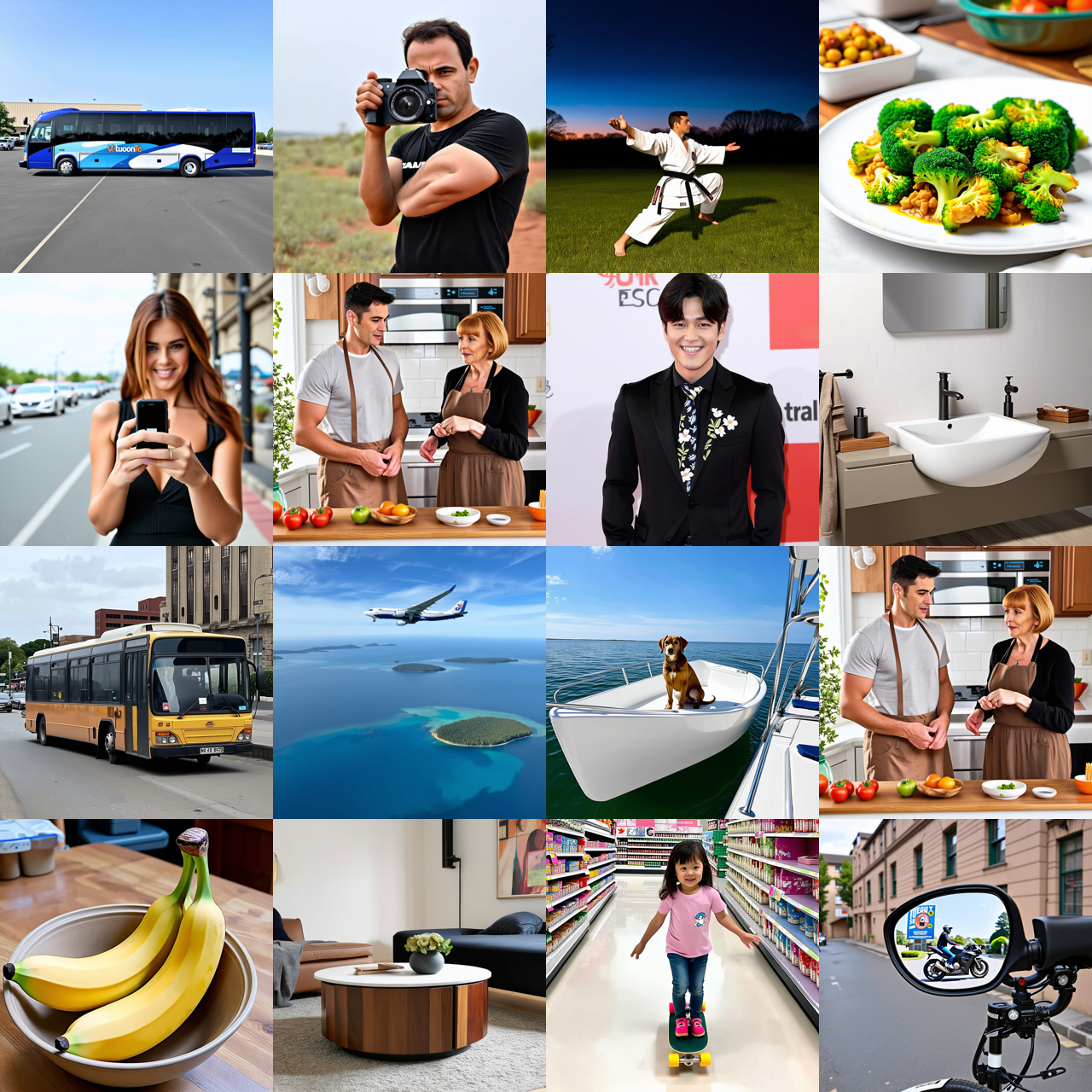}
      \caption{\textbf{Additional qualitative samples for conditional Text to Image generation using SD3 with EMAG with $ w_{emag} = 1.5 $}}

    \label{fig: section_D_cond_sd3_EMAG}
\end{figure*}

\begin{figure*}[t]
  \centering
  \includegraphics[width=\linewidth]{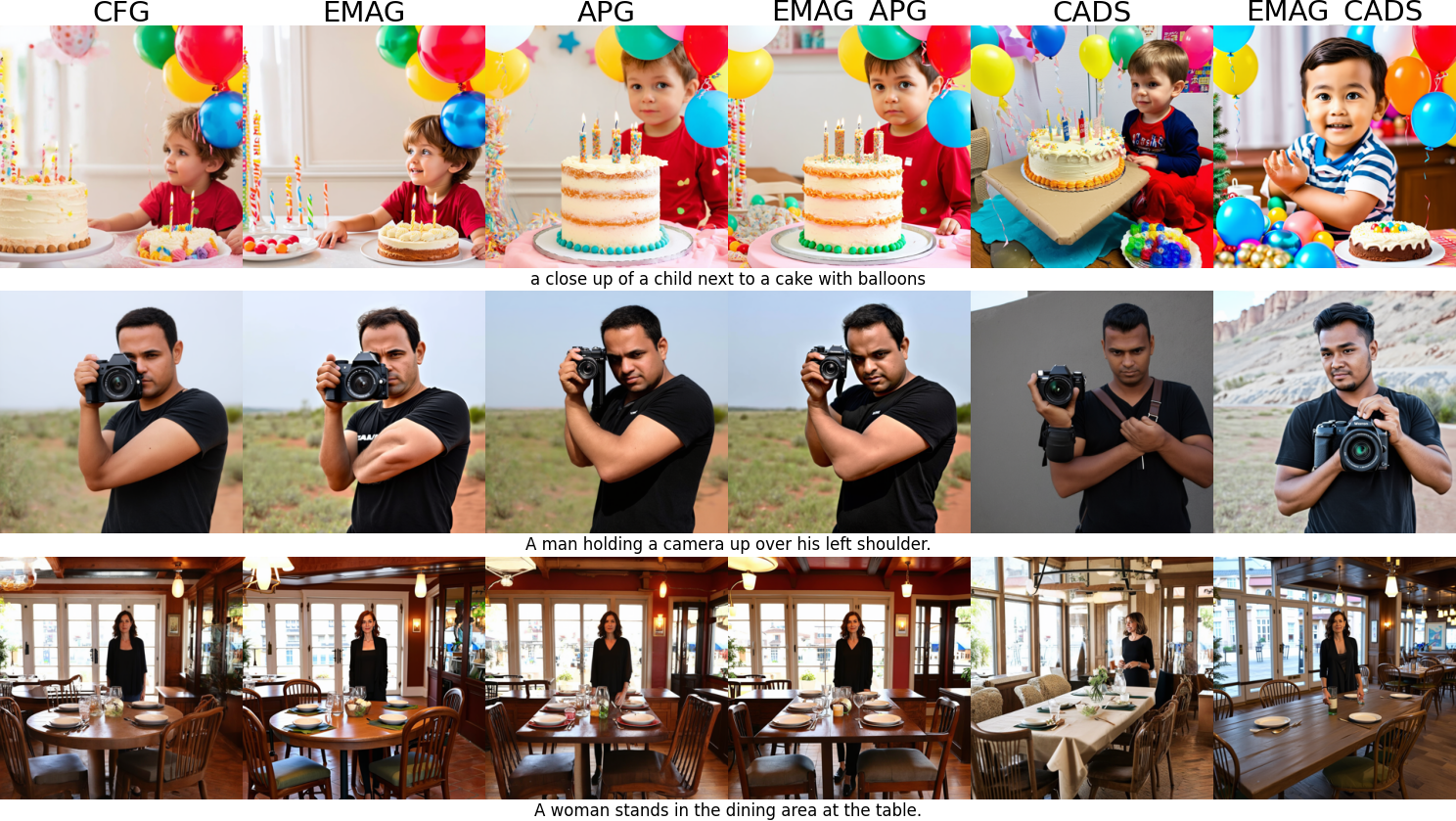}
      \caption{\textbf{Additional qualitative comparison of conditional text-to-image generation with SD3 using EMAG, APG, CADS, and their combinations.}}

    \label{fig: section_D_cond_APG_CADS_with_EMAG}
\end{figure*}

\begin{figure*}[t]
  \centering
  \includegraphics[width=\linewidth]{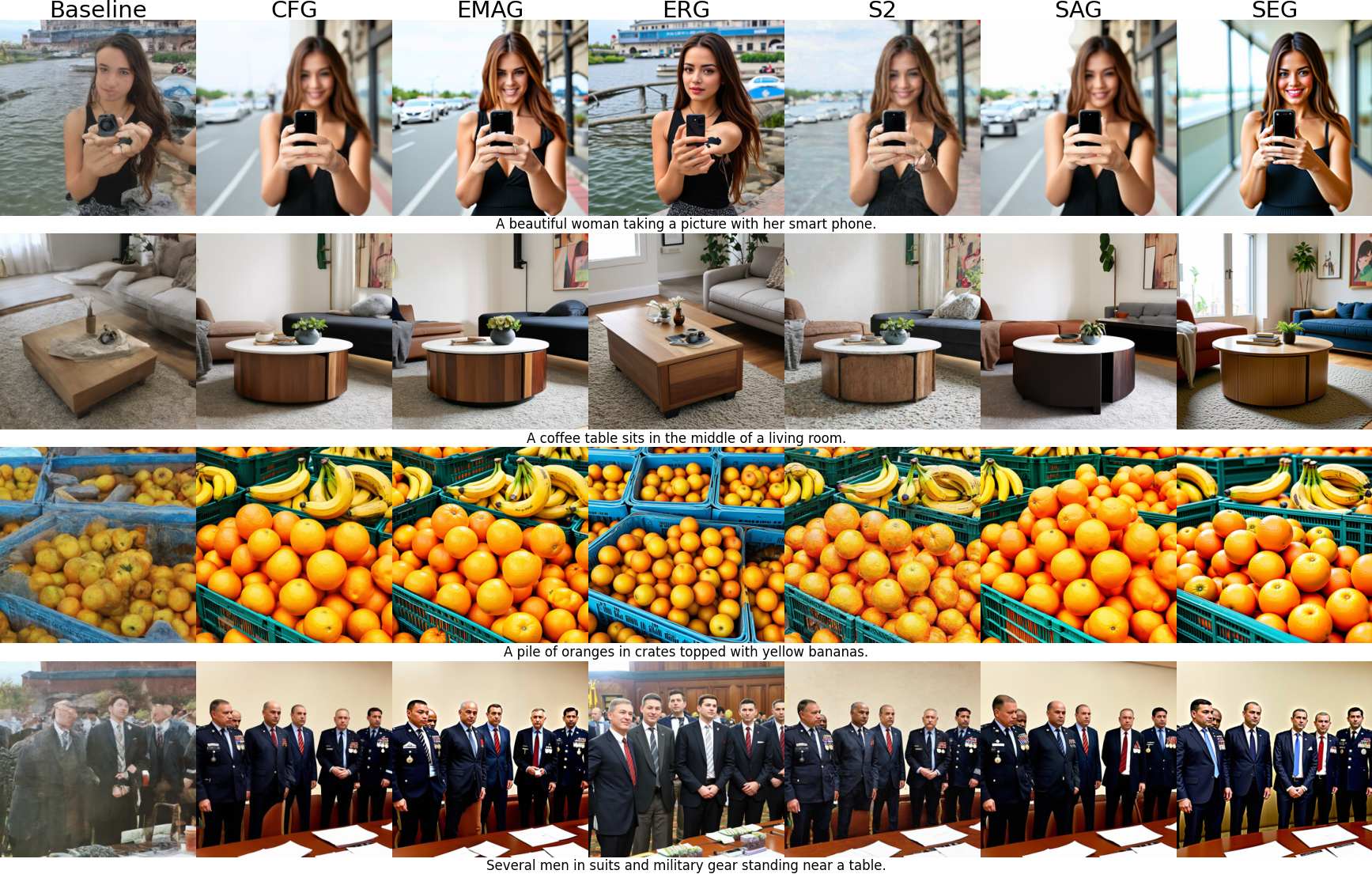}
      \caption{\textbf{Additional qualitative comparison of EMAG and other baseline guidance methods for conditional text-to-image generation with SD3.}}

    \label{fig: section_D_baseline_vs_emag}
\end{figure*}

\begin{figure*}[t]
  \centering
  \includegraphics[width=\linewidth]{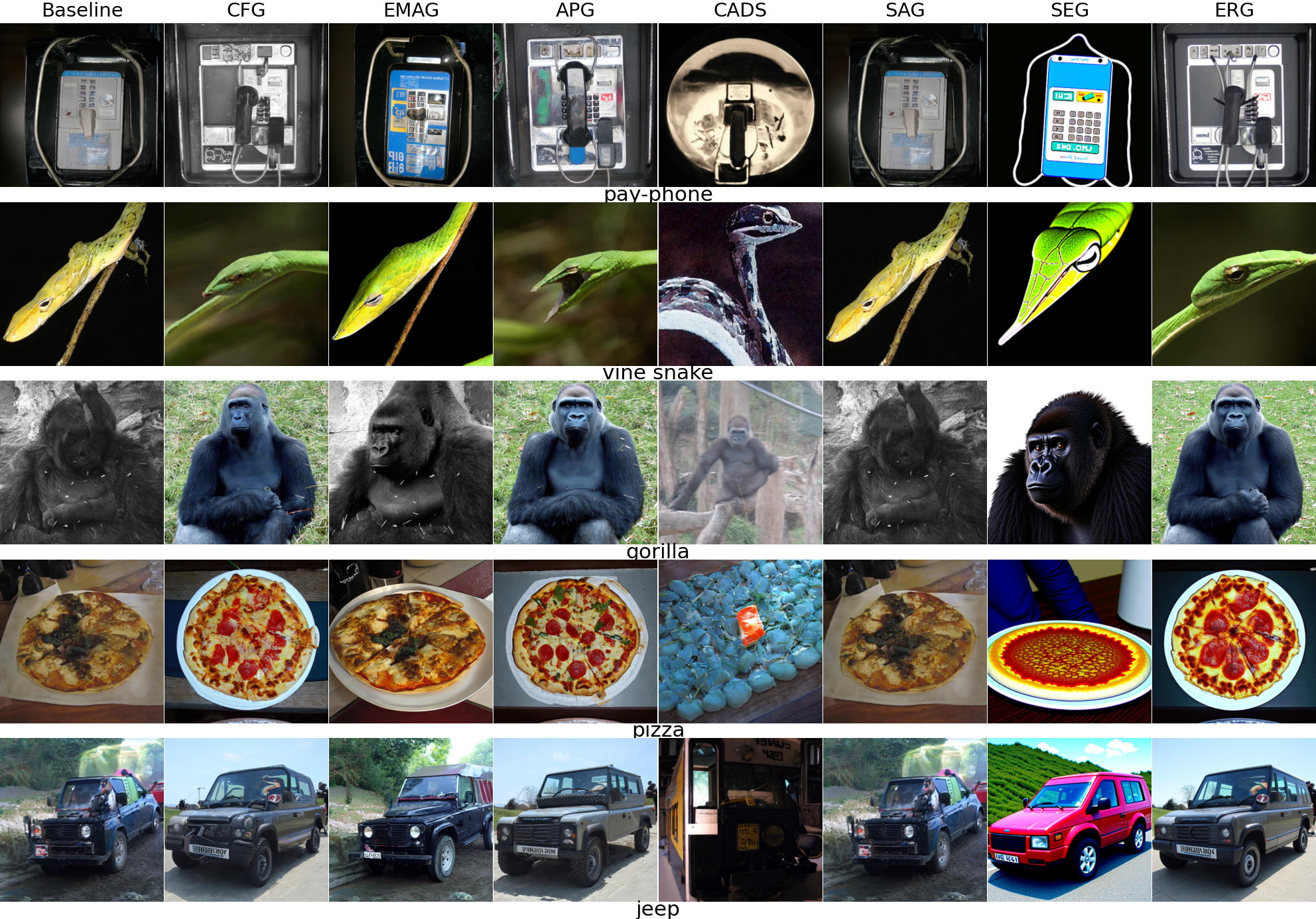}
      \caption{\textbf{Additional qualitative comparison of EMAG and other baseline guidance methods for class conditional generation with DiT-XL/2 512x512.}}

    \label{fig: section_D_baseline_vs_emag_512_dit}
\end{figure*}

\begin{figure*}[t]
  \centering
  \includegraphics[width=\linewidth]{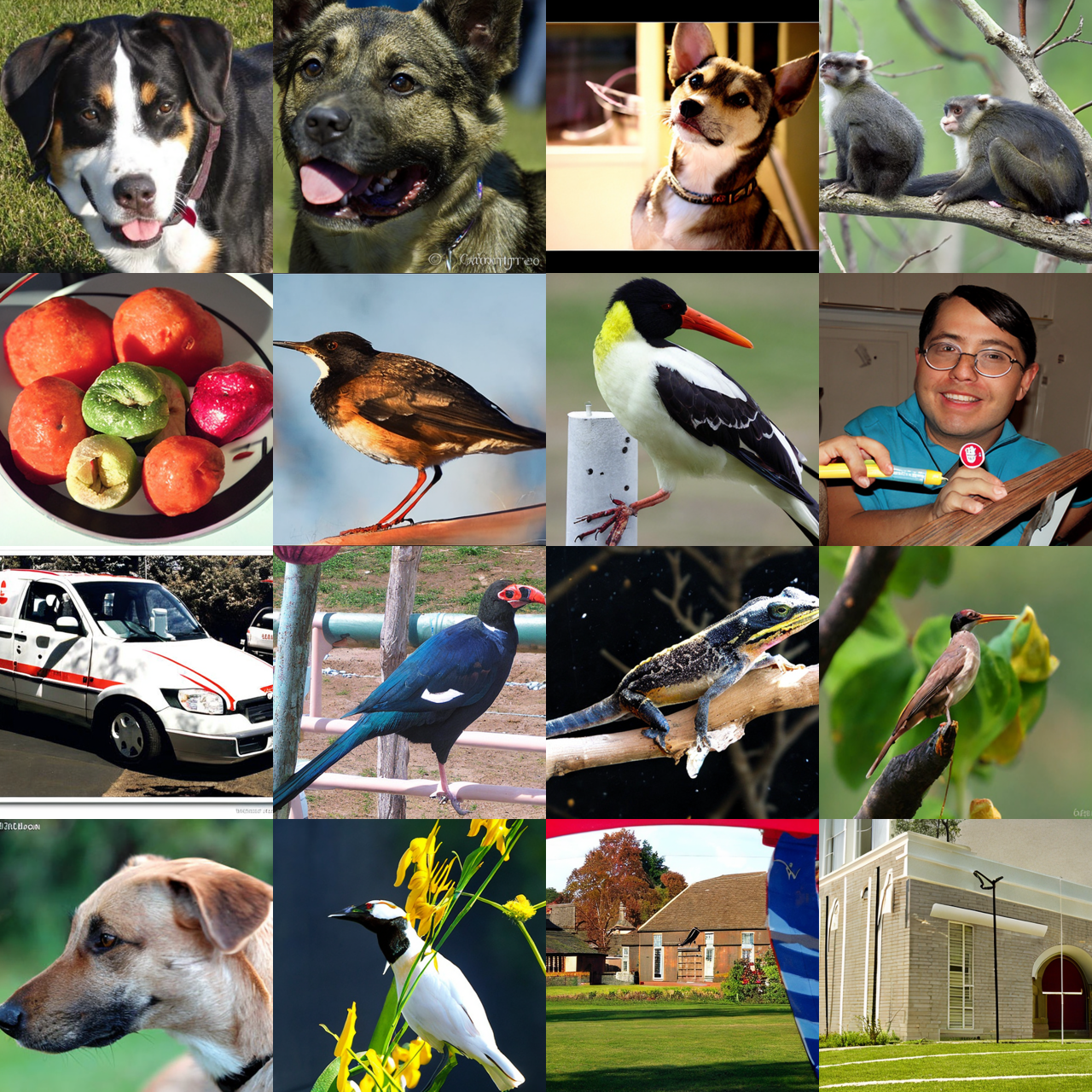}
      \caption{\textbf{Qualitative samples for unconditional EMAG with $ w_{emag} = 7 $} for DIT-XL/2 512x512 }

    \label{fig: section_D_uncond_sd3_dit_512}
\end{figure*}

In this section, we provide additional qualitative examples. 
Figures~\ref{fig: section_D_uncond_sd3_EMAG} and~\ref{fig: section_D_cond_sd3_EMAG} show unconditional and conditional samples from EMAG on SD3, respectively, while Fig.~\ref{fig: section_D_uncond_sd3_dit_512} presents unconditional generations from DiT-XL/2 at 512$\times$512 resolution. 
Fig.~\ref{fig: section_D_baseline_vs_emag} provides a qualitative comparison between EMAG and other guidance baselines on SD3, and Fig.~\ref{fig: section_D_cond_APG_CADS_with_EMAG} compares EMAG to orthogonal guidance strategies such as CFG, APG, and CADS, as well as their combinations with EMAG. 
Finally, Fig.~\ref{fig: section_D_baseline_vs_emag_512_dit} reports qualitative comparisons for DiT-XL/2 at 512$\times$512 resolution, contrasting EMAG-based guidance with competing baselines for class-conditional generation. All prompts used for the qualitative figures are listed in Table~\ref{tab:qual_prompts}.

\section{EMAG Computational Analysis and EMAG-Q}
\label{sec:EMAG-Q}

As shown in Tables~\ref{tab:timing_benchmark} and~\ref{tab:emag_q_comparison}, EMAG incurs significant computational overhead (+213\%) compared to other guidance methods, primarily due to the explicit tracking and renormalization of post-softmax attention maps, which prevents the use of fused SDPA kernels. While this precise attention map EMA is central to regulating hard negative quality, it creates a bottleneck in practice.

To address this, we introduce \textbf{EMAG-Q}, an efficient variant that maintains the EMA over pre-softmax query embeddings rather than post-softmax attention maps. By perturbing the query embeddings directly, EMAG-Q enables the use of fused SDPA kernels, reducing overhead from +213\% to +95\% and peak memory from 29.58\,GB to 16.96\,GB—closely matching the memory footprint of vanilla CFG. As shown in Table~\ref{tab:emag_q_comparison}, all generation quality metrics remain within marginal differences ($\Delta$FID\,=\,+0.63, $\Delta$HPS\,=\,$-$0.12).

To further validate that EMAG-Q faithfully approximates EMAG, we compute per-image Spearman rank correlation (SRCC) on paired outputs. The EMAG vs.\ EMAG-Q correlation ($\rho$\,=\,0.741) closely matches the inter-seed correlation of EMAG with itself ($\rho$\,=\,0.754), indicating that the approximation error introduced by query-space EMA falls within the inherent stochastic variation of the diffusion process.

To verify that EMAG-Q preserves the hard negative generation capability of EMAG, 
we compare the cosine similarity between positive and negative samples produced 
by both variants. As shown in Fig.~\ref{fig: section_D_EMAG-Q-VS-EMAG_bars}, 
the similarity scores are nearly identical across both CLIP-I 
($0.753$ vs.\ $0.741$) and DINOv2 ($0.548$ vs.\ $0.549$), confirming that 
query-space EMA produces comparable degradations to attention-map EMA. 
A qualitative comparison in Fig.~\ref{fig: section_D_EMAG-Q-VS-EMAG} further 
illustrates that both variants exhibit the same style of structural distortion 
and detail loss in their hard negatives.

\subsection{EMAG-Q: Efficient Query-Space EMA}
\label{sec:Architecure discussion EMAG-Q}
\subsubsection{Motivation}


While vanilla EMAG leverages EMA as a low-pass frequency filter to generate effective hard negatives, tracking the EMA over post-softmax attention maps incurs significant overhead: the EMA buffer requires substantial storage, and replacing the attention map at each timestep precludes the use of Fused SDPA, further increasing compute time (see Table~\ref{tab:timing_benchmark}). To address this, we introduce EMAG-Q, a lightweight variant that tracks the EMA over query embedding tokens rather than post-softmax attention maps, and restricts perturbation to the query embeddings only. This preserves the desirable properties of vanilla EMAG at substantially reduced computational cost. Qualitative results (Figures~\ref{fig: section_D_EMAG-Q-VS-EMAG_bars} and~\ref{fig: section_D_EMAG-Q-VS-EMAG}) show that CLIP-I and DINO similarity scores between generated images and their negatives remain comparable to vanilla EMAG. Quantitative analysis in Table~\ref{tab:emag_q_comparison} further confirms that EMAG-Q achieves performance within an acceptable range of vanilla EMAG, with comparable HPS SRCC and PLCC across seed variations.

\subsubsection{Formulation}
The Standard EMAG has the following formulation for EMA calculation and attention smoothing. See Eq.~\ref{eq:emag_orig} or Eq.~\ref{eq:EMA update} and \ref{eq: EMA replacement}.

\begin{align}
\label{eq:emag_orig}
    & \bar{A}_t = \beta \bar{A}_{t-1} + (1-\beta) \text{softmax}(Q_t K_t^\top / \sqrt{d}) \\
    & \tilde{A}_t = (1-\lambda)\text{softmax}(Q_t K_t^\top / \sqrt{d}) + \lambda \bar{A}_t 
\end{align}

Now for the EMAG-Q, the formulation remains the same, but instead of operating on the post softmax attention map $\text{softmax}(Q_t K_t^\top / \sqrt{d})$. We operate on the $Q_t$ see Eq.~\ref{eq:emag_q_formulation}

\begin{align}
\label{eq:emag_q_formulation}
    & \bar{Q}_t = \beta \bar{Q}_{t-1} + (1-\beta) Q_t \\
    & \tilde{Q}_t = (1-\lambda) \cdot Q_{t} + \lambda \cdot \bar{Q}_t 
\end{align}

This reformulation is motivated by two observations. First, the EMA on $Q_t$ acts as a temporal low-frequency filter on the query space, which induces a smoothed attention distribution after the softmax, analogous to the direct attention smoothing in vanilla EMAG. Second, since the softmax is applied \emph{after} blending, $\tilde{A}_t$ is a valid probability distribution by construction, requiring no renormalization. Crucially, this formulation avoids materializing the full $N \times N$
attention matrix for the EMA buffer, and permits the use of Fused SDPA since the attention computation remains unmodified.

\subsubsection{Why This Works}
We now show that EMAG-Q approximates vanilla EMAG to first order.

\paragraph{Step 1: Linearity of matrix multiplication.}
Since matrix multiplication distributes over addition, blending queries before multiplying by $K_t^\top$ is equivalent to blending the resulting logits. Let $L_t = Q_t K_t^\top / \sqrt{d}$ and $\bar{L}_t = \bar{Q}_t K_t^\top / \sqrt{d}$. 
Then:
\begin{align}
\tilde{Q}_t K_t^\top / \sqrt{d} 
&= \left[(1-\lambda)\,Q_t + \lambda\,\bar{Q}_t\right] K_t^\top / \sqrt{d} \nonumber \\
&= (1-\lambda)\,L_t + \lambda\,\bar{L}_t \label{eq:logit_blend}
\end{align}

\paragraph{Step 2: Interchangeability of EMA and softmax.}
In vanilla EMAG, $\bar{A}_t$ is the EMA of post-softmax attention maps. We show that this is approximately equal to the softmax of EMA logits, 

\begin{align}
    i.e., \bar{A}_t = \mathrm{EMA}[\mathrm{softmax}(L_\tau)] \approx \mathrm{softmax}(\mathrm{EMA}[L_\tau]) = \mathrm{softmax}(\bar{L}_t).
\end{align}
Writing the EMA explicitly as a weighted sum $\bar{A}_t = \sum_\tau w_\tau\,\mathrm{softmax}(L_\tau)$ with exponentially decaying weights $\sum_\tau w_\tau = 1$, and assuming each $L_\tau$ lies close to the weighted mean $\bar{L}_t$ (i.e., $L_\tau = \bar{L}_t + \epsilon_\tau$ with $\epsilon_\tau$ small), we expand each term to first order using the softmax Jacobian $J = \mathrm{diag}(\sigma) - \sigma\sigma^\top$ where $\sigma = \mathrm{softmax}(\bar{L}_t)$:
\begin{align}
\mathrm{softmax}(L_\tau) &\approx \sigma(\bar{L}_t) + J(\bar{L}_t)\,\epsilon_\tau \label{eq:taylor_ema}
\end{align}
Taking the weighted average:
\begin{align}
\sum_\tau w_\tau\,\mathrm{softmax}(L_\tau) 
&\approx \sigma(\bar{L}_t)\underbrace{\sum_\tau w_\tau}_{=\,1} + J(\bar{L}_t)\underbrace{\sum_\tau w_\tau\,\epsilon_\tau}_{=\,0} \nonumber \\
&= \mathrm{softmax}(\bar{L}_t) \label{eq:ema_swap}
\end{align}
The residual term vanishes since $\sum_\tau w_\tau\,\epsilon_\tau = \sum_\tau w_\tau(L_\tau - \bar{L}_t) = \bar{L}_t - \bar{L}_t = 0$ by definition of the weighted mean. Hence:
\begin{align}
\bar{A}_t \approx \mathrm{softmax}(\bar{L}_t) \label{eq:abar_approx}
\end{align}

\paragraph{Step 3: First-order equivalence.}
We can now show that both formulations produce the same result to first order. Applying a Taylor expansion of $\mathrm{softmax}$ around $L_t$ with perturbation $\delta = \lambda(\bar{L}_t - L_t)$:

\noindent\textit{EMAG-Q output} (from Eqs.~\ref{eq:emag_q_formulation} and~\ref{eq:logit_blend}):
\begin{align}
\mathrm{softmax}\!\big((1-\lambda)\,L_t + \lambda\,\bar{L}_t\big) 
&= \mathrm{softmax}\!\big(L_t + \lambda(\bar{L}_t - L_t)\big) \nonumber \\
&\approx A_t + \lambda\,J(L_t)\,(\bar{L}_t - L_t) \label{eq:emagq_taylor}
\end{align}

\noindent\textit{Vanilla EMAG output} (from Eqs.~\ref{eq:emag_orig} and~\ref{eq:abar_approx}):
\begin{align}
(1-\lambda)\,A_t + \lambda\,\bar{A}_t 
&\approx (1-\lambda)\,A_t + \lambda\,\mathrm{softmax}(\bar{L}_t) \nonumber \\
&\approx (1-\lambda)\,A_t + \lambda\big[A_t + J(L_t)\,(\bar{L}_t - L_t)\big] \nonumber \\
&= A_t + \lambda\,J(L_t)\,(\bar{L}_t - L_t) \label{eq:emag_taylor}
\end{align}

Since Eqs.~\ref{eq:emagq_taylor} and~\ref{eq:emag_taylor} are identical, we conclude:
\begin{align}
\boxed{\mathrm{softmax}\!\big((1\!-\!\lambda)\,L_t + \lambda\,\bar{L}_t\big) \;\approx\; (1\!-\!\lambda)\,A_t + \lambda\,\bar{A}_t + \mathcal{O}(\|\delta\|^2 + \mathcal{\mathrm{Var}_{\tau}[L_{\tau}]})}
\label{eq:equivalence}
\end{align}


The total approximation error has two sources: the Taylor expansion in Step~3, scaling with $\|\delta\|^2 = \lambda^2\|\bar{L}_t - L_t\|^2$, and the softmax nonlinearity under temporal logit variance in Step~2. Both remain small in practice because the EMA with $\beta$ close to 1 tracks the signal closely, ensuring $\bar{L}_t \approx L_t$ and suppressing temporal variance. In our setting $\lambda = 1$ (hard replacement), so the approximation quality is governed entirely by the EMA decay rate $\beta$. In diffusion transformers, where attention logits evolve smoothly across adjacent denoising steps, this condition is naturally satisfied.





\subsubsection{Computational Benefits}
Table~\ref{tab:timing_benchmark} quantifies the efficiency gains of EMAG-Q over vanilla EMAG. Since the EMA is tracked over query embeddings $\mathbb{R}^{N \times d}$ rather than the full attention map $\mathbb{R}^{N \times N}$, the memory footprint of the EMA buffer is reduced by a factor of $N/d$. Moreover, because EMAG-Q modifies only the query input and leaves the attention computation unchanged, it is fully compatible with Fused SDPA, yielding substantial wall-clock speedups. In contrast, vanilla EMAG requires explicit attention map materialization and replacement, precluding such kernel-level optimizations.

\subsubsection{Empirical Validation}
\begin{figure*}[t]
  \centering
  \includegraphics[width=\linewidth]{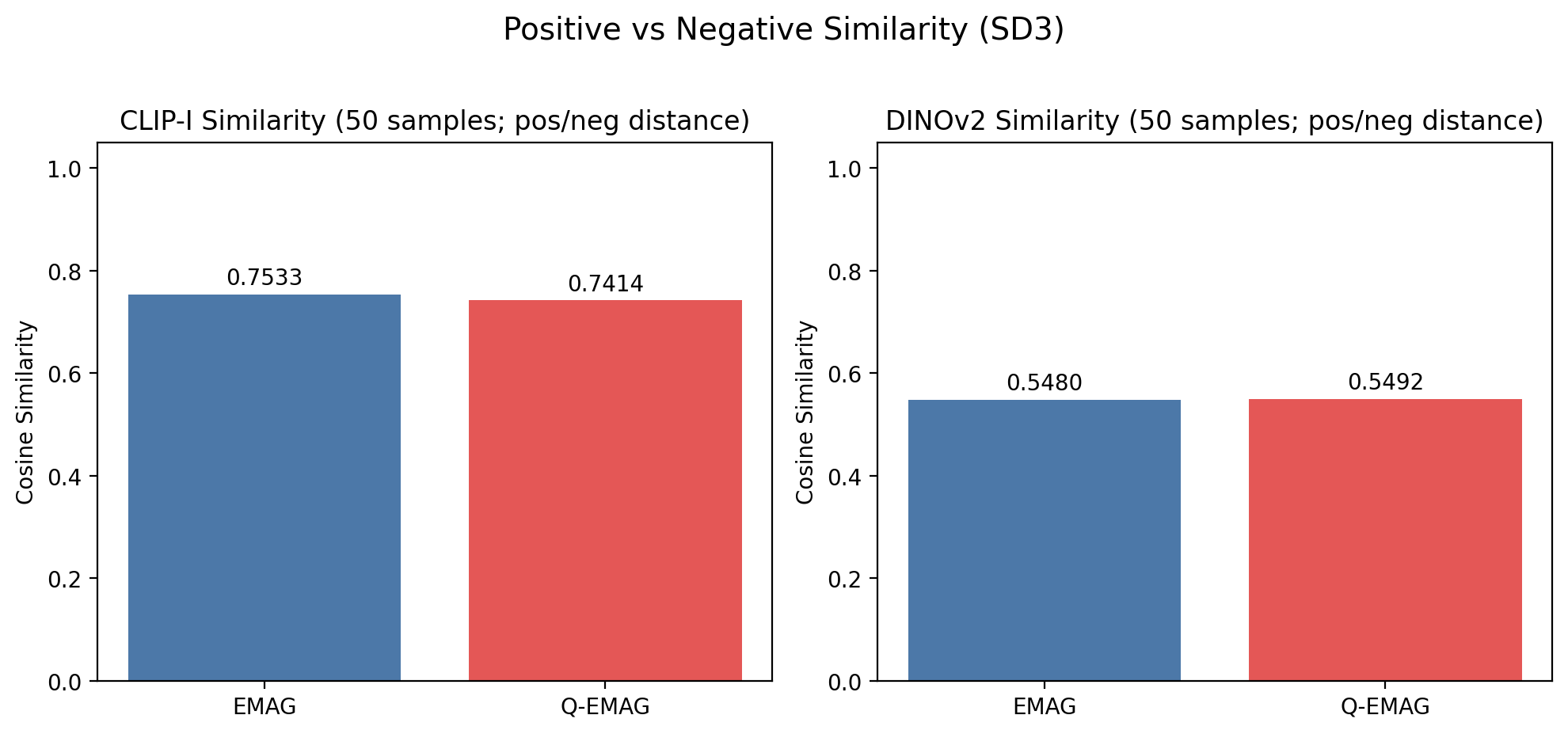}
      \caption{\textbf{Negative generation similarity comparison between EMAG-Q and EMAG }}

    \label{fig: section_D_EMAG-Q-VS-EMAG_bars}
\end{figure*}

\begin{figure*}[t]
  \centering
  \includegraphics[width=\linewidth]{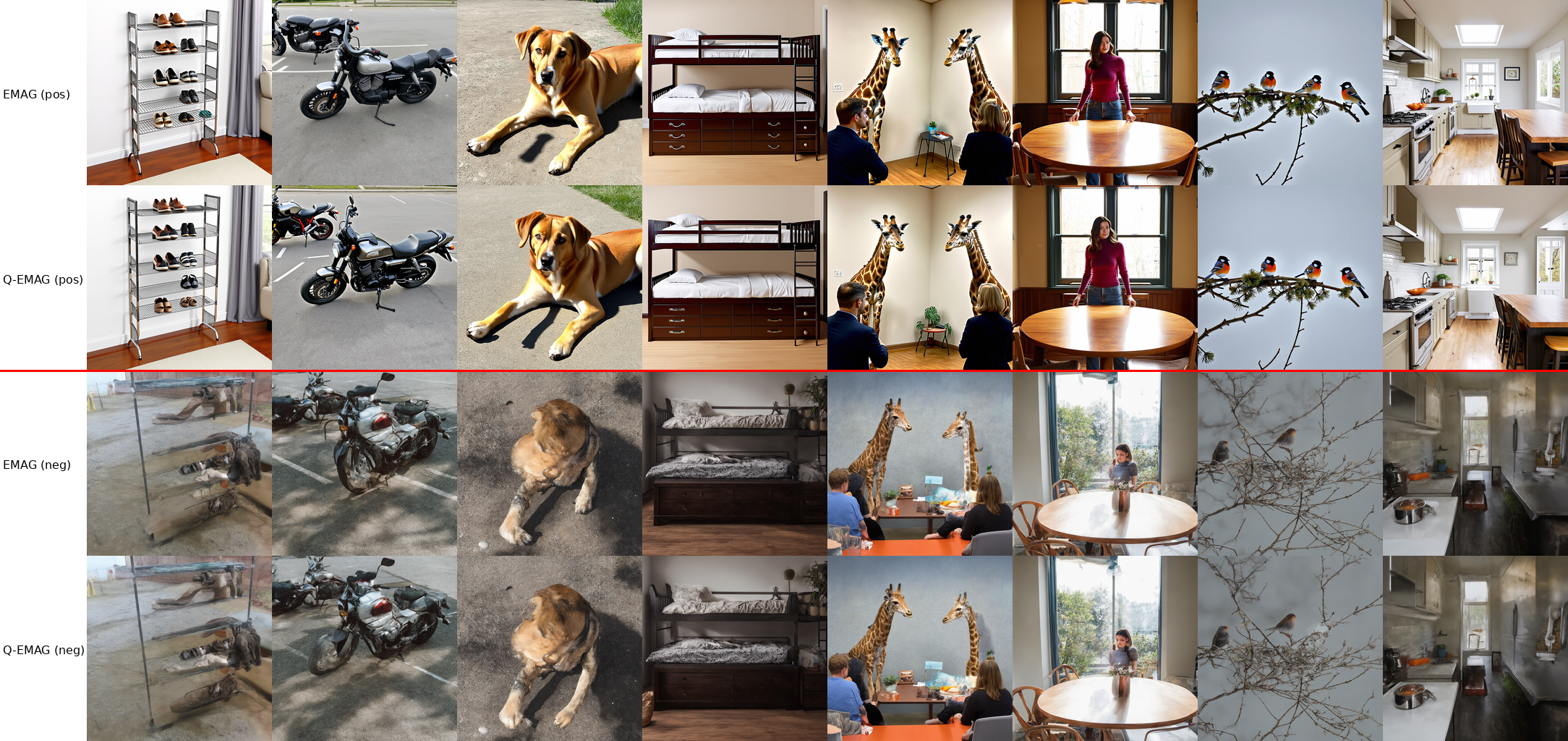}
      \caption{\textbf{Generation comparison between EMAG-Q and EMAG }}

    \label{fig: section_D_EMAG-Q-VS-EMAG}
\end{figure*}

We verify that the computational savings of EMAG-Q do not come at the cost of generation quality. Fig.~\ref{fig: section_D_EMAG-Q-VS-EMAG_bars} compares CLIP-I and DINO similarity scores between generated images and their corresponding negatives for both variants; the distributions are closely matched, confirming that EMAG-Q produces hard negatives of comparable fidelity. Fig.~\ref{fig: section_D_EMAG-Q-VS-EMAG} provides a qualitative comparison: using the same seed and initial latent, both approaches yield visually similar generations and negatives. Table~\ref{tab:emag_q_comparison} further supports this with quantitative metrics. Finally, we measure the per-image HPS correlation between EMAG-Q and vanilla EMAG outputs and find it comparable to the correlation observed across different seed variations of vanilla EMAG alone, indicating that the two variants are functionally interchangeable.




\begin{figure*}[t]
  \centering

  \includegraphics[width=\linewidth]{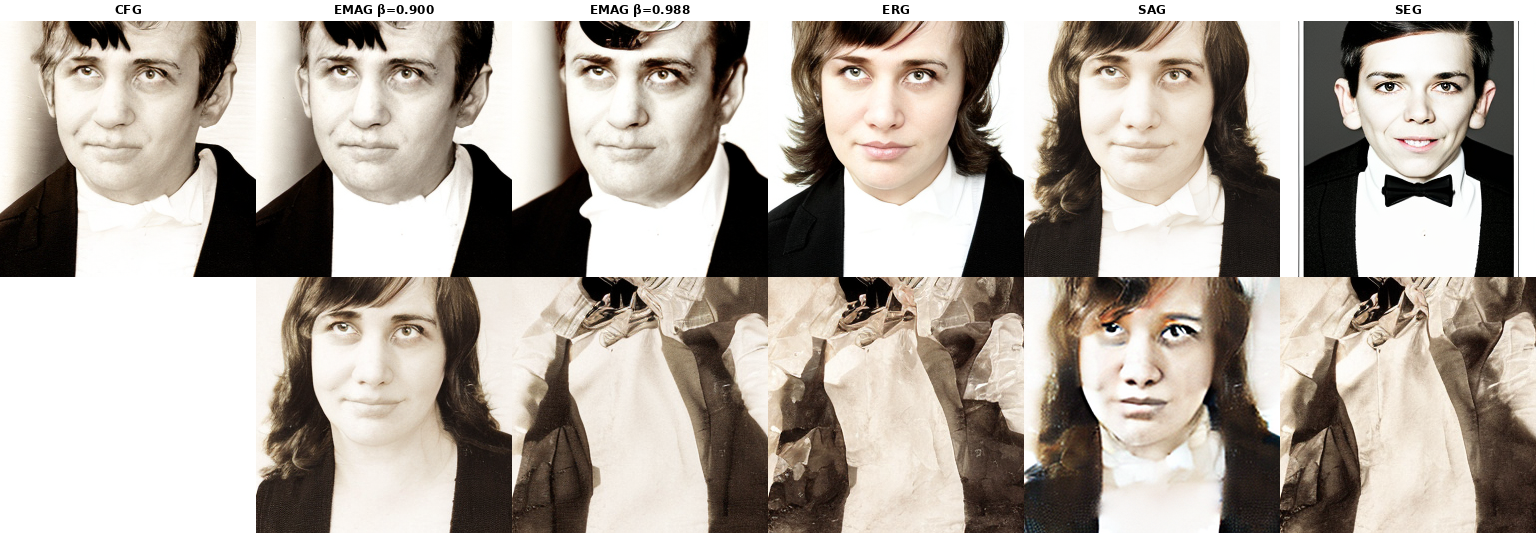}

  \vspace{0.6em} 

  \includegraphics[width=\linewidth]{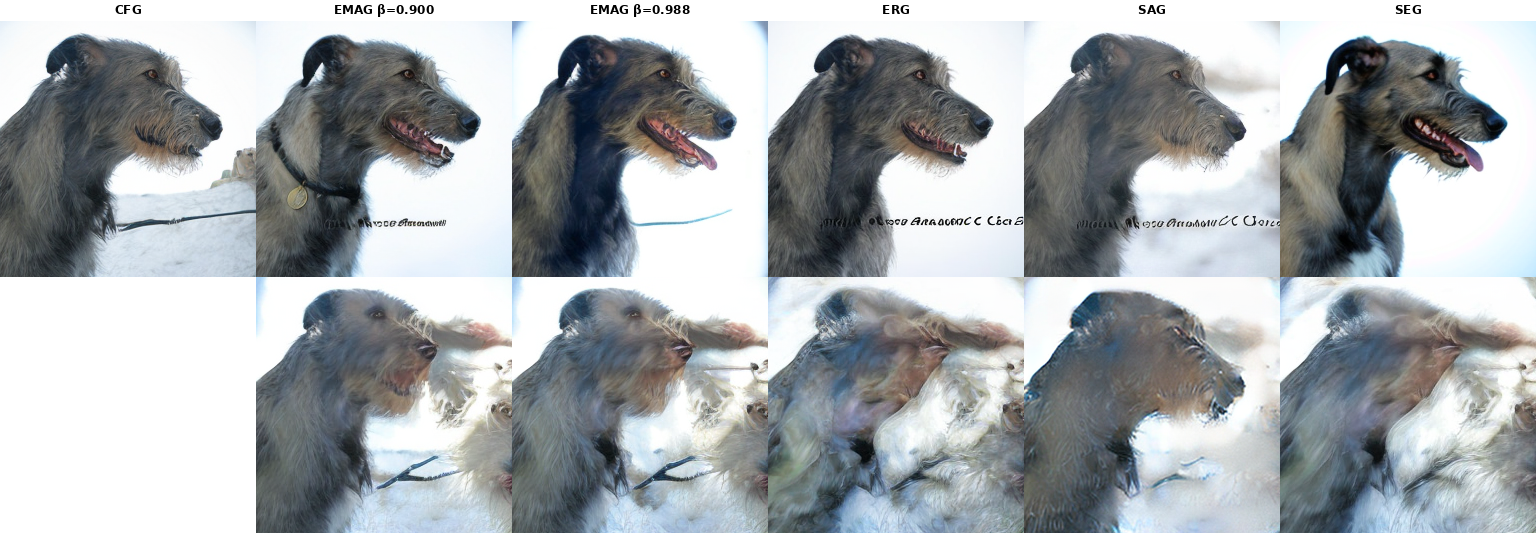}

  \caption{Qualitative examples of positive and negative samples for EMAG and competing baselines. The top row shows positive samples, while the bottom row shows the corresponding negative samples generated using the same conditioning class and random seed (seed = 8).}

  \label{fig:section_F_hard_negative_qualitative}
\end{figure*}

\section{Limitations and ethical considerations}

Our study focuses on image generation with Transformer backbones; extending EMAG to video/audio/3D is left for future work. As a test-time guidance method, EMAG inherits biases and misuse risks from its foundation models; we keep built-in safety filters enabled and avoid sensitive content.

\begin{table*}[t]
\centering
\small
\setlength{\tabcolsep}{6pt}

\begin{tabular}{l p{0.8\linewidth}}
\toprule
\textbf{Figure} & \textbf{Prompt} \\
\midrule
Fig.~\ref{fig:front_image}  & Ferry deck in windy daylight: white railings with a safety placard ``MUSTER STATION B''; a life ring labeled ``MV CORAL''; a paper cup with tea pressed by a metal spoon; distant headland under broken clouds; a timetable brochure on a bench marked ``11:40 $\rightarrow$ 12:15''. Bright, high-key light; ripples on water; clean maritime palette of whites, blues, and orange accents. \\
Fig.~\ref{fig: section_D_cond_sd3_EMAG}  & A bus parked in a large parking lot. \\
Fig.~\ref{fig: section_D_cond_sd3_EMAG}  & A man holding a camera up over his left shoulder. \\
Fig.~\ref{fig: section_D_cond_sd3_EMAG}  & A person doing karate in a field at night. \\
Fig.~\ref{fig: section_D_cond_sd3_EMAG}  & A close up of a plate of food containing broccoli. \\
Fig.~\ref{fig: section_D_cond_sd3_EMAG}  & A beautiful woman taking a picture with her smart phone. \\
Fig.~\ref{fig: section_D_cond_sd3_EMAG}  & Two people standing in a kitchen looking around. \\
Fig.~\ref{fig: section_D_cond_sd3_EMAG}  & A young man wearing black attire and a flowered tie is standing and smiling. \\
Fig.~\ref{fig: section_D_cond_sd3_EMAG}  & A sink is shown with a mirror and lights. \\
Fig.~\ref{fig: section_D_cond_sd3_EMAG} & A bus that is sitting in the street. \\
Fig.~\ref{fig: section_D_cond_sd3_EMAG} & A plane flies over water with two islands nearby. \\
Fig.~\ref{fig: section_D_cond_sd3_EMAG} & A dog sitting on the inside of a white boat. \\
Fig.~\ref{fig: section_D_cond_sd3_EMAG} & Two people standing in a kitchen looking around. \\
Fig.~\ref{fig: section_D_cond_sd3_EMAG} & A banana that is sitting in a bowl on the table. \\
Fig.~\ref{fig: section_D_cond_sd3_EMAG} & A coffee table sits in the middle of a living room.\\
Fig.~\ref{fig: section_D_cond_sd3_EMAG} & A cute little girl riding a skateboard in a store. \\
Fig.~\ref{fig: section_D_cond_sd3_EMAG} & A rear view mirror on a bike is reflecting a poster on building. \\
Fig.~\ref{fig:cond-samples} & This wire metal rack holds several pairs of shoes and sandals. \\
Fig.~\ref{fig:cond-samples} & A small kitchen with low a ceiling. \\
Fig.~\ref{fig:cond-samples} & A woman stands in the dining area at the table. \\
Fig.~\ref{fig: section_D_EMAG-Q-VS-EMAG} & This wire metal rack holds several pairs of shoes and sandals. \\
Fig.~\ref{fig: section_D_EMAG-Q-VS-EMAG} & A motorcycle parked in a parking space next to another motorcycle. \\
Fig.~\ref{fig: section_D_EMAG-Q-VS-EMAG} & A picture of a dog laying on the ground. \\
Fig.~\ref{fig: section_D_EMAG-Q-VS-EMAG} & A loft bed with a dresser underneath it \\
Fig.~\ref{fig: section_D_EMAG-Q-VS-EMAG} & Two giraffes in a room with people looking at them. \\
Fig.~\ref{fig: section_D_EMAG-Q-VS-EMAG} & A woman stands in the dining area at the table. \\
Fig.~\ref{fig: section_D_EMAG-Q-VS-EMAG} & Birds perch on a bunch of twigs in the winter. \\
Fig.~\ref{fig: section_D_EMAG-Q-VS-EMAG} & A small kitchen with low a ceiling. \\
\bottomrule
\end{tabular}
\caption{Prompts used for qualitative examples.}
\label{tab:qual_prompts}
\end{table*}

\begin{table*}[t]
\centering
\small
\setlength{\tabcolsep}{6pt}
\caption{We ablate the impact of the different layers. The experiments are conducted for 5000 samples. All runs use the identical COCO 2014 validation set, sampling steps, and evaluation code.}
\label{tab:Adaptive_layer_selection_ablation_supp2}
\resizebox{\columnwidth}{!}{
\begin{tabular}{lccccccc}
\toprule
\textbf{Model / Guidance} & \textbf{FID} $\downarrow$ & \textbf{CLIP Score} $\uparrow$ & \textbf{Precision} $\uparrow$ & \textbf{Recall} $\uparrow$ & \textbf{Density} $\uparrow$ & \textbf{Coverage} $\uparrow$ & \textbf{HPS} $\uparrow$\\
\midrule
\multicolumn{8}{l}{\textit{SD3 (MMDiT)}} \\
\midrule
EMAG (L 1)     & 27.34 & 26.759 & 0.827 & 0.578 & 1.104 & 0.450 & 29.64 \\ 
EMAG (L 2)     & 28.58 & 26.836 & 0.787 & 0.539 & 0.970 & 0.410 & 29.37 \\ 
EMAG (L 3)     & 27.46 & 26.792 & 0.820 & 0.571 & 1.113 & 0.450 & 29.69 \\ 
EMAG (L 4)     & 28.11 & 26.725 & 0.815 & 0.550 & 1.048 & 0.434 & 29.66 \\ 
EMAG (L 5)     & 28.03 & 26.699 & 0.795 & 0.538 & 0.982 & 0.419 & 29.71 \\ 
\midrule
EMAG (L 6)     & 27.47 & 26.604 & 0.824 & 0.548 & 1.102 & 0.452 & 29.54 \\ 
EMAG (L 7)     & 28.94 & 26.733 & 0.788 & 0.508 & 1.008 & 0.424 & 29.51 \\ 
EMAG (L 8)     & 28.23 & 26.768 & 0.810 & 0.543 & 1.040 & 0.431 & 29.66 \\ 
\midrule
EMAG (L 9)     & 29.57 & 26.555 & 0.824 & 0.505 & 1.102 & 0.438 & 30.52 \\ 
EMAG (L 10)    & 28.90 & 26.697 & 0.819 & 0.526 & 1.057 & 0.433 & 30.18 \\ 
EMAG (L 11)    & 31.40 & 26.742 & 0.771 & 0.481 & 0.912 & 0.384 & 30.27 \\ 
EMAG (L 12)    & 29.58 & 26.694 & 0.803 & 0.514 & 1.026 & 0.418 & 30.32 \\ 
EMAG (L 13)    & 29.07 & 26.920 & 0.794 & 0.527 & 0.943 & 0.411 & 29.64 \\ 
EMAG (L 14)    & 29.77 & 26.615 & 0.797 & 0.513 & 1.066 & 0.431 & 30.40 \\ 
EMAG (L 15)    & 30.36 & 26.706 & 0.799 & 0.510 & 0.978 & 0.407 & 30.15 \\ 
EMAG (L 16)    & 33.12 & 26.895 & 0.753 & 0.462 & 0.836 & 0.359 & 29.74 \\ 
EMAG (L 17)    & 31.55 & 26.848 & 0.772 & 0.485 & 0.899 & 0.385 & 29.88 \\ 
EMAG (L 18)    & 32.51 & 26.910 & 0.762 & 0.489 & 0.886 & 0.371 & 29.08 \\ 
\midrule
EMAG (L all)      & 35.60 & 26.849 & 0.688 & 0.465 & 0.671 & 0.311 & 28.80 \\ 
EMAG (Adaptive)   & 28.52 & 26.733 & 0.806 & 0.530 & 1.040 & 0.430 & 29.60 \\ 
\bottomrule
\end{tabular}
}
\end{table*}


\begin{table*}[t]
\centering
\small
\setlength{\tabcolsep}{6pt}
\caption{Ablation study comparing different operation ranges for EMAG ($\tau_{s}$ and $\tau_{e}$). The experiments are conducted for 1000 samples. All runs use the identical COCO 2014 validation set, sampling steps, and evaluation code.}
\label{tab:timestep_ablation_supp}
\resizebox{\columnwidth}{!}{
\begin{tabular}{ccccccccc}
\toprule
$\tau_{s}$ & $\tau_{e}$ & \textbf{FID} $\downarrow$ & \textbf{CLIP Score} $\uparrow$ & \textbf{Precision} $\uparrow$ & \textbf{Recall} $\uparrow$ & \textbf{Density} $\uparrow$ & \textbf{Coverage} $\uparrow$ & \textbf{HPS} $\uparrow$\\
\midrule
\multicolumn{9}{l}{\textit{SD3 (MMDiT)}} \\
\midrule
10 & 5  & 55.63 & 26.643 & 0.843 & 0.695 & 1.168 & 0.167 & 29.49 \\ 
14 & 5  & 55.37 & 26.646 & 0.838 & 0.702 & 1.152 & 0.165 & 29.57 \\ 
18 & 5  & 55.15 & 26.689 & 0.839 & 0.683 & 1.148 & 0.164 & 29.70 \\ 
26 & 0  & 56.62 & 26.709 & 0.808 & 0.651 & 1.062 & 0.150 & 30.00 \\ 
26 & 2  & 56.62 & 26.709 & 0.808 & 0.651 & 1.062 & 0.150 & 30.00 \\ 
26 & 5  & 55.30 & 26.748 & 0.821 & 0.682 & 1.046 & 0.151 & 29.82 \\ 
26 & 8  & 56.62 & 26.709 & 0.808 & 0.651 & 1.062 & 0.150 & 30.00 \\ 
26 & 10 & 55.70 & 26.775 & 0.823 & 0.667 & 1.167 & 0.162 & 29.98 \\ 
26 & 15 & 56.62 & 26.709 & 0.808 & 0.651 & 1.062 & 0.150 & 30.00 \\ 
\bottomrule
\end{tabular}
}
\end{table*}

\begin{table}[t]
\centering
\caption{Head-to-head comparison of EMAG and its efficient variant EMAG-Q on 5K COCO captions (seed=8). Both use SD3 Medium with $w{=}7$, $w_e{=}1.5$, $\beta{=}0.988$, layers $\{6,7,8\}$, adaptive mode 2, 28 steps. EMAG-Q replaces post-softmax attention map EMA with pre-softmax query EMA, enabling fused SDPA kernels. Per-image correlations (SRCC/PLCC) are computed on 1K paired samples. The seed-variation row shows EMAG vs itself with a different seed, establishing the upper bound on achievable per-image correlation.}

\label{tab:emag_q_comparison}
\small
\setlength{\tabcolsep}{4pt}
\renewcommand{\arraystretch}{1.05}
\resizebox{\columnwidth}{!}{
\begin{tabular}{lccccccccccc}
\toprule
\textbf{Variant}
& \textbf{FID$\downarrow$}
& \textbf{HPS$\uparrow$}
& \textbf{CLIP$\uparrow$}
& \textbf{Prec$\uparrow$}
& \textbf{Rec$\uparrow$}
& \textbf{Den$\uparrow$}
& \textbf{Cov$\uparrow$}
& \textbf{s/img}
& \textbf{Mem} 
& \textbf{HPS\_SRCC$\uparrow$}
& \textbf{HPS\_PLCC$\uparrow$} 
\\
\midrule
EMAG       & 28.51 & 29.70 & 26.68 & 0.833 & 0.551 & 1.110 & 0.443 & 12.66 & 29.58\,GB & -- & -- \\
EMAG-Q     & 29.14 & 29.58 & 26.70 & 0.812 & 0.535 & 1.127 & 0.445 & 7.86  & 16.96\,GB & -- & -- \\
\midrule
$\Delta$   & $+$0.63 & $-$0.12 & $+$0.02 & $-$0.021 & $-$0.016 & $+$0.017 & $+$0.002 & $-$38\% & $-$43\% & 0.741 & 0.749 \\
\midrule
EMAG (seed var.) & -- & -- & -- & -- & -- & -- & -- & -- & -- & 0.754 & 0.758 \\
\bottomrule
\end{tabular}
}
\end{table}

%
\begin{table*}[t]
\centering
\small
\setlength{\tabcolsep}{5pt}
\caption{Standalone guidance comparison \textbf{without CFG}. Each method operates as the sole guidance signal (no classifier-free guidance). All experiments use 1000 samples from COCO 2014 validation, 28 sampling steps, and seed 8. Best result per standalone method is \underline{underlined}; overall best is \textbf{bolded}. The ``Combined'' rows show the best CFG-combined configuration for each method, evaluated under identical 1K settings.}
\label{tab:standalone_guidance_comparison}
\resizebox{\textwidth}{!}{%
\begin{tabular}{llccccccc}
\toprule
\textbf{Method} & \textbf{Scale} & \textbf{CFG Scale} & \textbf{FID} $\downarrow$ & \textbf{Precision} $\uparrow$ & \textbf{Recall} $\uparrow$ & \textbf{Density} $\uparrow$ & \textbf{Coverage} $\uparrow$ & \textbf{HPS} $\uparrow$\\
\midrule
\multicolumn{9}{l}{\textit{Baselines}} \\
\midrule
No guidance & — & 0 & 58.76 & 0.670 & 0.758 & 0.599 & 0.101 & 21.86  \\
CFG only    & — & 7 & 56.93 & 0.834 & 0.669 & 1.158 & 0.165 & 29.22  \\
\midrule
\multicolumn{9}{l}{\textit{SAG Standalone (no CFG)}} \\
\midrule
SAG & $s=0.1$  & 0 & 56.44  & 0.690 & \underline{0.780} & 0.673 & 0.113 & 22.30 \\
SAG & $s=0.2$  & 0 & 54.11  & 0.702 & 0.780 & 0.742 & 0.124 & 22.75 \\
SAG & $s=0.3$  & 0 & 52.57  & 0.728 & 0.779 & 0.792 & 0.132 & 23.11 \\
SAG & $s=0.5$  & 0 & 50.74  & 0.748 & 0.777 & 0.826 & 0.137 & 23.71 \\
SAG & $s=0.75$ & 0 & \underline{50.09}  & 0.752 & 0.760 & 0.866 & \underline{0.143} & 24.26 \\
SAG & $s=1.0$  & 0 & 50.53  & \underline{0.758} & 0.738 & \underline{0.880} & \underline{0.143} & 24.62 \\
SAG & $s=1.5$  & 0 & 53.16  & 0.731 & 0.697 & 0.828 & 0.134 & 25.10 \\
SAG & $s=2.0$  & 0 & 56.85  & 0.716 & 0.673 & 0.730 & 0.117 & \underline{25.17} \\
SAG & $s=3.0$  & 0 & 65.10  & 0.662 & 0.619 & 0.562 & 0.090 & 24.89 \\
SAG + CFG   & $s=0.25$ & 7 & 56.42 & 0.840 & 0.669 & 1.158 & 0.161 & 29.28 \\
\midrule
\multicolumn{9}{l}{\textit{SEG Standalone (no CFG)}} \\
\midrule
SEG & $s=0.5$  & 0 & 176.76 & 0.099 & 0.443 & 0.026 & 0.004 & 10.38 \\
SEG & $s=1.0$  & 0 & 59.53  & 0.661 & \underline{0.773} & 0.609 & 0.102 & 21.89 \\
SEG & $s=2.0$  & 0 & \underline{51.90}  & \underline{\textbf{0.846}} & 0.679 & \underline{\textbf{1.162}} & \underline{\textbf{0.178}} & 28.39 \\
SEG & $s=3.0$  & 0 & 56.06  & 0.803 & 0.621 & 1.063 & 0.158 & 29.24 \\
SEG & $s=4.0$  & 0 & 60.69  & 0.786 & 0.595 & 0.913 & 0.137 & \underline{29.25} \\
SEG & $s=5.0$  & 0 & 63.80  & 0.735 & 0.555 & 0.860 & 0.126 & 29.05 \\
SEG & $s=6.0$  & 0 & 67.25  & 0.681 & 0.525 & 0.689 & 0.103 & 28.71 \\
SEG + CFG   & $s=2.0$ & 7 & 60.10 & 0.786 & 0.610 & 0.971 & 0.139 & \textbf{30.18} \\
\midrule
\multicolumn{9}{l}{\textit{PAG Standalone (no CFG)}} \\
\midrule
PAG & $s=1.0$  & 0 & \underline{81.21}  & \underline{0.635} & \underline{0.594} & \underline{0.615} & \underline{0.095} & 22.78 \\
PAG & $s=2.0$  & 0 & 93.76  & 0.561 & 0.585 & 0.545 & 0.082 & \underline{22.83} \\
PAG & $s=3.0$  & 0 & 97.53  & 0.516 & 0.579 & 0.475 & 0.075 & 22.77 \\
PAG & $s=4.0$  & 0 & 99.46  & 0.483 & 0.560 & 0.412 & 0.067 & 22.58 \\
PAG & $s=5.0$  & 0 & 101.26 & 0.447 & 0.553 & 0.353 & 0.056 & 22.44 \\
PAG + CFG   & $s=1.0$ & 5 & 58.05 & 0.805 & 0.645 & 1.102 & 0.157 & 28.84 \\
\midrule
\multicolumn{9}{l}{\textit{EMAG Standalone (no CFG)}} \\
\midrule
EMAG & $w_e=1.75$  & 0 & 50.00 & 0.773 & 0.778 & 0.813 & 0.136 & 24.03 \\
EMAG & $w_e=2.0$   & 0 & 48.49 & 0.784 & \underline{\textbf{0.781}} & 0.886 & 0.146 & 24.50 \\
EMAG & $w_e=2.5$   & 0 & 47.15 & 0.792 & 0.763 & 0.991 & 0.162 & 25.42 \\
EMAG & $w_e=3.0$   & 0 & \underline{\textbf{46.71}} & 0.812 & 0.758 & 1.056 & 0.170 & 26.27 \\
EMAG & $w_e=4.0$   & 0 & 47.92 & \underline{0.835} & 0.724 & \underline{1.083} & \underline{0.172} & 27.26 \\
EMAG & $w_e=5.0$   & 0 & 48.16 & 0.823 & 0.717 & 1.011 & 0.163 & 27.64 \\
EMAG & $w_e=7.0$   & 0 & 51.54 & 0.784 & 0.691 & 0.869 & 0.139 & 28.04 \\
EMAG & $w_e=9.0$   & 0 & 55.44 & 0.756 & 0.654 & 0.793 & 0.124 & \underline{28.08} \\
EMAG & $w_e=12.0$  & 0 & 65.15 & 0.675 & 0.571 & 0.633 & 0.096 & 27.62 \\
EMAG + CFG  & $w_e=1.5$ & 7 & 56.98 & 0.843 & 0.681 & 1.085 & 0.155 & 29.50 \\
\bottomrule
\end{tabular}
}
\end{table*}

\begin{table*}[t]
\centering
\small
\setlength{\tabcolsep}{5pt}
\caption{Summary of best standalone (no CFG) versus best CFG-combined configuration for each guidance method on SD3. Standalone scale is selected by best HPS from the full sweep in Table~\ref{tab:standalone_guidance_comparison}. The scale ratio (Standalone$\div$Combined) quantifies how much the method's optimal scale decreases when CFG is added; a larger ratio suggests the method can substantially reduce its own contribution when CFG handles coarse semantic alignment.}
\label{tab:standalone_summary}
\resizebox{\textwidth}{!}{%
\begin{tabular}{llcccccccc}
\toprule
\textbf{Method} & \textbf{Mode} & \textbf{Scale} & \textbf{CFG} & \textbf{FID} $\downarrow$ & \textbf{Precision} $\uparrow$ & \textbf{Recall} $\uparrow$ & \textbf{Density} $\uparrow$ & \textbf{HPS} $\uparrow$ & \textbf{Scale Ratio}\\
\midrule
\multicolumn{4}{l}{\textit{Baselines}} \\
No guidance & — & — & 0 & 58.76 & 0.670 & 0.758 & 0.599 & 21.86 & — \\
CFG only    & — & — & 7 & 56.93 & 0.834 & 0.669 & 1.158 & 29.22 & — \\
\midrule
\multirow{2}{*}{SAG}
  & Standalone & $s=2.0$    & 0 & 56.85 & 0.716 & 0.673 & 0.730 & 25.17 & \multirow{2}{*}{$8.0\times$} \\
  & Combined   & $s=0.25$   & 7 & 56.42 & 0.840 & 0.669 & 1.158 & 29.28 & \\
\midrule
\multirow{2}{*}{SEG}
  & Standalone & $s=4.0$    & 0 & 60.69 & 0.786 & 0.595 & 0.913 & 29.25 & \multirow{2}{*}{$2.0\times$} \\
  & Combined   & $s=2.0$    & 7 & 60.10 & 0.786 & 0.610 & 0.971 & \textbf{30.18} & \\
\midrule
\multirow{2}{*}{PAG}
  & Standalone & $s=2.0$    & 0 & 93.76 & 0.561 & 0.585 & 0.545 & 22.83 & \multirow{2}{*}{$2.0\times$} \\
  & Combined   & $s=1.0$    & 5 & 58.05 & 0.805 & 0.645 & 1.102 & 28.84 & \\
\midrule
\multirow{2}{*}{\textbf{EMAG}}
  & Standalone & $w_e=9.0$  & 0 & 55.44 & 0.756 & 0.654 & 0.793 & 28.08 & \multirow{2}{*}{$\mathbf{6.0\times}$} \\
  & Combined   & $w_e=1.5$  & 7 & 56.98 & 0.843 & 0.681 & 1.085 & 29.50 & \\
\bottomrule
\end{tabular}
}
\end{table*}

\begin{table*}[t]
\centering
\small
\setlength{\tabcolsep}{5pt}
\caption{PAG hyperparameter sweep on SD3. Block~1 fixes PAG scale and sweeps CFG; Block~2 fixes CFG and sweeps PAG; Block~3 tests PAG standalone without CFG. All experiments use 1000 samples from COCO 2014 validation, 28 steps, and seed 8. Corresponding Pareto frontier curves are shown in Fig.~\ref{fig:pag_pareto_frontier_curve}.}
\label{tab:pag_pareto_sweep}
\resizebox{\textwidth}{!}{%
\begin{tabular}{llccccccc}
\toprule
\textbf{Method} & \textbf{PAG Scale} & \textbf{CFG Scale} & \textbf{FID} $\downarrow$ & \textbf{Precision} $\uparrow$ & \textbf{Recall} $\uparrow$ & \textbf{Density} $\uparrow$ & \textbf{Coverage} $\uparrow$ & \textbf{HPS} $\uparrow$\\
\midrule
\multicolumn{9}{l}{\textit{Block 1: Fixed PAG$=1.0$, sweep CFG}} \\
\midrule
PAG + CFG & $s=1.0$ &  3 & 60.01 & 0.817 & 0.640 & 1.130 & 0.158 & 29.10 \\
PAG + CFG & $s=1.0$ &  5 & 58.05 & 0.805 & 0.645 & 1.102 & 0.157 & 29.13 \\
PAG + CFG & $s=1.0$ &  7 & 56.98 & 0.713 & 0.680 & 0.806 & 0.122 & 27.88 \\
PAG + CFG & $s=1.0$ &  9 & 66.10 & 0.557 & 0.717 & 0.537 & 0.084 & 25.61 \\
PAG + CFG & $s=1.0$ & 11 & 83.14 & 0.471 & 0.716 & 0.321 & 0.051 & 23.18 \\
\midrule
\multicolumn{9}{l}{\textit{Block 2: Fixed CFG$=7$, sweep PAG}} \\
\midrule
PAG + CFG & $s=1.0$ & 7 & 56.98 & 0.713 & 0.680 & 0.806 & 0.122 & 27.88 \\
PAG + CFG & $s=2.0$ & 7 & 81.86 & 0.478 & 0.693 & 0.348 & 0.056 & 23.67 \\
PAG + CFG & $s=3.0$ & 7 & 120.27 & 0.322 & 0.645 & 0.137 & 0.019 & 20.39 \\
PAG + CFG & $s=4.0$ & 7 & 148.54 & 0.233 & 0.550 & 0.057 & 0.006 & 18.29 \\
PAG + CFG & $s=5.0$ & 7 & 169.75 & 0.172 & 0.289 & 0.039 & 0.003 & 16.94 \\
\midrule
\multicolumn{9}{l}{\textit{Block 3: PAG Standalone (no CFG)}} \\
\midrule
PAG & $s=1.0$ & 0 & 81.21 & 0.635 & 0.594 & 0.615 & 0.095 & 22.78 \\
PAG & $s=2.0$ & 0 & 93.76 & 0.561 & 0.585 & 0.545 & 0.082 & 22.83 \\
PAG & $s=3.0$ & 0 & 97.53 & 0.516 & 0.579 & 0.475 & 0.075 & 22.77 \\
PAG & $s=4.0$ & 0 & 99.46 & 0.483 & 0.560 & 0.412 & 0.067 & 22.58 \\
PAG & $s=5.0$ & 0 & 101.26 & 0.447 & 0.553 & 0.353 & 0.056 & 22.44 \\
\bottomrule
\end{tabular}
}
\end{table*}

\begin{table*}[t]
\centering
\small
\setlength{\tabcolsep}{5pt}
\caption{EMAG performance under reduced inference step regimes on SD3. Block~1 evaluates the default EMAG configuration ($\beta{=}0.988$) across step counts to test graceful degradation. Block~2 adapts $\beta$ to the step budget (targeting EMA half-life $\approx T/5$) to recover guidance effectiveness at low steps. CFG-only baselines at each step count are provided for computing relative improvement ($\Delta$HPS). All experiments use 1000 samples from COCO 2014 validation, CFG scale 7.0, $w_e{=}1.50$, and seed 8.}
\label{tab:emag_fewstep_ablation}
\resizebox{\textwidth}{!}{%
\begin{tabular}{llcccccccc}
\toprule
\textbf{Method} & \textbf{Steps} & $\boldsymbol{\beta}$ & \textbf{FID} $\downarrow$ & \textbf{Precision} $\uparrow$ & \textbf{Recall} $\uparrow$ & \textbf{Density} $\uparrow$ & \textbf{Coverage} $\uparrow$ & \textbf{HPS} $\uparrow$ & $\boldsymbol{\Delta}$\textbf{HPS}\\
\midrule
\multicolumn{10}{l}{\textit{CFG-only baselines (no EMAG)}} \\
\midrule
CFG only & 28 & — & 55.87 & 0.834 & 0.698 & 1.147 & 0.164 & 29.58 & — \\
CFG only & 20 & — & 56.14 & 0.832 & 0.675 & 1.081 & 0.156 & 29.13 & — \\
CFG only & 14 & — & 55.85 & 0.815 & 0.673 & 1.067 & 0.153 & 28.09 & — \\
CFG only &  8 & — & 58.49 & 0.693 & 0.666 & 0.721 & 0.107 & 25.27 & — \\
CFG only &  4 & — & 99.78 & 0.293 & 0.589 & 0.139 & 0.024 & 16.33 & — \\
\midrule
\multicolumn{10}{l}{\textit{Block 1: Default EMAG ($\beta{=}0.988$, half-life $\approx$ 57 steps)}} \\
\midrule
EMAG & 28 & 0.988 & 56.19 & 0.830 & 0.669 & 1.089 & 0.156 & 30.15 & +0.57 \\
EMAG & 20 & 0.988 & 56.46 & 0.817 & 0.649 & 1.025 & 0.149 & 30.24 & +1.11 \\
EMAG & 14 & 0.988 & 56.26 & 0.811 & 0.647 & 1.001 & 0.147 & 28.77 & +0.68 \\
EMAG &  8 & 0.988 & 56.73 & 0.715 & 0.671 & 0.755 & 0.113 & 26.42 & +1.15 \\
EMAG &  4 & 0.988 & 99.72 & 0.290 & 0.627 & 0.130 & 0.023 & 16.47 & +0.15 \\
\midrule
\multicolumn{10}{l}{\textit{Block 2: Adapted $\beta$ for low-step regimes (half-life $\approx T/5$)}} \\
\midrule
EMAG & 14 & 0.75 & 55.46 & 0.805 & 0.668 & 1.066 & 0.156 & 27.89 & $-$0.20 \\
EMAG &  8 & 0.70 & 58.07 & 0.714 & 0.682 & 0.706 & 0.107 & 25.60 & +0.33 \\
EMAG &  8 & 0.50 & 58.24 & 0.686 & 0.675 & 0.712 & 0.107 & 25.49 & +0.22 \\
EMAG &  4 & 0.50 & 98.88 & 0.314 & 0.594 & 0.142 & 0.025 & 16.50 & +0.18 \\
EMAG &  4 & 0.30 & 99.09 & 0.291 & 0.593 & 0.142 & 0.025 & 16.33 & +0.01 \\
\bottomrule
\end{tabular}
}
\end{table*}

\begin{table*}[t]
\centering
\small
\setlength{\tabcolsep}{5pt}
\caption{\textbf{Effect of EMA decay rate $\beta$ on guidance quality.}
$\Delta$HPS reports improvement over the CFG-only baseline.
All experiments use SD3-Medium on 1{,}000 COCO 2014 validation prompts, 28 steps, CFG scale 7.0, $w_e{=}1.5$, seed 8.}
\label{tab:emag_beta_ablation}
\resizebox{\textwidth}{!}{%
\begin{tabular}{ccccccccc}
\toprule
$\boldsymbol{\beta}$ & \textbf{Half-life (steps)} & \textbf{FID} $\downarrow$ & \textbf{Precision} $\uparrow$ & \textbf{Recall} $\uparrow$ & \textbf{Density} $\uparrow$ & \textbf{Coverage} $\uparrow$ & \textbf{HPS} $\uparrow$ & $\boldsymbol{\Delta}$\textbf{HPS} \\
\midrule
\multicolumn{9}{l}{\textit{CFG-only baseline (no EMAG)}} \\
\midrule
--- & --- & 55.87 & 0.834 & 0.698 & 1.147 & 0.164 & 29.58 & --- \\
\midrule
\multicolumn{9}{l}{\textit{EMAG with varying $\beta$}} \\
\midrule
0.100 & $<$1  & 55.71 & 0.847 & 0.691 & 1.177 & 0.167 & 29.17 & $-$0.41 \\
0.500 & 1     & 55.52 & 0.856 & 0.690 & 1.157 & 0.165 & 29.20 & $-$0.38 \\
0.600 & 1.4   & 55.60 & 0.843 & 0.701 & 1.171 & 0.167 & 29.21 & $-$0.37 \\
0.800 & 3.1   & 55.67 & 0.836 & 0.696 & 1.148 & 0.164 & 29.27 & $-$0.31 \\
0.900 & 6.6   & 55.94 & 0.830 & 0.687 & 1.084 & 0.157 & 29.40 & $-$0.18 \\
0.950 & 13.5  & 56.31 & 0.829 & 0.676 & 1.142 & 0.162 & 29.51 & $-$0.07 \\
0.960 & 17.0  & 56.19 & 0.827 & 0.676 & 1.113 & 0.157 & 29.55 & $-$0.03 \\
\rowcolor{gray!10} 0.965 & 19.5  & 56.27 & 0.832 & 0.673 & 1.086 & 0.153 & 29.58 & $+$0.00 \\
0.975 & 27.4  & 56.27 & 0.842 & 0.670 & 1.080 & 0.153 & 29.62 & $+$0.04 \\
0.982 & 38.2  & 56.28 & 0.834 & 0.675 & 1.081 & 0.153 & 29.67 & $+$0.09 \\
0.988 & 57.4  & 56.19 & 0.830 & 0.669 & 1.089 & 0.156 & 29.72 & $+$0.14 \\
0.995 & 138   & 56.38 & 0.823 & 0.665 & 1.078 & 0.154 & 29.77 & $+$0.19 \\
0.999 & 693   & 56.39 & 0.829 & 0.662 & 1.071 & 0.152 & 29.80 & $+$0.22\\
\bottomrule
\end{tabular}
}
\end{table*}

\begin{table*}[t]
\centering
\small
\setlength{\tabcolsep}{5pt}
\caption{\textbf{Cross-seed standard deviation for SD3 text-conditional experiments.}
Each method is evaluated independently across 3 seeds (8, 56, 64) with 1{,}000 samples per seed.
Values report the standard deviation of each metric across the three per-seed evaluations, quantifying sensitivity to random initialisation.
Low values across all methods suggest that the metric rankings in the main table are stable.}
\label{tab:sd3_cond_std}
\resizebox{\textwidth}{!}{%
\begin{tabular}{lccccccc}
\toprule
\textbf{Method} & \textbf{FID} & \textbf{Precision} & \textbf{Recall} & \textbf{Density} & \textbf{Coverage} & \textbf{CLIP} & \textbf{HPS} $(\times 100)$ \\
\midrule
\multicolumn{8}{l}{\textit{SD3 (Text-Conditional) --- Std.\ dev.\ across 3 seeds (1K samples each)}} \\
\midrule
No Guidance & 0.25 & 0.020 & 0.012 & 0.044 & 0.0022 & 0.06 & 0.11 \\
CFG         & 0.88 & 0.008 & 0.013 & 0.043 & 0.0020 & 0.05 & 0.05 \\
\midrule
SAG+CFG     & 0.40 & 0.001 & 0.002 & 0.045 & 0.0021 & 0.03 & 0.04 \\
SEG+CFG     & 0.44 & 0.008 & 0.007 & 0.034 & 0.0009 & 0.01 & 0.04 \\
ERG         & 0.56 & 0.017 & 0.006 & 0.068 & 0.0038 & 0.04 & 0.03 \\
PAG+CFG     & 0.11 & 0.016 & 0.018 & 0.062 & 0.0047 & 0.04 & 0.04 \\
APG         & 0.78 & 0.009 & 0.024 & 0.043 & 0.0024 & 0.07 & 0.02 \\
S$^2$       & 0.44 & 0.009 & 0.006 & 0.035 & 0.0028 & 0.03 & 0.00 \\
\midrule
EMAG        & 0.16 & 0.003 & 0.012 & 0.064 & 0.0032 & 0.06 & 0.03 \\
EMAG+APG    & 1.08 & 0.015 & 0.008 & 0.052 & 0.0014 & 0.05 & 0.04 \\
EMAG+CADS   & 0.33 & 0.014 & 0.016 & 0.070 & 0.0030 & 0.03 & 0.02 \\
\bottomrule
\end{tabular}
}
\end{table*}

\clearpage
{\scriptsize
\begin{longtable}{llccccccc}
\caption{Full comparison of guidance methods with varying hyperparameters on DiT-XL/2 at $256{\times}256$ resolution. All experiments are conducted on 1{,}000 samples from the ImageNet 1K validation set using identical sampling steps, evaluation protocols, and a fixed seed of 8. Corresponding Pareto frontier curves are shown in Fig.~\ref{fig: Dit_256_pareto_frontier_by_method}.}
\label{tab:dit_guidance_ablation_256} \\
\toprule
\textbf{Method} & \textbf{Scale} & \textbf{CFG} & \textbf{FID} $\downarrow$ & \textbf{Precision} $\uparrow$ & \textbf{Recall} $\uparrow$ & \textbf{Density} $\uparrow$ & \textbf{Coverage} $\uparrow$ & \textbf{HPS} $\uparrow$\\
\midrule
\endfirsthead
\multicolumn{9}{l}{\small\textit{Table~\ref{tab:dit_guidance_ablation_256} continued from previous page}} \\
\toprule
\textbf{Method} & \textbf{Scale} & \textbf{CFG} & \textbf{FID} $\downarrow$ & \textbf{Precision} $\uparrow$ & \textbf{Recall} $\uparrow$ & \textbf{Density} $\uparrow$ & \textbf{Coverage} $\uparrow$ & \textbf{HPS} $\uparrow$\\
\midrule
\endhead
\midrule
\multicolumn{9}{r}{\small\textit{Continued on next page}} \\
\endfoot
\bottomrule
\endlastfoot
\multicolumn{9}{l}{\textit{EMAG (Ours)}} \\
\midrule
EMAG & $w_e=1.5$ & 1.5 & 45.38 & 0.846 & 0.726 & 1.205 & 0.429 & 23.65 \\
EMAG & $w_e=1.5$ & 3 & 57.00 & 0.924 & 0.701 & 1.338 & 0.462 & 25.38 \\
EMAG & $w_e=1.5$ & 5 & 61.88 & 0.914 & 0.686 & 1.217 & 0.415 & 25.77 \\
EMAG & $w_e=1.5$ & 7 & 61.80 & 0.899 & 0.697 & 1.146 & 0.395 & 25.82 \\
EMAG & $w_e=1.5$ & 9 & 60.86 & 0.862 & 0.707 & 1.076 & 0.374 & 25.65 \\
EMAG & $w_e=2.0$ & 1.5 & 46.18 & 0.848 & 0.719 & 1.213 & 0.432 & 23.94 \\
EMAG & $w_e=2.0$ & 3 & 58.00 & 0.916 & 0.701 & 1.306 & 0.452 & 25.43 \\
EMAG & $w_e=2.0$ & 5 & 61.28 & 0.912 & 0.702 & 1.207 & 0.414 & 25.79 \\
EMAG & $w_e=2.0$ & 7 & 60.35 & 0.866 & 0.705 & 1.097 & 0.383 & 25.72 \\
EMAG & $w_e=2.0$ & 9 & 60.50 & 0.853 & 0.707 & 1.057 & 0.363 & 25.50 \\
EMAG & $w_e=2.5$ & 1.5 & 47.06 & 0.856 & 0.718 & 1.241 & 0.439 & 24.16 \\
EMAG & $w_e=2.5$ & 3 & 57.42 & 0.912 & 0.709 & 1.241 & 0.433 & 25.46 \\
EMAG & $w_e=2.5$ & 5 & 59.72 & 0.892 & 0.703 & 1.168 & 0.408 & 25.73 \\
EMAG & $w_e=2.5$ & 7 & 59.19 & 0.834 & 0.712 & 1.088 & 0.380 & 25.62 \\
EMAG & $w_e=2.5$ & 9 & 58.60 & 0.794 & 0.711 & 1.007 & 0.350 & 25.33 \\
EMAG & $w_e=3.0$ & 1.5 & 47.61 & 0.873 & 0.709 & 1.272 & 0.444 & 24.32 \\
EMAG & $w_e=3.0$ & 3 & 55.98 & 0.909 & 0.700 & 1.226 & 0.431 & 25.46 \\
EMAG & $w_e=3.0$ & 5 & 58.15 & 0.878 & 0.700 & 1.150 & 0.402 & 25.64 \\
EMAG & $w_e=3.0$ & 7 & 58.49 & 0.824 & 0.707 & 1.051 & 0.370 & 25.44 \\
EMAG & $w_e=3.0$ & 9 & 57.70 & 0.763 & 0.703 & 0.944 & 0.338 & 25.07 \\
EMAG & $w_e=5.0$ & 1.5 & 48.52 & 0.847 & 0.703 & 1.256 & 0.441 & 24.67 \\
EMAG & $w_e=5.0$ & 3 & 53.49 & 0.848 & 0.707 & 1.158 & 0.418 & 25.18 \\
EMAG & $w_e=5.0$ & 5 & 55.04 & 0.787 & 0.709 & 1.016 & 0.369 & 24.95 \\
EMAG & $w_e=5.0$ & 7 & 56.76 & 0.704 & 0.711 & 0.842 & 0.306 & 24.47 \\
EMAG & $w_e=5.0$ & 9 & 63.29 & 0.575 & 0.690 & 0.669 & 0.249 & 23.93 \\
\midrule
\multicolumn{9}{l}{\textit{SAG}} \\
\midrule
SAG & $s=0.1$ & 1.5 & 55.75 & 0.928 & 0.709 & 1.358 & 0.460 & 25.42 \\
SAG & $s=0.1$ & 3 & 62.33 & 0.926 & 0.691 & 1.243 & 0.421 & 26.00 \\
SAG & $s=0.1$ & 5 & 63.72 & 0.916 & 0.690 & 1.168 & 0.395 & 26.11 \\
SAG & $s=0.1$ & 7 & 62.86 & 0.881 & 0.690 & 1.124 & 0.382 & 25.95 \\
SAG & $s=0.1$ & 9 & 62.38 & 0.840 & 0.708 & 1.019 & 0.353 & 25.71 \\
SAG & $s=0.25$ & 1.5 & 55.53 & 0.921 & 0.692 & 1.360 & 0.462 & 25.57 \\
SAG & $s=0.25$ & 3 & 62.33 & 0.922 & 0.683 & 1.248 & 0.424 & 26.09 \\
SAG & $s=0.25$ & 5 & 63.34 & 0.922 & 0.688 & 1.176 & 0.402 & 26.14 \\
SAG & $s=0.25$ & 7 & 62.33 & 0.873 & 0.680 & 1.130 & 0.385 & 25.99 \\
SAG & $s=0.25$ & 9 & 62.28 & 0.841 & 0.709 & 1.022 & 0.357 & 25.76 \\
SAG & $s=0.5$ & 1.5 & 55.49 & 0.917 & 0.684 & 1.339 & 0.456 & 25.68 \\
SAG & $s=0.5$ & 3 & 61.05 & 0.920 & 0.684 & 1.255 & 0.422 & 26.15 \\
SAG & $s=0.5$ & 5 & 62.38 & 0.904 & 0.694 & 1.180 & 0.401 & 26.14 \\
SAG & $s=0.5$ & 7 & 62.04 & 0.873 & 0.692 & 1.093 & 0.375 & 26.02 \\
SAG & $s=0.5$ & 9 & 62.30 & 0.834 & 0.703 & 1.004 & 0.350 & 25.81 \\
SAG & $s=0.75$ & 1.5 & 54.29 & 0.905 & 0.691 & 1.323 & 0.456 & 25.64 \\
SAG & $s=0.75$ & 3 & 60.07 & 0.919 & 0.687 & 1.230 & 0.420 & 26.10 \\
SAG & $s=0.75$ & 5 & 62.36 & 0.913 & 0.688 & 1.163 & 0.396 & 26.13 \\
SAG & $s=0.75$ & 7 & 61.43 & 0.861 & 0.695 & 1.089 & 0.376 & 26.00 \\
SAG & $s=0.75$ & 9 & 61.42 & 0.839 & 0.711 & 1.026 & 0.357 & 25.83 \\
SAG & $s=1.0$ & 1.5 & 53.68 & 0.896 & 0.672 & 1.297 & 0.449 & 25.56 \\
SAG & $s=1.0$ & 3 & 59.21 & 0.905 & 0.692 & 1.214 & 0.415 & 26.03 \\
SAG & $s=1.0$ & 5 & 61.43 & 0.907 & 0.691 & 1.153 & 0.395 & 26.07 \\
SAG & $s=1.0$ & 7 & 61.58 & 0.850 & 0.693 & 1.071 & 0.370 & 25.96 \\
SAG & $s=1.0$ & 9 & 61.06 & 0.836 & 0.701 & 1.035 & 0.362 & 25.82 \\
\midrule
\multicolumn{9}{l}{\textit{SEG}} \\
\midrule
SEG & $s=0.5$ & 1.5 & 49.11 & 0.913 & 0.694 & 1.381 & 0.472 & 25.58 \\
SEG & $s=0.5$ & 3 & 59.10 & 0.925 & 0.688 & 1.312 & 0.447 & 26.37 \\
SEG & $s=0.5$ & 5 & 62.32 & 0.910 & 0.683 & 1.225 & 0.416 & 26.48 \\
SEG & $s=0.5$ & 7 & 63.70 & 0.891 & 0.696 & 1.140 & 0.390 & 26.26 \\
SEG & $s=0.5$ & 9 & 61.86 & 0.853 & 0.692 & 1.090 & 0.376 & 26.00 \\
SEG & $s=1.0$ & 1.5 & 51.11 & 0.871 & 0.682 & 1.273 & 0.441 & 26.17 \\
SEG & $s=1.0$ & 3 & 58.28 & 0.896 & 0.688 & 1.246 & 0.427 & 26.65 \\
SEG & $s=1.0$ & 5 & 60.65 & 0.906 & 0.682 & 1.199 & 0.415 & 26.56 \\
SEG & $s=1.0$ & 7 & 61.71 & 0.882 & 0.684 & 1.157 & 0.393 & 26.33 \\
SEG & $s=1.0$ & 9 & 61.05 & 0.833 & 0.693 & 1.048 & 0.360 & 26.06 \\
SEG & $s=2.0$ & 1.5 & 52.80 & 0.842 & 0.647 & 1.145 & 0.398 & 26.15 \\
SEG & $s=2.0$ & 3 & 55.52 & 0.854 & 0.668 & 1.141 & 0.406 & 26.49 \\
SEG & $s=2.0$ & 5 & 57.77 & 0.867 & 0.671 & 1.165 & 0.404 & 26.41 \\
SEG & $s=2.0$ & 7 & 59.15 & 0.835 & 0.674 & 1.116 & 0.379 & 26.19 \\
SEG & $s=2.0$ & 9 & 59.66 & 0.810 & 0.685 & 1.015 & 0.350 & 25.91 \\
SEG & $s=3.0$ & 1.5 & 54.78 & 0.770 & 0.618 & 1.047 & 0.366 & 25.71 \\
SEG & $s=3.0$ & 3 & 55.43 & 0.782 & 0.645 & 1.051 & 0.372 & 26.00 \\
SEG & $s=3.0$ & 5 & 57.04 & 0.805 & 0.652 & 1.024 & 0.366 & 25.93 \\
SEG & $s=3.0$ & 7 & 58.42 & 0.769 & 0.663 & 0.995 & 0.344 & 25.84 \\
SEG & $s=3.0$ & 9 & 59.97 & 0.741 & 0.657 & 0.937 & 0.332 & 25.55 \\
SEG & $s=4.0$ & 1.5 & 58.86 & 0.681 & 0.584 & 0.917 & 0.322 & 25.05 \\
SEG & $s=4.0$ & 3 & 57.66 & 0.726 & 0.603 & 0.964 & 0.339 & 25.36 \\
SEG & $s=4.0$ & 5 & 58.54 & 0.705 & 0.627 & 0.927 & 0.331 & 25.48 \\
SEG & $s=4.0$ & 7 & 59.97 & 0.707 & 0.633 & 0.913 & 0.316 & 25.34 \\
SEG & $s=4.0$ & 9 & 61.69 & 0.680 & 0.645 & 0.804 & 0.289 & 25.07 \\
\midrule
\multicolumn{9}{l}{\textit{ERG}} \\
\midrule
ERG & -- & 1.5 & 45.26 & 0.862 & 0.698 & 1.276 & 0.453 & 24.44 \\
ERG & -- & 3 & 53.18 & 0.863 & 0.657 & 1.239 & 0.419 & 25.93 \\
ERG & -- & 5 & 55.93 & 0.766 & 0.627 & 1.046 & 0.371 & 25.39 \\
ERG & -- & 7 & 60.78 & 0.627 & 0.585 & 0.819 & 0.291 & 24.63 \\
ERG & -- & 9 & 71.02 & 0.518 & 0.534 & 0.610 & 0.220 & 23.81 \\
\midrule
\multicolumn{9}{l}{\textit{S$^2$}} \\
\midrule
S$^2$ & $w=0.10$ & 1.5 & 84.05 & 0.364 & 0.622 & 0.369 & 0.147 & 21.92 \\
S$^2$ & $w=0.10$ & 3 & 56.17 & 0.767 & 0.714 & 0.947 & 0.344 & 24.57 \\
S$^2$ & $w=0.10$ & 5 & 58.87 & 0.842 & 0.715 & 1.047 & 0.373 & 25.24 \\
S$^2$ & $w=0.10$ & 7 & 59.90 & 0.840 & 0.718 & 1.032 & 0.367 & 25.31 \\
S$^2$ & $w=0.10$ & 9 & 60.32 & 0.779 & 0.722 & 0.945 & 0.335 & 25.04 \\
S$^2$ & $w=0.15$ & 1.5 & 137.11 & 0.231 & 0.480 & 0.147 & 0.044 & 20.06 \\
S$^2$ & $w=0.15$ & 3 & 67.34 & 0.530 & 0.678 & 0.576 & 0.225 & 23.38 \\
S$^2$ & $w=0.15$ & 5 & 60.31 & 0.696 & 0.706 & 0.822 & 0.301 & 24.44 \\
S$^2$ & $w=0.15$ & 7 & 61.30 & 0.694 & 0.710 & 0.821 & 0.295 & 24.58 \\
S$^2$ & $w=0.15$ & 9 & 62.80 & 0.645 & 0.711 & 0.735 & 0.273 & 24.41 \\
S$^2$ & $w=0.25$ & 1.5 & 230.60 & 0.184 & 0.081 & 0.121 & 0.004 & 16.70 \\
S$^2$ & $w=0.25$ & 3 & 150.23 & 0.210 & 0.392 & 0.127 & 0.031 & 20.03 \\
S$^2$ & $w=0.25$ & 5 & 97.92 & 0.327 & 0.549 & 0.283 & 0.108 & 22.04 \\
S$^2$ & $w=0.25$ & 7 & 92.48 & 0.349 & 0.586 & 0.331 & 0.123 & 22.29 \\
S$^2$ & $w=0.25$ & 9 & 94.17 & 0.341 & 0.553 & 0.329 & 0.116 & 22.11 \\
S$^2$ & $w=0.30$ & 1.5 & 270.70 & 0.190 & 0.000 & 0.178 & 0.001 & 15.49 \\
S$^2$ & $w=0.30$ & 3 & 211.70 & 0.186 & 0.136 & 0.116 & 0.007 & 17.24 \\
S$^2$ & $w=0.30$ & 5 & 158.16 & 0.197 & 0.361 & 0.121 & 0.028 & 19.21 \\
S$^2$ & $w=0.30$ & 7 & 137.69 & 0.229 & 0.426 & 0.156 & 0.047 & 19.83 \\
S$^2$ & $w=0.30$ & 9 & 142.14 & 0.177 & 0.436 & 0.133 & 0.042 & 19.31 \\
\end{longtable}
}

\clearpage
{\scriptsize
\begin{longtable}{llccccccc}
\caption{Full comparison of guidance methods with varying hyperparameters on DiT-XL/2 at $512{\times}512$ resolution. All experiments are conducted on 1{,}000 samples from the ImageNet 1K validation set using identical sampling steps, evaluation protocols, and a fixed seed of 8. Corresponding Pareto frontier curves are shown in Fig.~\ref{fig: Dit_512_pareto_frontier_by_method}.}
\label{tab:dit_guidance_ablation_512} \\
\toprule
\textbf{Method} & \textbf{Scale} & \textbf{CFG} & \textbf{FID} $\downarrow$ & \textbf{Precision} $\uparrow$ & \textbf{Recall} $\uparrow$ & \textbf{Density} $\uparrow$ & \textbf{Coverage} $\uparrow$ & \textbf{HPS} $\uparrow$\\
\midrule
\endfirsthead
\multicolumn{9}{l}{\small\textit{Table~\ref{tab:dit_guidance_ablation_512} continued from previous page}} \\
\toprule
\textbf{Method} & \textbf{Scale} & \textbf{CFG} & \textbf{FID} $\downarrow$ & \textbf{Precision} $\uparrow$ & \textbf{Recall} $\uparrow$ & \textbf{Density} $\uparrow$ & \textbf{Coverage} $\uparrow$ & \textbf{HPS} $\uparrow$\\
\midrule
\endhead
\midrule
\multicolumn{9}{r}{\small\textit{Continued on next page}} \\
\endfoot
\bottomrule
\endlastfoot
\multicolumn{9}{l}{\textit{EMAG (Ours)}} \\
\midrule
EMAG & $w_e=1.5$ & 1.5 & 44.30 & 0.858 & 0.697 & 1.324 & 0.977 & 23.70 \\
EMAG & $w_e=1.5$ & 3 & 55.40 & 0.851 & 0.655 & 1.222 & 0.963 & 25.71 \\
EMAG & $w_e=1.5$ & 5 & 59.56 & 0.854 & 0.673 & 1.002 & 0.946 & 26.24 \\
EMAG & $w_e=1.5$ & 7 & 60.78 & 0.844 & 0.667 & 0.985 & 0.936 & 26.32 \\
EMAG & $w_e=1.5$ & 9 & 60.04 & 0.827 & 0.678 & 1.002 & 0.926 & 26.23 \\
EMAG & $w_e=2.0$ & 1.5 & 44.52 & 0.855 & 0.668 & 1.416 & 0.981 & 24.10 \\
EMAG & $w_e=2.0$ & 3 & 54.24 & 0.857 & 0.665 & 1.305 & 0.971 & 25.80 \\
EMAG & $w_e=2.0$ & 5 & 58.05 & 0.846 & 0.676 & 0.973 & 0.958 & 26.17 \\
EMAG & $w_e=2.0$ & 7 & 58.17 & 0.835 & 0.676 & 0.973 & 0.935 & 26.11 \\
EMAG & $w_e=2.0$ & 9 & 57.34 & 0.829 & 0.687 & 0.995 & 0.941 & 25.92 \\
EMAG & $w_e=2.5$ & 1.5 & 44.41 & 0.873 & 0.678 & 1.418 & 0.985 & 24.37 \\
EMAG & $w_e=2.5$ & 3 & 52.68 & 0.858 & 0.692 & 1.159 & 0.969 & 25.82 \\
EMAG & $w_e=2.5$ & 5 & 55.57 & 0.844 & 0.680 & 1.078 & 0.963 & 25.96 \\
EMAG & $w_e=2.5$ & 7 & 55.55 & 0.825 & 0.692 & 0.959 & 0.948 & 25.75 \\
EMAG & $w_e=2.5$ & 9 & 55.41 & 0.820 & 0.678 & 0.945 & 0.928 & 25.45 \\
EMAG & $w_e=3.0$ & 1.5 & 44.82 & 0.875 & 0.668 & 1.432 & 0.982 & 24.52 \\
EMAG & $w_e=3.0$ & 3 & 51.27 & 0.864 & 0.678 & 1.163 & 0.968 & 25.72 \\
EMAG & $w_e=3.0$ & 5 & 53.26 & 0.850 & 0.677 & 1.064 & 0.963 & 25.67 \\
EMAG & $w_e=3.0$ & 7 & 53.26 & 0.824 & 0.668 & 1.032 & 0.948 & 25.32 \\
EMAG & $w_e=3.0$ & 9 & 54.39 & 0.784 & 0.676 & 0.924 & 0.939 & 24.99 \\
EMAG & $w_e=5.0$ & 1.5 & 45.35 & 0.868 & 0.669 & 1.399 & 0.982 & 24.71 \\
EMAG & $w_e=5.0$ & 3 & 49.58 & 0.855 & 0.668 & 1.143 & 0.961 & 24.90 \\
EMAG & $w_e=5.0$ & 5 & 54.60 & 0.770 & 0.652 & 0.984 & 0.933 & 24.08 \\
EMAG & $w_e=5.0$ & 7 & 62.68 & 0.730 & 0.600 & 0.931 & 0.929 & 23.50 \\
EMAG & $w_e=5.0$ & 9 & 74.76 & 0.669 & 0.554 & 0.778 & 0.864 & 23.00 \\
EMAG & $w_e=7.0$ & 1.5 & 47.11 & 0.864 & 0.650 & 1.387 & 0.979 & 24.52 \\
EMAG & $w_e=7.0$ & 3 & 54.52 & 0.791 & 0.640 & 1.076 & 0.938 & 23.76 \\
EMAG & $w_e=7.0$ & 5 & 73.34 & 0.666 & 0.556 & 0.921 & 0.930 & 22.70 \\
EMAG & $w_e=7.0$ & 7 & 90.56 & 0.623 & 0.456 & 0.673 & 0.722 & 22.09 \\
EMAG & $w_e=7.0$ & 9 & 108.02 & 0.518 & 0.411 & 0.562 & 0.630 & 21.34 \\
\midrule
\multicolumn{9}{l}{\textit{SAG}} \\
\midrule
SAG & $s=0.1$ & 1.5 & 53.65 & 0.857 & 0.698 & 1.190 & 0.971 & 25.71 \\
SAG & $s=0.1$ & 3 & 60.78 & 0.850 & 0.663 & 1.056 & 0.959 & 26.48 \\
SAG & $s=0.1$ & 5 & 63.12 & 0.820 & 0.683 & 0.949 & 0.921 & 26.68 \\
SAG & $s=0.1$ & 7 & 63.08 & 0.822 & 0.674 & 0.971 & 0.940 & 26.61 \\
SAG & $s=0.1$ & 9 & 62.82 & 0.801 & 0.663 & 0.908 & 0.931 & 26.46 \\
SAG & $s=0.25$ & 1.5 & 53.32 & 0.865 & 0.684 & 1.240 & 0.969 & 25.86 \\
SAG & $s=0.25$ & 3 & 59.92 & 0.848 & 0.662 & 1.002 & 0.947 & 26.52 \\
SAG & $s=0.25$ & 5 & 62.57 & 0.826 & 0.674 & 0.946 & 0.928 & 26.77 \\
SAG & $s=0.25$ & 7 & 62.79 & 0.813 & 0.673 & 1.013 & 0.929 & 26.65 \\
SAG & $s=0.25$ & 9 & 62.24 & 0.814 & 0.659 & 0.862 & 0.928 & 26.48 \\
SAG & $s=0.5$ & 1.5 & 53.16 & 0.852 & 0.667 & 1.204 & 0.963 & 25.96 \\
SAG & $s=0.5$ & 3 & 59.83 & 0.828 & 0.665 & 1.065 & 0.949 & 26.52 \\
SAG & $s=0.5$ & 5 & 62.46 & 0.817 & 0.646 & 1.036 & 0.945 & 26.65 \\
SAG & $s=0.5$ & 7 & 62.35 & 0.814 & 0.665 & 1.099 & 0.935 & 26.63 \\
SAG & $s=0.5$ & 9 & 62.69 & 0.804 & 0.659 & 0.973 & 0.937 & 26.43 \\
SAG & $s=0.75$ & 1.5 & 52.24 & 0.851 & 0.653 & 1.176 & 0.967 & 26.00 \\
SAG & $s=0.75$ & 3 & 58.40 & 0.828 & 0.643 & 1.170 & 0.953 & 26.49 \\
SAG & $s=0.75$ & 5 & 61.22 & 0.810 & 0.648 & 1.059 & 0.943 & 26.61 \\
SAG & $s=0.75$ & 7 & 61.04 & 0.810 & 0.651 & 1.159 & 0.944 & 26.60 \\
SAG & $s=0.75$ & 9 & 61.27 & 0.811 & 0.653 & 0.988 & 0.940 & 26.40 \\
SAG & $s=1.0$ & 1.5 & 51.64 & 0.853 & 0.649 & 1.205 & 0.966 & 25.91 \\
SAG & $s=1.0$ & 3 & 56.91 & 0.827 & 0.650 & 1.187 & 0.959 & 26.41 \\
SAG & $s=1.0$ & 5 & 59.96 & 0.817 & 0.646 & 1.093 & 0.943 & 26.53 \\
SAG & $s=1.0$ & 7 & 60.42 & 0.812 & 0.658 & 1.021 & 0.948 & 26.49 \\
SAG & $s=1.0$ & 9 & 60.41 & 0.813 & 0.656 & 0.984 & 0.947 & 26.33 \\
\midrule
\multicolumn{9}{l}{\textit{SEG}} \\
\midrule
SEG & $s=0.5$ & 1.5 & 47.73 & 0.864 & 0.662 & 1.414 & 0.976 & 25.62 \\
SEG & $s=0.5$ & 3 & 55.89 & 0.862 & 0.642 & 1.232 & 0.960 & 26.60 \\
SEG & $s=0.5$ & 5 & 59.30 & 0.838 & 0.678 & 0.980 & 0.934 & 26.84 \\
SEG & $s=0.5$ & 7 & 60.20 & 0.834 & 0.653 & 1.053 & 0.949 & 26.72 \\
SEG & $s=0.5$ & 9 & 60.68 & 0.808 & 0.658 & 1.150 & 0.946 & 26.66 \\
SEG & $s=1.0$ & 1.5 & 49.80 & 0.856 & 0.641 & 1.271 & 0.966 & 25.60 \\
SEG & $s=1.0$ & 3 & 54.13 & 0.833 & 0.653 & 1.201 & 0.962 & 26.36 \\
SEG & $s=1.0$ & 5 & 57.16 & 0.842 & 0.660 & 1.068 & 0.943 & 26.61 \\
SEG & $s=1.0$ & 7 & 58.30 & 0.841 & 0.631 & 1.205 & 0.965 & 26.55 \\
SEG & $s=1.0$ & 9 & 58.68 & 0.828 & 0.660 & 1.020 & 0.944 & 26.46 \\
SEG & $s=2.0$ & 1.5 & 56.50 & 0.791 & 0.571 & 1.405 & 0.962 & 24.66 \\
SEG & $s=2.0$ & 3 & 55.51 & 0.805 & 0.606 & 1.291 & 0.952 & 25.38 \\
SEG & $s=2.0$ & 5 & 55.74 & 0.826 & 0.626 & 1.196 & 0.953 & 25.68 \\
SEG & $s=2.0$ & 7 & 57.49 & 0.803 & 0.623 & 1.089 & 0.957 & 25.77 \\
SEG & $s=2.0$ & 9 & 58.45 & 0.819 & 0.641 & 1.067 & 0.961 & 25.84 \\
SEG & $s=3.0$ & 1.5 & 68.23 & 0.694 & 0.522 & 1.075 & 0.908 & 23.48 \\
SEG & $s=3.0$ & 3 & 63.82 & 0.746 & 0.538 & 1.043 & 0.926 & 24.20 \\
SEG & $s=3.0$ & 5 & 61.08 & 0.769 & 0.574 & 0.991 & 0.936 & 24.64 \\
SEG & $s=3.0$ & 7 & 59.70 & 0.769 & 0.591 & 0.970 & 0.935 & 24.91 \\
SEG & $s=3.0$ & 9 & 60.15 & 0.771 & 0.597 & 1.013 & 0.951 & 25.05 \\
SEG & $s=4.0$ & 1.5 & 81.78 & 0.608 & 0.449 & 0.767 & 0.818 & 22.39 \\
SEG & $s=4.0$ & 3 & 74.08 & 0.668 & 0.499 & 0.843 & 0.830 & 23.12 \\
SEG & $s=4.0$ & 5 & 69.14 & 0.700 & 0.509 & 0.894 & 0.891 & 23.71 \\
SEG & $s=4.0$ & 7 & 67.13 & 0.709 & 0.533 & 0.933 & 0.903 & 24.06 \\
SEG & $s=4.0$ & 9 & 65.98 & 0.717 & 0.560 & 1.010 & 0.926 & 24.20 \\
\midrule
\multicolumn{9}{l}{\textit{ERG}} \\
\midrule
ERG & $\text{--}$ & 1.5 & 44.58 & 0.854 & 0.665 & 1.371 & 0.969 & 24.28 \\
ERG & $\text{--}$ & 3 & 55.65 & 0.793 & 0.596 & 1.408 & 0.969 & 24.70 \\
ERG & $\text{--}$ & 5 & 70.95 & 0.683 & 0.506 & 1.006 & 0.910 & 23.35 \\
ERG & $\text{--}$ & 7 & 88.83 & 0.575 & 0.440 & 0.716 & 0.753 & 22.37 \\
ERG & $\text{--}$ & 9 & 105.14 & 0.483 & 0.368 & 0.537 & 0.575 & 21.55 \\
\midrule
\multicolumn{9}{l}{\textit{S$^2$}} \\
\midrule
S$^2$ & $w=0.10$ & 1.5 & 123.62 & 0.345 & 0.506 & 0.266 & 0.603 & 20.65 \\
S$^2$ & $w=0.10$ & 3 & 58.26 & 0.740 & 0.674 & 0.793 & 0.926 & 24.27 \\
S$^2$ & $w=0.10$ & 5 & 57.35 & 0.798 & 0.691 & 0.844 & 0.937 & 25.53 \\
S$^2$ & $w=0.10$ & 7 & 58.38 & 0.801 & 0.680 & 0.916 & 0.933 & 25.83 \\
S$^2$ & $w=0.10$ & 9 & 59.08 & 0.795 & 0.686 & 0.783 & 0.916 & 25.84 \\
S$^2$ & $w=0.15$ & 1.5 & 195.93 & 0.126 & 0.218 & 0.077 & 0.153 & 18.54 \\
S$^2$ & $w=0.15$ & 3 & 94.37 & 0.482 & 0.563 & 0.393 & 0.752 & 22.53 \\
S$^2$ & $w=0.15$ & 5 & 62.71 & 0.711 & 0.676 & 0.596 & 0.901 & 24.37 \\
S$^2$ & $w=0.15$ & 7 & 59.40 & 0.767 & 0.699 & 0.725 & 0.932 & 24.91 \\
S$^2$ & $w=0.15$ & 9 & 58.66 & 0.761 & 0.681 & 0.747 & 0.915 & 25.09 \\
S$^2$ & $w=0.25$ & 1.5 & 305.63 & 0.163 & 0.001 & 0.042 & 0.017 & 15.36 \\
S$^2$ & $w=0.25$ & 3 & 239.29 & 0.115 & 0.141 & 0.044 & 0.070 & 15.92 \\
S$^2$ & $w=0.25$ & 5 & 161.76 & 0.213 & 0.325 & 0.176 & 0.462 & 16.78 \\
S$^2$ & $w=0.25$ & 7 & 120.36 & 0.351 & 0.442 & 0.301 & 0.609 & 17.26 \\
S$^2$ & $w=0.25$ & 9 & 102.15 & 0.438 & 0.513 & 0.387 & 0.736 & 17.34 \\
S$^2$ & $w=0.30$ & 1.5 & 383.36 & 0.040 & 0.000 & 0.009 & 0.003 & 15.28 \\
S$^2$ & $w=0.30$ & 3 & 338.12 & 0.055 & 0.000 & 0.014 & 0.008 & 15.35 \\
S$^2$ & $w=0.30$ & 5 & 266.07 & 0.065 & 0.076 & 0.033 & 0.061 & 15.59 \\
S$^2$ & $w=0.30$ & 7 & 217.38 & 0.086 & 0.311 & 0.060 & 0.171 & 15.87 \\
S$^2$ & $w=0.30$ & 9 & 190.46 & 0.123 & 0.289 & 0.071 & 0.167 & 16.07 \\
\end{longtable}
}



\begin{figure}[h]
\vspace{-1.0em}
    \centering
    \begin{subfigure}[b]{1\linewidth}
        \centering
        \includegraphics[width=\linewidth]{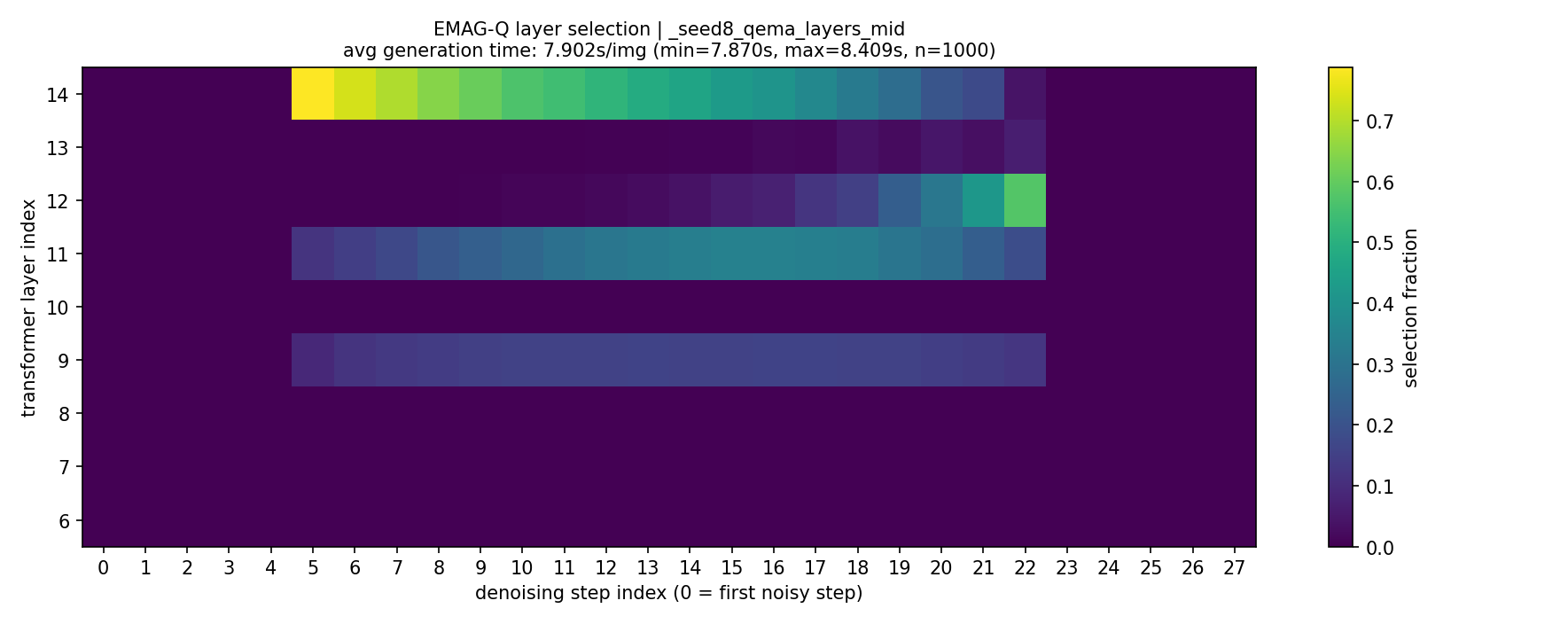}
        \caption{EMAG-Q (L[9-14]).}
        \label{fig:heatmap_qema_mid}
    \end{subfigure}

    \vspace{0.5em}

    \begin{subfigure}[b]{1\linewidth}
        \centering
        \includegraphics[width=\linewidth]{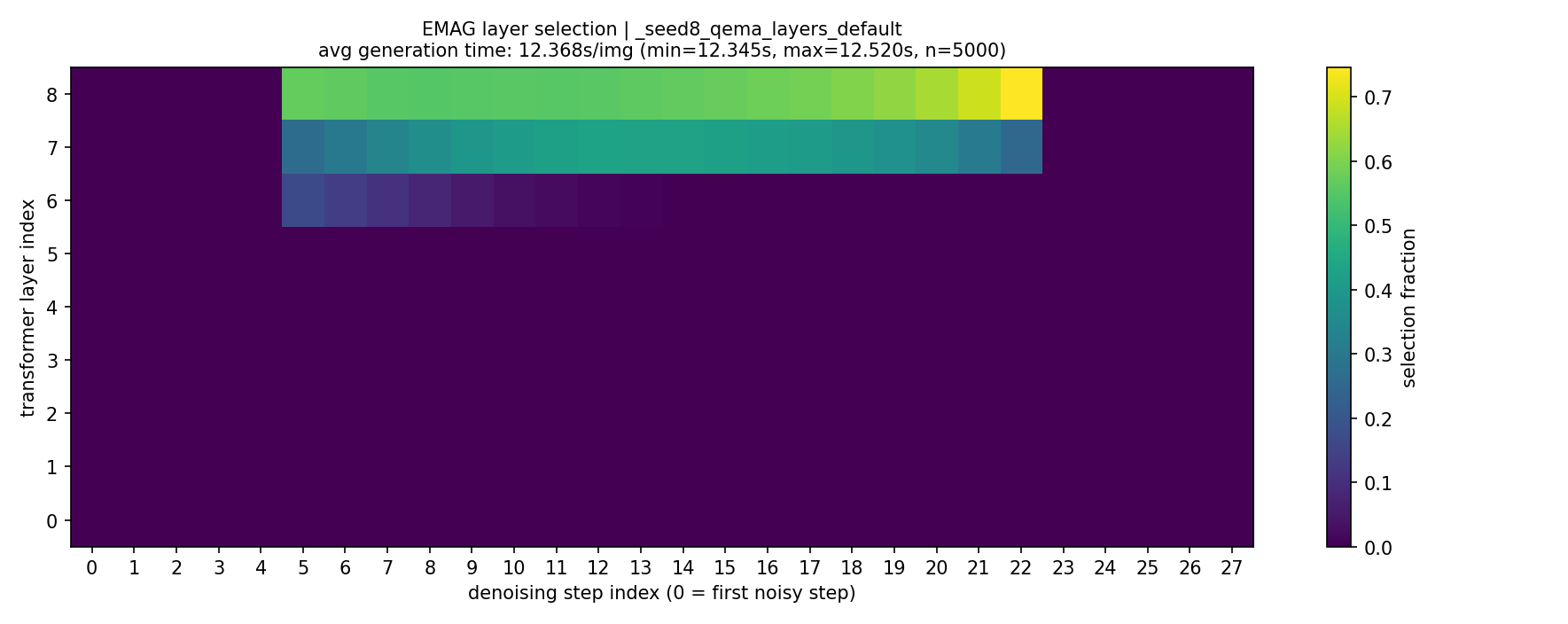}
        \caption{EMAG (L[6-8]).}
        \label{fig:heatmap_full_emag}
    \end{subfigure}
    \caption{\textbf{Adaptive layer selection is timestep-varying.} Frequency with which each layer is selected across timesteps, aggregated over 1000 generations. Attention-gradient statistics drive a non-uniform, timestep-dependent choice of perturbation layer rather than a fixed one.}
    \label{fig:layer_selection_heatmaps}
\end{figure}

%
%

\end{document}